\documentclass{article}

\usepackage[final, nonatbib]{neurips_2021}
\usepackage[numbers]{natbib}

\usepackage[utf8]{inputenc} 
\usepackage[T1]{fontenc}    
\usepackage{hyperref}       
\usepackage{url}            
\usepackage{booktabs}       
\usepackage{amsfonts}       
\usepackage{nicefrac}       
\usepackage{microtype}      
\usepackage{xcolor}         
\usepackage{adjustbox}

\usepackage{amsmath,amsthm,amssymb}
\usepackage{algorithm}
\usepackage{algorithmic}
\usepackage{comment}
\usepackage{mathtools}
\usepackage{caption}
\usepackage{float}
\usepackage{graphicx}
\usepackage{subfigure}
\usepackage{multirow}
\usepackage{mathrsfs}
\usepackage{newfloat}
\usepackage{relsize}
\usepackage{enumitem}
\usepackage{nicefrac}
\usepackage{pifont}

\title{How Powerful are Performance Predictors\\ in Neural Architecture Search?}

\author{Colin White$^1$\thanks{\texttt{\{colin, yang\}@abacus.ai}, 
\texttt{\{zelaa, fh\}@cs.uni-freiburg.de}, \texttt{robin@robots.ox.ac.uk}
} 
, Arber Zela$^2$, Binxin Ru$^3$, Yang Liu$^1$, 
Frank Hutter$^{2,4}$ \\
    $^1$ Abacus.AI, $^2$ University of Freiburg, $^3$ University of Oxford, \\
    $^4$ Bosch Center for Artificial Intelligence\\
  }

\begin{document}

\maketitle

\begin{abstract}
Early methods in the rapidly developing field of neural architecture search (NAS) required fully training thousands of neural networks. 
To reduce this extreme computational cost, 
dozens of techniques have since been proposed to predict the final performance of neural architectures.
Despite the success of such performance prediction methods, it is not well-understood how
different families of techniques compare to one another, due to
the lack of an agreed-upon evaluation metric and optimization for different constraints on 
the initialization time and query time.
In this work, we give the first large-scale study of performance predictors by analyzing 
31 techniques ranging from learning curve extrapolation, to weight-sharing, to supervised learning, to zero-cost proxies.
We test a number of correlation- and rank-based performance measures in a variety
of settings, as well as the ability of each technique to speed up predictor-based 
NAS frameworks. Our results act as recommendations for the best predictors to use in 
different settings, and we show that certain families of predictors can 
be combined to achieve even better predictive power, opening up
promising research directions. 
Our code, featuring a library of 31 performance predictors, is available at 
\url{https://github.com/automl/naslib}.
\end{abstract}

\section{Introduction} \label{sec:intro}
Neural architecture search (NAS) is a popular area of machine learning, which aims to automate the process of developing neural architectures 
for a given dataset. Since 2017, a wide variety of NAS
techniques have been proposed~\citep{zoph2017neural, enas, darts, real2019regularized}.
While the first NAS techniques trained thousands of 
architectures to completion and then evaluated the performance using the 
final validation accuracy~\citep{zoph2017neural}, modern algorithms use more
efficient strategies to estimate the performance of partially-trained or even
untrained neural networks~\citep{domhan2015speeding,baker2017accelerating,bonas,seminas,mellor2020neural}. 

Recently, many performance prediction methods have
been proposed based on training a model to predict the final validation accuracy
of an architecture just from an encoding of the architecture. Popular choices for these models include Gaussian processes~\citep{SweDuvSnoHutOsb13, nasbot, nasbowl},
neural networks~\citep{ma2019deep, bonas, alphax, bananas}, 
tree-based methods~\citep{luo2020neural, nasbench301}, and so on.
However, these methods often require hundreds of fully-trained architectures to be 
used as training data, thus incurring high initialization time.
In contrast, learning curve extrapolation 
methods~\citep{domhan2015speeding, baker2017accelerating, lcnet} need little or no 
initialization time, but each individual prediction requires partially training the architecture, incurring high query time.
Very recently, a few techniques have been introduced which are fast both
in query time and initialization time~\citep{mellor2020neural, abdelfattah2021zerocost}, 
computing predictions based on a single minibatch of data.
Finally, using shared weights~\citep{enas, bender2018understanding, darts} is a popular paradigm 
for NAS~\citep{zela2020understanding, li2020geometry},
although the effectiveness of these methods in ranking architectures
is disputed~\citep{sciuto2019evaluating, zela2020bench, zhang2020deeper}.

Despite the widespread use of performance predictors, it is not known 
how methods from different families compare to one 
another. While there have been some analyses on the best predictors within
each class~\citep{ning2020surgery, yu2020train}, for many predictors,
the only evaluation is from the original work that proposed the method.
Furthermore, no work has previously compared the predictors \emph{across} 
different families of performance predictors.
This leads to two natural questions: how do zero-cost methods, 
model-based methods, learning curve extrapolation methods,
and weight sharing methods compare to one another across different constraints 
on initialization time and query time? Furthermore, can predictors from different 
families be combined to achieve even better performance?

In this work, we answer the above questions by giving the first large-scale study of 
performance predictors for NAS. We study 31 
predictors across four 
popular search spaces and four datasets:
NAS-Bench-201 \citep{nasbench201} with CIFAR-10, CIFAR-100, and ImageNet16-120,
NAS-Bench-101 \citep{nasbench} and DARTS \citep{darts} with CIFAR-10, and
NAS-Bench-NLP \citep{nasbenchnlp} with Penn TreeBank.
In order to give a fair comparison among different classes of predictors,
we run a full portfolio of experiments, measuring the Pearson correlation
and rank correlation metrics (Spearman, Kendall Tau, and sparse Kendall Tau),
across a variety of initialization time and query time budgets.
We run experiments using a training and test set of architectures generated both uniformly at random, as well as by mutating the highest-performing architectures (the latter potentially more closely resembling distributions encountered during an actual NAS run).
Finally, we test the ability of each predictor to speed up NAS algorithms,
namely Bayesian optimization~\citep{ma2019deep, bonas, bananas, nasbowl}
and predictor-guided evolution~\citep{npenas, sun2020new}.

\begin{figure}
\centering
\raisebox{0.1\height}{\includegraphics[width=.39\columnwidth]{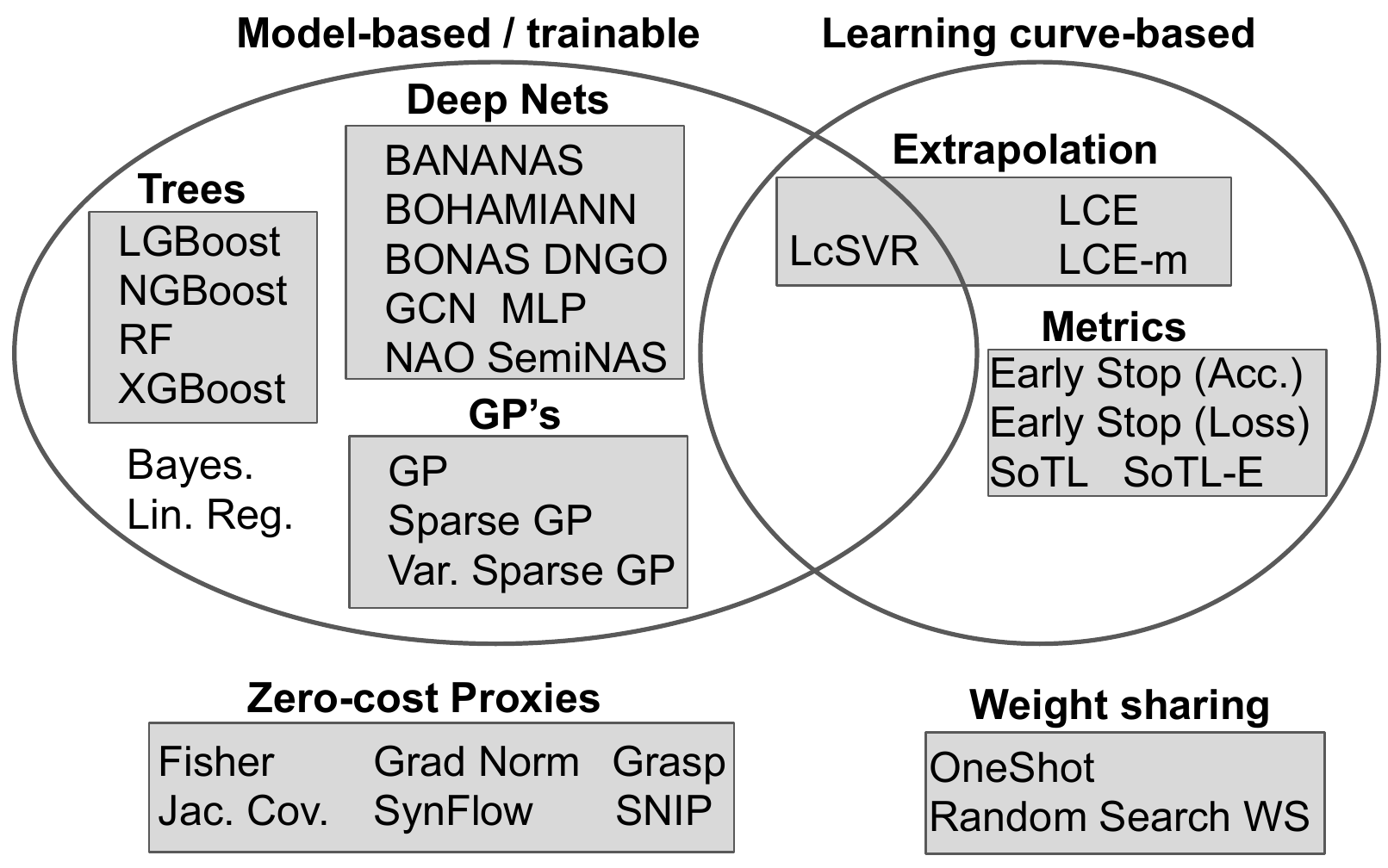}}
\raisebox{0.0\height}{\includegraphics[width=.6\columnwidth]{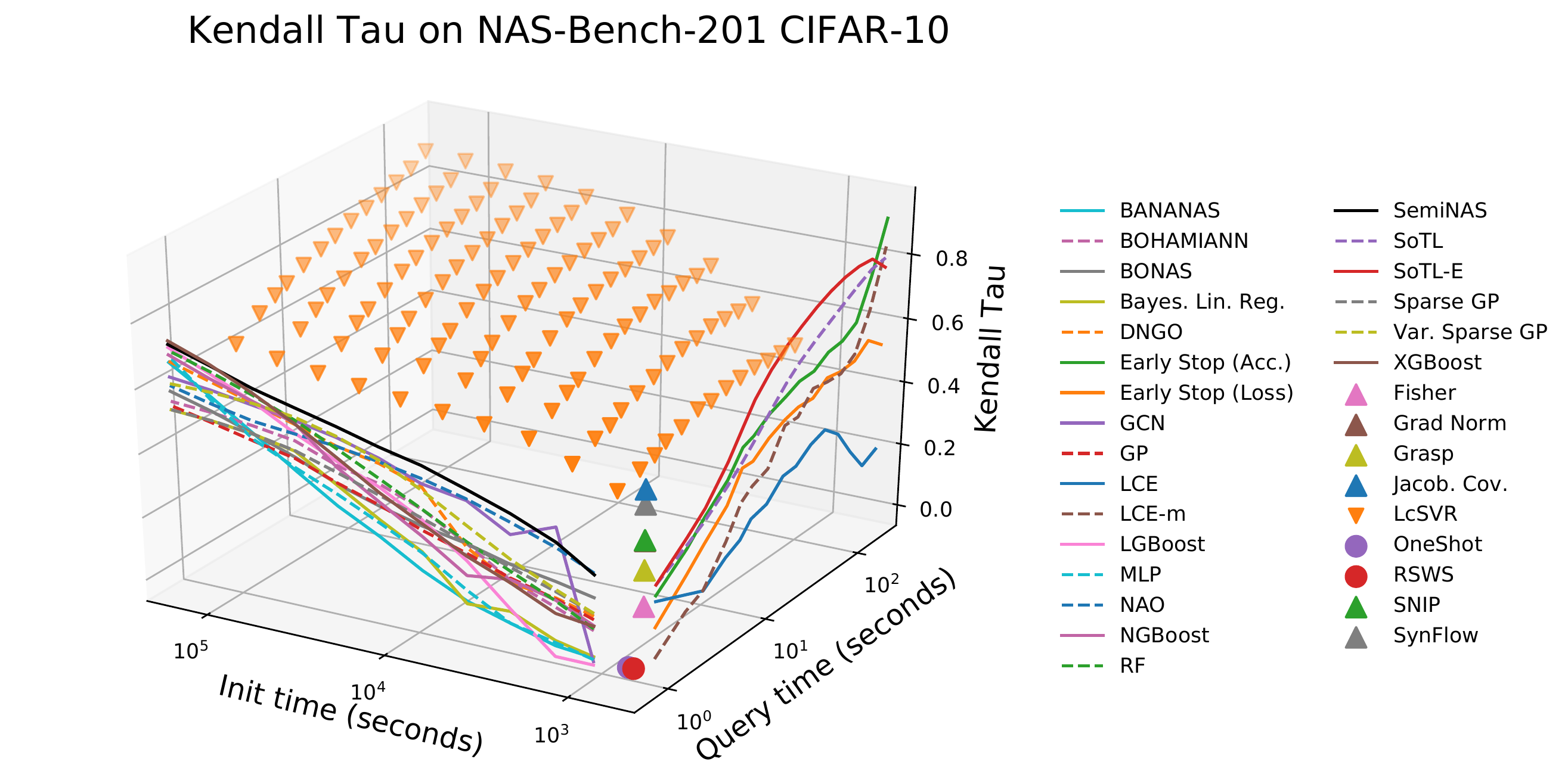}}
\caption{
Categories of performance predictors (left). 
%
Kendall Tau rank correlation for performance predictors 
with respect to initialization time and query time (right). Each type of
predictor is plotted differently based on whether it allows
variable initialization time and/or variable query time. For example, 
the sixteen model-based predictors have a fixed query time and
variable initialization time, so they are plotted as curves parallel to the
X-Z plane.
}
\label{fig:3d}
\end{figure}

Since many predictors so far had only been evaluated on one search space, 
our work shows which predictors have consistent performance across search
spaces. 
Furthermore, by conducting a study with three axes of comparison 
(see Figure~\ref{fig:3d}),
and by comparing various types of predictors, we see 
a more complete view of the state of performance predictor techniques
that leads to interesting insights. 
Notably, we show that the performance of predictors from different families
are complementary and can be combined to achieve significantly higher performance.
The success of these experiments opens up promising avenues for future work.



Overall, our experiments bridge multiple areas of NAS research and act as 
recommendations for the best predictors to use under different runtime constraints.
Our code, based on the NASLib library~\citep{ruchte2020naslib}, 
can be used as a testing ground for future performance prediction techniques.
In order to ensure reproducibility of the original results, we created a table to clarify which of the 31 predictors had previously published results on a NAS-Bench search space, and how these published results compared to our results (Table \ref{tab:reproduce}).
We also adhere to the specialized NAS best practices checklist~\citep{lindauer2019best}.

\noindent\textbf{Our contributions.}
We summarize our main contributions below.
\begin{itemize}[topsep=0pt, itemsep=2pt, parsep=0pt, leftmargin=5mm]
    \item 
    We conduct the first large-scale study of performance predictors for 
    neural architecture search by comparing model-based methods, learning
    curve extrapolation methods, zero-cost methods, and weight sharing
    methods across a variety of settings.
    \item 
    We release a comprehensive library of 31 performance predictors
    on four different search spaces.
    \item 
    We show that different families of performance predictors can be 
    combined to achieve substantially 
    better predictive power than any single predictor.
\end{itemize}

\section{Related Work} \label{sec:relatedwork}

NAS has been studied since at least the 1990s~\citep{kitano1990designing, stanley2002evolving}, 
and has been revitalized in the last few years~\citep{zoph2017neural}.
While initial techniques focused on reinforcement learning~\citep{zoph2017neural, enas}
and evolutionary search~\citep{maziarz2018evolutionary, real2019regularized}, 
one-shot NAS algorithms~\citep{darts, gdas, bender2018understanding} and
predictor-based NAS algorithms~\citep{alphax, bonas, bananas} 
have recently become popular.
We give a brief survey of performance prediction techniques in Section~\ref{sec:prelim}.
For a survey on NAS, see~\citep{nas-survey}.
The most widely used type of search space in prior work is the cell-based search 
space~\citep{zoph2018learning},
where the architecture search is over a relatively small directed acyclic 
graph representing an architecture.


A few recent works have compared different performance predictors on
popular cell-based search spaces for NAS.
Siems et al.~\citep{nasbench301} 
studied graph neural networks and tree-based methods,
and found that gradient-boosted trees and graph isomorphism networks performed the best. However, the
comparison was only on a single search space and dataset, 
and the explicit goal was to achieve maximum performance
given a training set of around 60\,000 architectures.
Another recent paper~\citep{ning2020surgery} studied various aspects of supernetwork training,
and separately compared four model-based methods: random forest, MLP, LSTM, 
and GATES~\citep{ning2020generic}.
However, the comparisons were again on a single search space and dataset 
and did not compare between multiple families of performance predictors.
Other papers have proposed new model-based predictors and compared the new 
predictors to other model-based baselines~\citep{seminas, alphax, bonas, bananas}.
Finally, a recent paper analyzed training heuristics to make weight-sharing more effective
at ranking architectures~\citep{yu2020train}.
To the best of our knowledge, no prior work has conducted comparisons across multiple families 
of performance predictors.

\section{Performance Prediction Methods for NAS} \label{sec:prelim}

In NAS, given a search space $\mathcal{A}$, the goal is to find
$a^*=\text{argmin}_{a\in\mathcal{A}} f(a)$, where $f$ denotes the validation error
of architecture $a$ after training on a fixed dataset for a fixed number of epochs $E$.
Since evaluating $f(a)$ typically takes hours (as it requires training a neural
network from scratch), many NAS algorithms make use of performance
predictors to speed up this process.
A \emph{performance predictor} $f'$ is defined generally as any function which predicts
the final accuracy or ranking of architectures, without fully training the architectures.
That is, evaluating $f'$ should take less time than evaluating $f$,
and $\{f'(a)\mid a\in\mathcal{A}\}$ should ideally have high correlation or rank correlation with
$\{f(a)\mid a\in\mathcal{A}\}$.

Each performance predictor is defined by two main routines: 
an \textbf{initialization} routine which performs general pre-computation,
and a \textbf{query} routine which performs the final architecture-specific computation: it
takes as input an architecture specification, and outputs its predicted accuracy.
For example, one of the simplest performance predictors is 
early stopping: for any $\textbf{query}(a)$, train $a$ for $E/2$ epochs instead
of $E$~\citep{zhou2020econas}. In this case, there is no general pre-computation, so initialization time is zero. On the other hand, the query time for each input architecture is high because it involves training the architecture for $E/2$ epochs. 
In fact, the runtime of the initialization and query routines varies substantially based on 
the type of predictor.
In the context of NAS algorithms, the initialization routine is typically performed once at
the start of the algorithm, and the query routine is typically performed many times 
throughout the NAS algorithm. Some performance predictors also make use of an \textbf{update} 
routine, when part of the computation from initialization needs to be updated without
running the full procedure again (for example, in a NAS algorithm, a model may be updated 
periodically based on newly trained architectures).
Now we give an overview of the main families of 
predictors. See Figure~\ref{fig:3d} (left) for a taxonomy of performance predictors.

\paragraph{Model-based (trainable) methods.}
The most common type of predictor, the model-based predictor, is based on
supervised learning. The initialization routine consists of fully training many architectures
(i.e., evaluating $f(a)$ for many architectures $a\in\mathcal{A}$) to build
a training set of datapoints $\{a, f(a)\}$. Then a model $f'$ is trained to predict $f(a)$ given
$a$.
While the initialization time for model-based predictors is very high, the query time
typically takes less than a second, which allows thousands of predictions to be made
throughout a NAS algorithm. The model is also updated regularly based on the new datapoints.
These predictors are typically used within 
BO frameworks~\citep{ma2019deep, bonas}, 
evolutionary frameworks~\citep{npenas}, or by themselves~\citep{wen2019neural}, 
to perform NAS. Popular choices for the model include tree-based methods 
(where the features are the adjacency matrix representation of the architectures)~\citep{luo2020neural, nasbench301},
graph neural networks~\citep{ma2019deep,bonas},
Gaussian processes~\citep{gpml, nasbowl},
and neural networks based on specialized encodings of the architecture~\citep{bananas, ning2020generic}.

\paragraph{Learning curve-based methods.}
Another family predicts the final performance of architectures using only a 
partially trained network, by extrapolating the learning curve. 
This is accomplished by fitting the partial learning curve to an ensemble of parametric
models~\citep{domhan2015speeding}, or by simply summing the training losses observed so far~\citep{ru2020revisiting}.
Early stopping as described earlier is also a learning curve-based method.
Learning curve methods do not require any initialization time, yet the query time
typically takes minutes or hours, which is orders of magnitude slower 
than the query time in model-based methods.
Learning curve-based methods can be used in conjunction with 
multi-fidelty algorithms, such as Hyperband or 
BOHB~\citep{hyperband, bohb, li2018system}.
%

\paragraph{Hybrid methods.}
Some predictors are hybrids between learning curve and model-based methods. 
These predictors train a model at initialization time to predict
$f(a)$ given both $a$ and a partial learning curve of $a$ as features.
Models in prior work include an SVR~\citep{baker2017accelerating},
or a Bayesian neural network~\citep{lcnet}.
Although the query time and initialization time are both high, 
hybrid predictors tend to have strong performance.

\paragraph{Zero-cost methods.}
Another class of predictors have no initialization time and very short query times (so-called
``zero-cost'' methods). These predictors compute statistics from just a 
single forward/backward propagation pass for a single minibatch of data,
by computing the correlation of activations within a network~\citep{mellor2020neural},
or by adapting saliency metrics proposed in pruning-at-initialization literatures~\citep{lee2018snip,abdelfattah2021zerocost}. 
Similar to learning curve-based methods,
since the only computation is specific to each architecture, the initialization time is zero.
Zero-cost methods have recently been used to warm start NAS algorithms~\citep{abdelfattah2021zerocost}.

\paragraph{Weight sharing methods.}

Weight sharing~\citep{enas} is a popular approach to substantially speed up NAS, 
especially in conjunction with a one-shot algorithm~\citep{darts, gdas}.
In this approach, all architectures in the search space are combined to form a single over-parameterized supernetwork. 
By training the weights of the supernetwork, all architectures in the search space can 
be evaluated quickly using this set of weights. To this end, the supernetwork 
can be used as a performance predictor. 
This results in NAS algorithms~\citep{darts, darts+} which are significantly faster than sequential NAS algorithms, such
as evolution or Bayesian optimization.
Recent work has shown that although the shared weights are sometimes
not effective at ranking 
architectures~\citep{sciuto2019evaluating, zela2020bench, zhang2020deeper}, one-shot NAS techniques
using shared weights still achieve strong performance~\citep{zela2020understanding, li2020geometry}.

\paragraph{Tradeoff between intialization and query time.}
The main families mentioned above all have different initialization and query times.
The tradeoffs between initialization time, query time, and performance depend on a few factors such as the type of NAS algorithm and its total runtime budget, and different settings are needed in different situations. For example, if there are many architectures whose performance we want to estimate, then we should have a low query time, and if we have a high total runtime budget, then we can afford a high initialization time. We may also change our runtime budget throughout the run of a single NAS algorithm. For example, at the start of a NAS algorithm, we may want to have coarse estimates of a large number of architectures (low initialization time, low query time such as zero-cost predictors). As the NAS algorithm progresses, it is more desirable to receive higher-fidelity predictions on a smaller set of architectures (model-based or hybrid predictors). The exact budgets depend on the type of NAS algorithm.

\paragraph{Choice of performance predictors.}


We analyze 31 performance predictors defined in prior work:
BANANAS~\citep{bananas},
Bayesian Linear Regression~\citep{bishop2006pattern},
BOHAMIANN~\citep{springenberg2016bayesian},
BONAS~\citep{bonas},
DNGO~\citep{snoek2015scalable},
Early Stopping with Val.\ Acc.\ (e.g.~\citep{zhou2020econas, hyperband, bohb, zoph2018learning})
Early Stopping with Val.\ Loss.~\citep{ru2020revisiting},
Fisher~\citep{abdelfattah2021zerocost},
Gaussian Process (GP)~\citep{rasmussen2003gaussian},
GCN~\citep{zhang2020neural},
Grad Norm~\citep{abdelfattah2021zerocost},
Grasp~\citep{wang2019picking},
Jacobian Covariance~\citep{mellor2020neural},
LCE~\citep{domhan2015speeding},
LCE-m~\citep{lcnet},
LcSVR~\citep{baker2017accelerating},
LGBoost/GBDT~\citep{luo2020neural},
MLP~\citep{bananas},
NAO~\citep{luo2018neural},
NGBoost~\citep{nasbench301},
OneShot~\citep{zela2020understanding},
Random Forest (RF)~\citep{nasbench301},
Random Search with Weight Sharing (RSWS)~\citep{randomnas},
SemiNAS~\citep{seminas},
SNIP~\citep{lee2018snip},
SoTL~\citep{ru2020revisiting},
SoTL-E~\citep{ru2020revisiting},
Sparse GP~\citep{bauer2016understanding}, 
SynFlow~\cite{tanaka2020pruning},
Variational Sparse GP~\citep{titsias2009variational},
and XGBoost~\citep{nasbench301}.
For any method that did not have an architecture encoding already defined 
(such as the tree-based
methods, GP-based methods, and Bayesian Linear Regression),
we use the standard adjacency matrix encoding, which consists of the adjacency matrix
of the architecture along with a one-hot list of the 
operations~\citep{nasbench, white2020study}.
By open-sourcing our code, we encourage implementing more (existing and
future) performance predictors which can then be compared to the 31 which
we focus on in this work.
In Section~\ref{subsec:descriptions}, we give descriptions 
and detailed implementation details for each performance predictor.
In Section \ref{app:reproducibility}, we give a table that describes for which predictors we were able to reproduce published results, and for which predictors it is not possible (e.g., since some predictors were released before the creation of NAS benchmarks).
\section{Experiments} \label{sec:experiments}

We now discuss our experimental setup and results.
We discuss reproducibility in Sections \ref{app:nas_checklist} and \ref{app:reproducibility}, 
and our code (based on the NASLib library~\citep{ruchte2020naslib}) is available at \url{https://github.com/automl/naslib}.
We split up our experiments into two categories: 
evaluating the performance of each predictor with respect to 
various correlation metrics (Section~\ref{sec:evaluation}), 
and evaluating the ability of each predictor to speed up 
predictor-based NAS algorithms (Section~\ref{sec:nas}).
We start by describing the four NAS benchmarks used in our experiments. 


\begin{figure}
    \centering
    \begin{tabular}{c@{}c@{}}
      \multirow{-9}{*}{\includegraphics[width=.49\columnwidth]{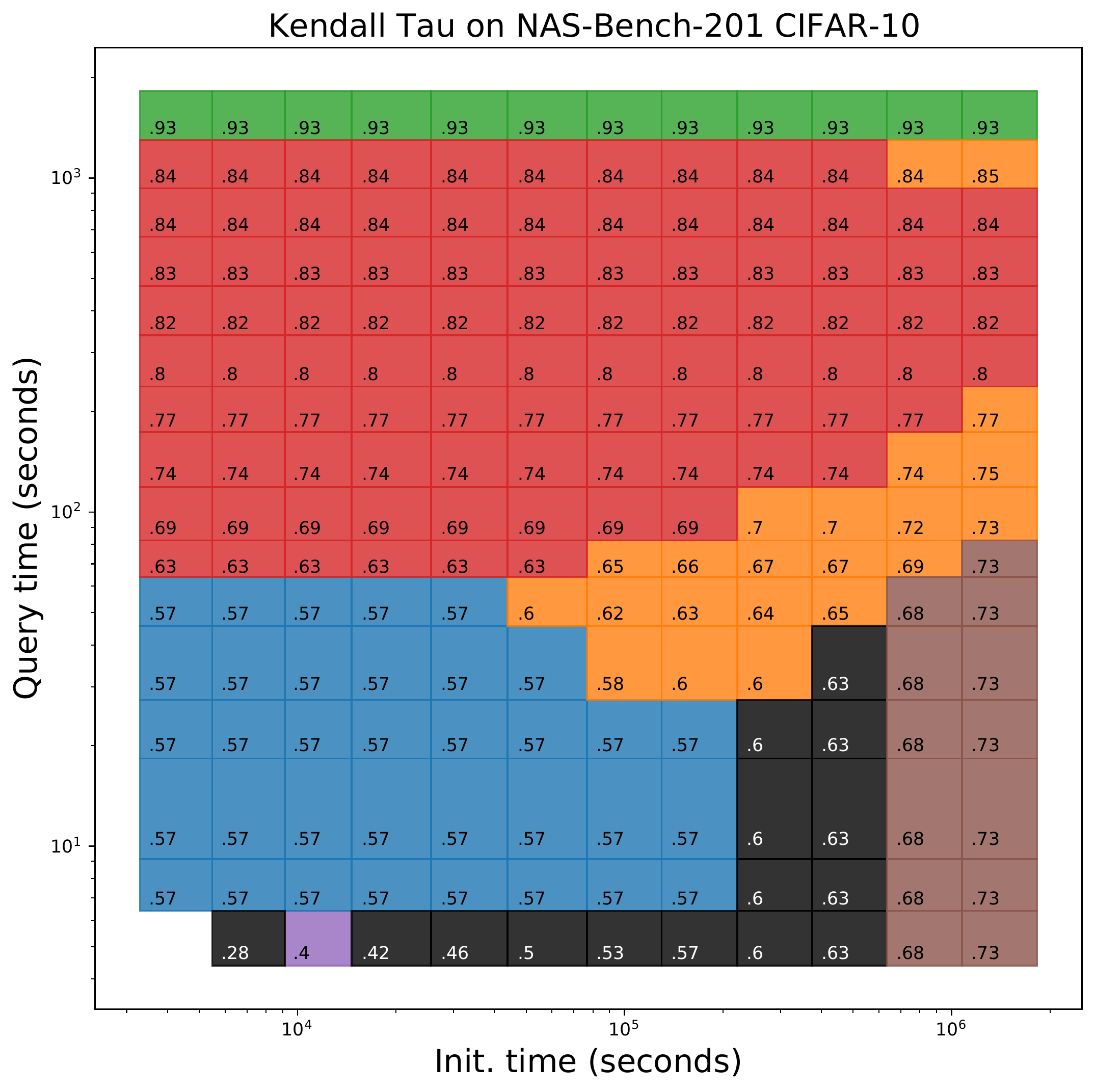}}
        &
      \begin{tabular}{l}
        \raisebox{0.0\height}{\includegraphics[width=.24\columnwidth]{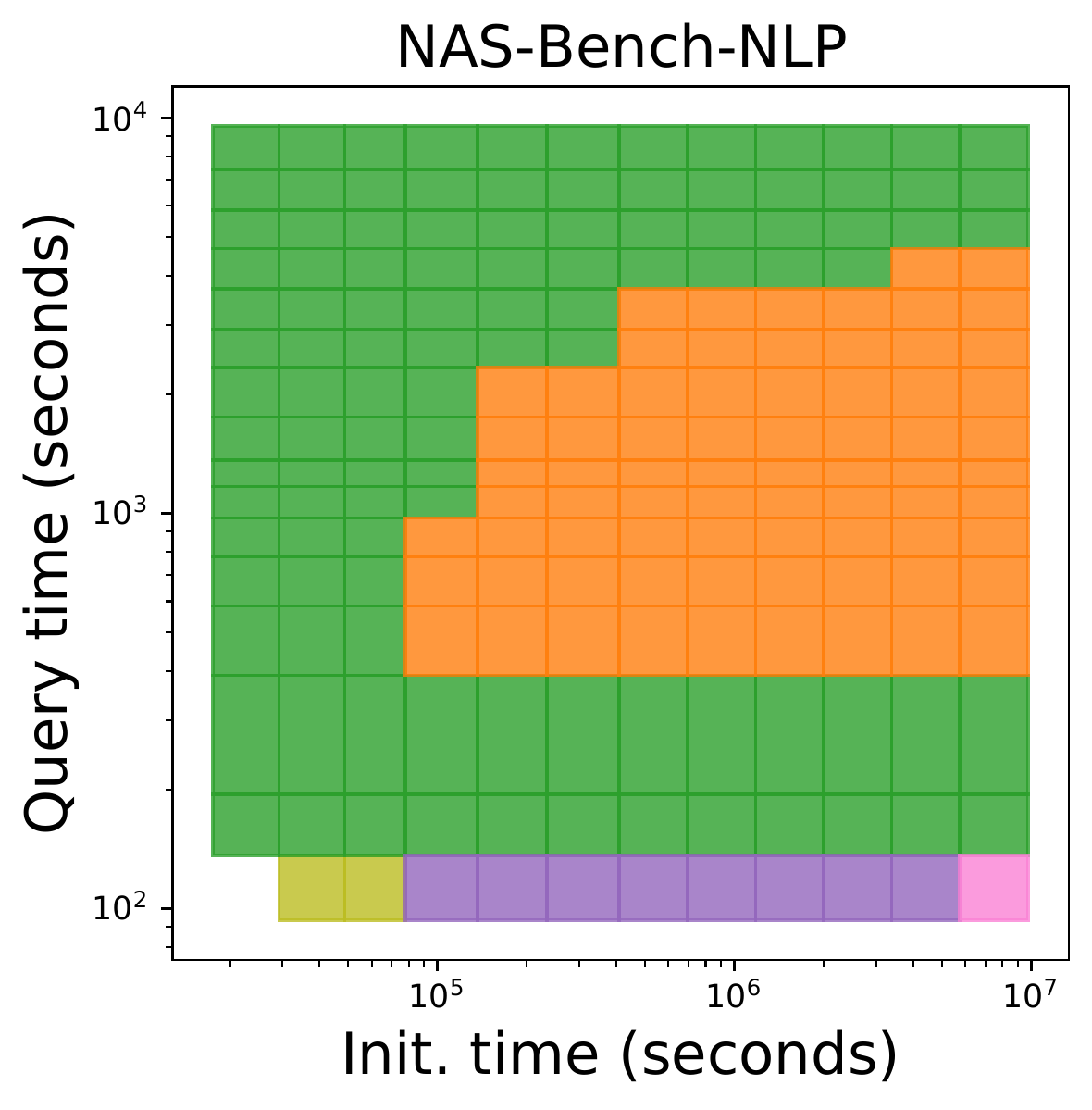}}
        \raisebox{0.0\height}{\includegraphics[width=.25\columnwidth]{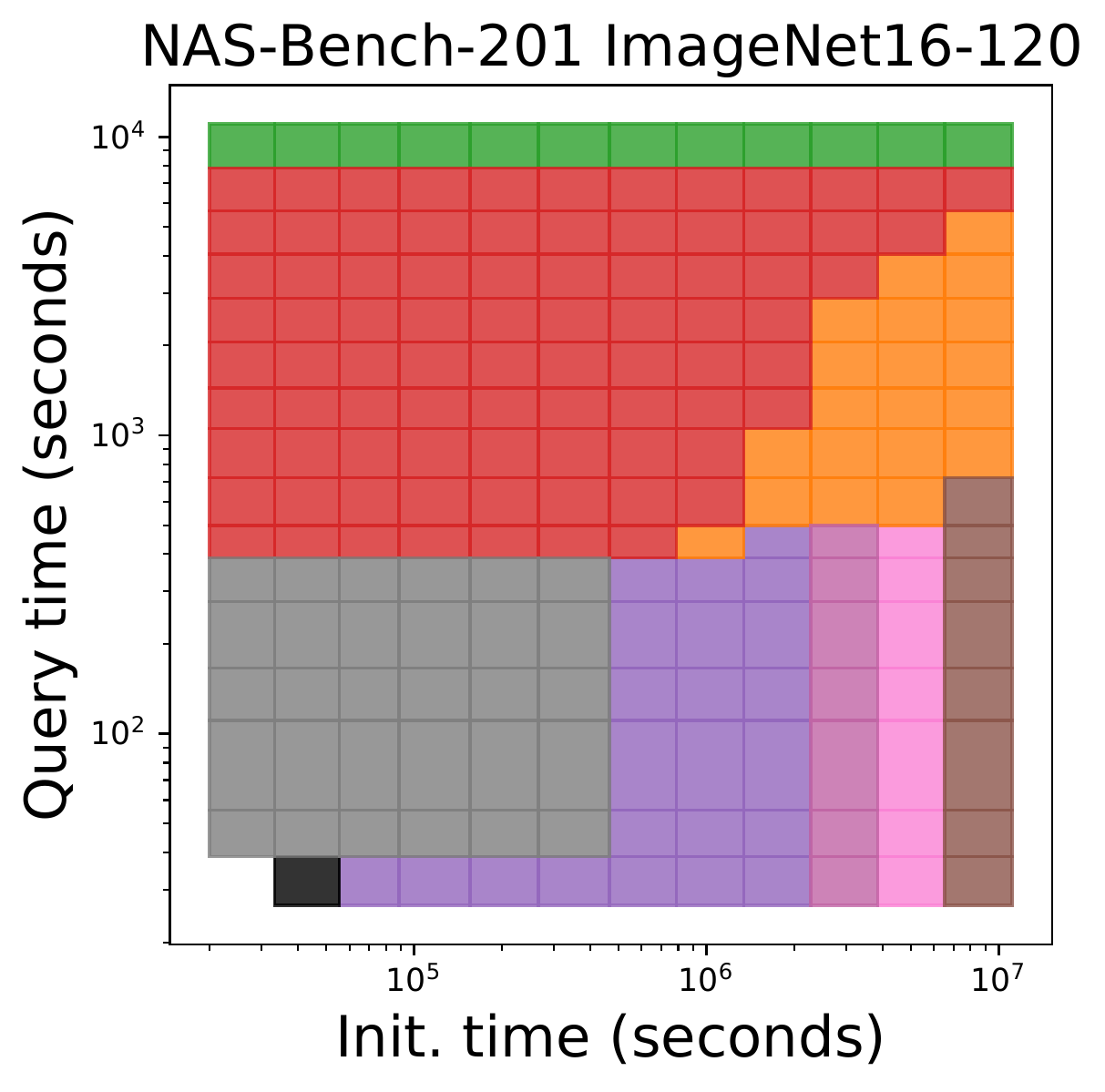}}
        \tabularnewline
        \raisebox{0.0\height}{\includegraphics[width=.24\columnwidth]{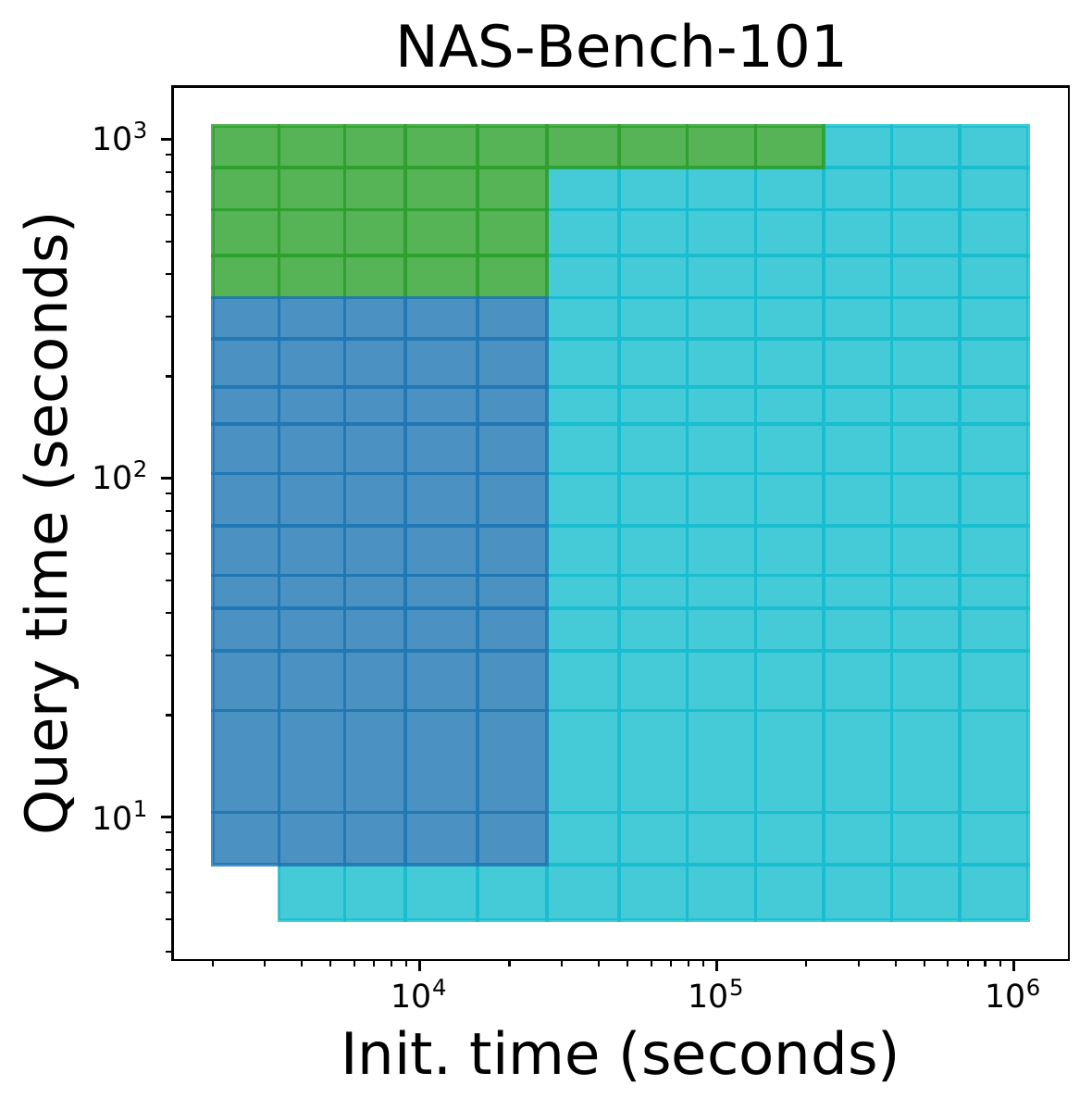}}
        \raisebox{0.0\height}{\includegraphics[width=.24\columnwidth]{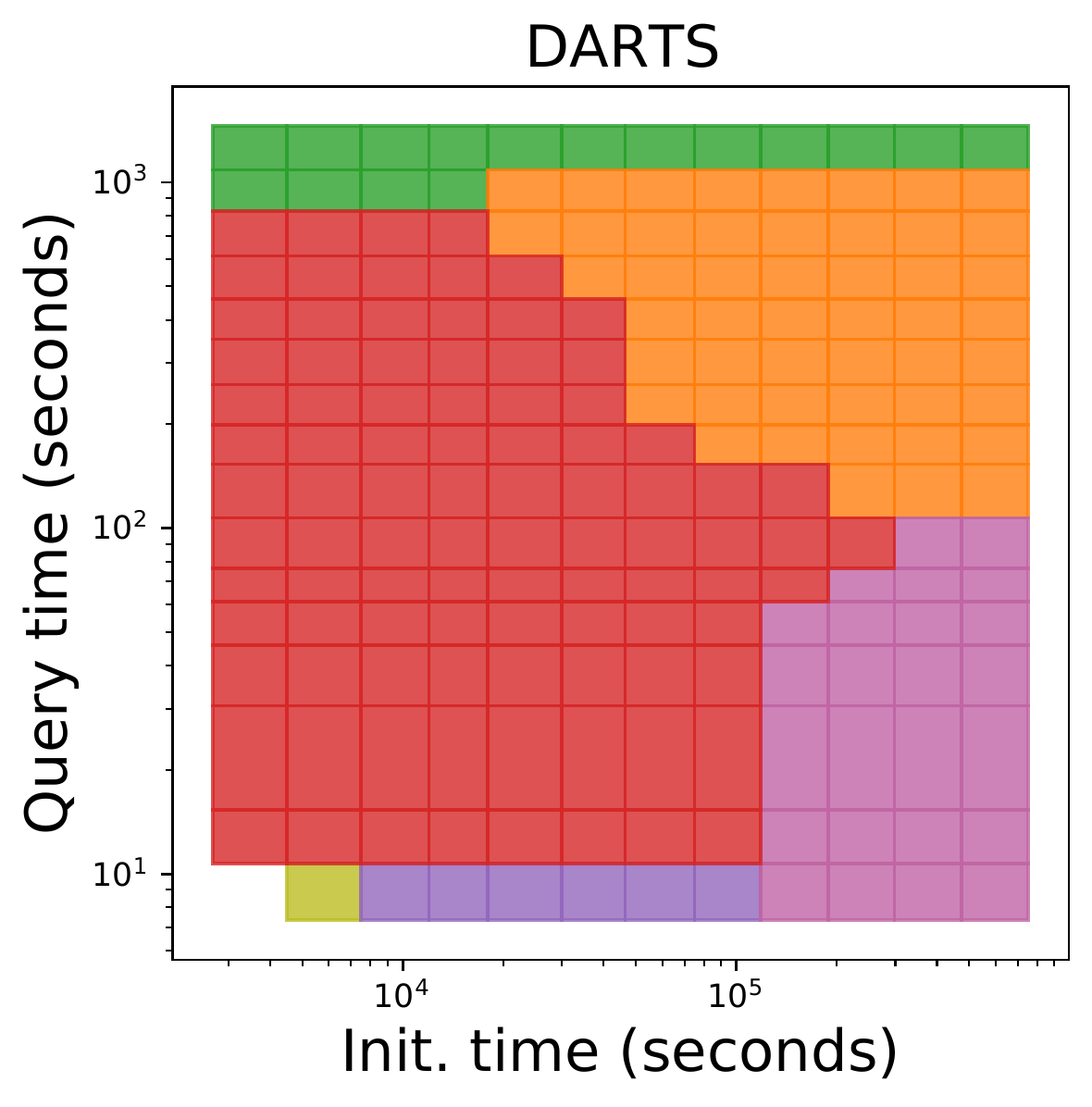}} 
        \tabularnewline
      \end{tabular} \tabularnewline
    \end{tabular}
    \includegraphics[width=.8\columnwidth]{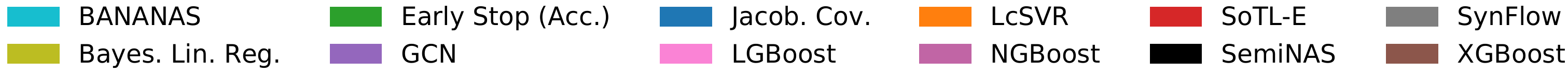}
    \caption{The performance predictors with the highest Kendall Tau values
for all initialization time and query time budgets on NAS-Bench-201, NAS-Bench-101, NAS-Bench-NLP and DARTS.
For example, on NAS-Bench-201 CIFAR-10 (left) with an initialization time of $10^6$ seconds and query time of $10$ seconds, XGBoost achieves a Kendall Tau value of $.73$ which is the highest value out of the 31 predictors that we tested at that budget. 
}
    \label{fig:search_spaces}
\end{figure}

\paragraph{NAS benchmark datasets.}
NAS-Bench-101~\cite{nasbench} consists of over 
$423\,000$ unique neural architectures 
with precomputed training, validation, and test accuracies after training 
for 4, 12, 36, and 108 epochs on CIFAR-10 \citep{nasbench}.
The cell-based search space consists of five nodes which can take on any directed acyclic graph (DAG) structure, and each node can be one of three operations.
Since learning curve information is only available at four epochs, 
it is not possible to run most learning curve extrapolation methods on 
NAS-Bench-101.
%
NAS-Bench-201~\cite{nasbench201} consists of $15\,625$  architectures
(out of which $6\,466$ are unique after removing isomorphisms \citep{nasbench201}). 
Each architecture has full learning curve information for training, 
validation, and test losses/accuracies for 200 epochs on 
CIFAR-10~\citep{CIFAR10}, CIFAR-100, and ImageNet-16-120~\citep{DownsampledImageNet}.
The search space consists of a cell which is a complete DAG
with 4 nodes.
Each edge can take one of five different operations.
%
The DARTS search space~\citep{darts} is significantly
larger with roughly $10^{18}$ architectures. 
The search space consists of two cells,
each with seven nodes. 
The first two nodes are inputs from previous layers, and the intermediate four nodes 
can take on any DAG structure such that each node has two incident edges. 
The last node is the output node.
Each edge can take one of eight operations.
In our experiments, we make use of the training data from 
NAS-Bench-301~\citep{nasbench301}, which consists of $23\,000$ architectures drawn 
uniformly at random and trained on CIFAR-10 for 100 epochs.
%
Finally, the NAS-Bench-NLP search space~\citep{nasbenchnlp} is even larger, at $10^{53}$
LSTM-like cells, each with at most 25 nodes in any DAG structure.
Each cell can take one of seven operations.
In our experiments, we use the NAS-Bench-NLP dataset, which consists
of $14\,000$ architectures drawn uniformly at random and trained on 
Penn Tree Bank~\citep{penntreebank} for 50 epochs.

\paragraph{Hyperparameter tuning.}
Although we used the code directly from the original repositories
(sometimes making changes when necessary to adapt to NAS-Bench search spaces), 
the predictors had significantly different levels of hyperparameter tuning.
For example, some of the predictors had undergone heavy hyperparameter
tuning on the DARTS search space (used in NAS-Bench-301), while other predictors (particularly
those from 2017 or earlier) had never been run on cell-based
search spaces. 
Furthermore, most predictor-based NAS algorithms can utilize
cross-validation to tune the predictor periodically throughout the
NAS algorithm. This is because the bottleneck for predictor-based 
NAS algorithms is typically the training of architectures, not fitting
the predictor~\citep{snoek2015scalable, lindauer2017smac, bohb}.
Therefore, it is fairer and also more informative to compare performance 
predictors which have had the same level of hyperparameter tuning through 
cross-validation. For each search space, we run random search on each 
performance predictor for 5000 iterations, with a maximum total runtime of 15 minutes.
The final evaluation uses a separate test set.
The hyperparameter value ranges for each predictor can be found in
Section~\ref{subsec:hpo}.

\subsection{Performance Predictor Evaluation}\label{sec:evaluation}
We evaluate each predictor based on three axes of comparison:
initialization time, query time, and performance.
We measured performance with respect to several different metrics:
Pearson correlation 
and three different rank correlation metrics (Spearman, 
Kendall Tau, and sparse Kendall Tau~\cite{yu2020train, nasbench301}).
The experimental setup is as follows:
the predictors are tested with 11 different initialization time budgets
and 14 different query time budgets, leading to a total of 154 settings.
On NAS-Bench-201 CIFAR-10, the 11 initialization time budgets are spaced 
logarithmically from 1 second to $1.8\times 10^{7}$ seconds on a 1080 Ti GPU
(which corresponds to training 1000 random architectures on average) 
which is consistent with experiments conducted in prior 
work~\citep{alphax, bananas, seminas}.
For other search spaces, these times are adjusted based on the average
time to train 1000 architectures.
The 14 query time budgets are spaced logarithmically from 1 second to $1.8\times 10^4$ seconds
(which corresponds to training an architecture for 199 epochs).
These times are adjusted for other search spaces based on the training time and different number of epochs. 
Once the predictor is initialized,
we draw a test set of 200 architectures uniformly at random from the search
space. 
For each architecture in the test set, the predictor uses the specified
query time budget to make a prediction.
We then evaluate the quality of the predictions using the metrics described above.
We average the results over 100 trials for each (initialization time, query time) pair.

\paragraph{Results and discussion.}

\begin{figure}
\centering
\includegraphics[width=.27\columnwidth]{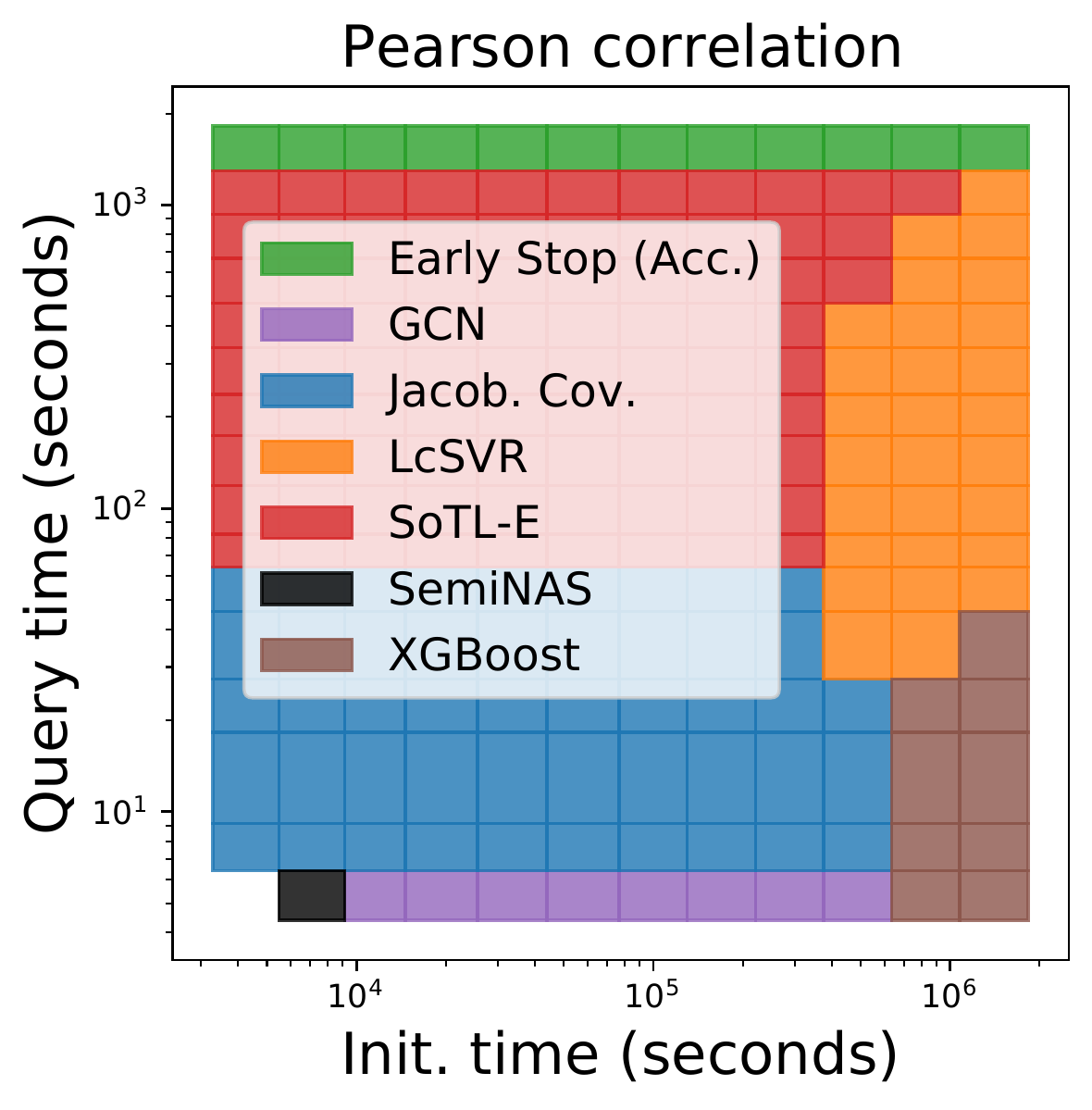}
\includegraphics[width=.27\columnwidth]{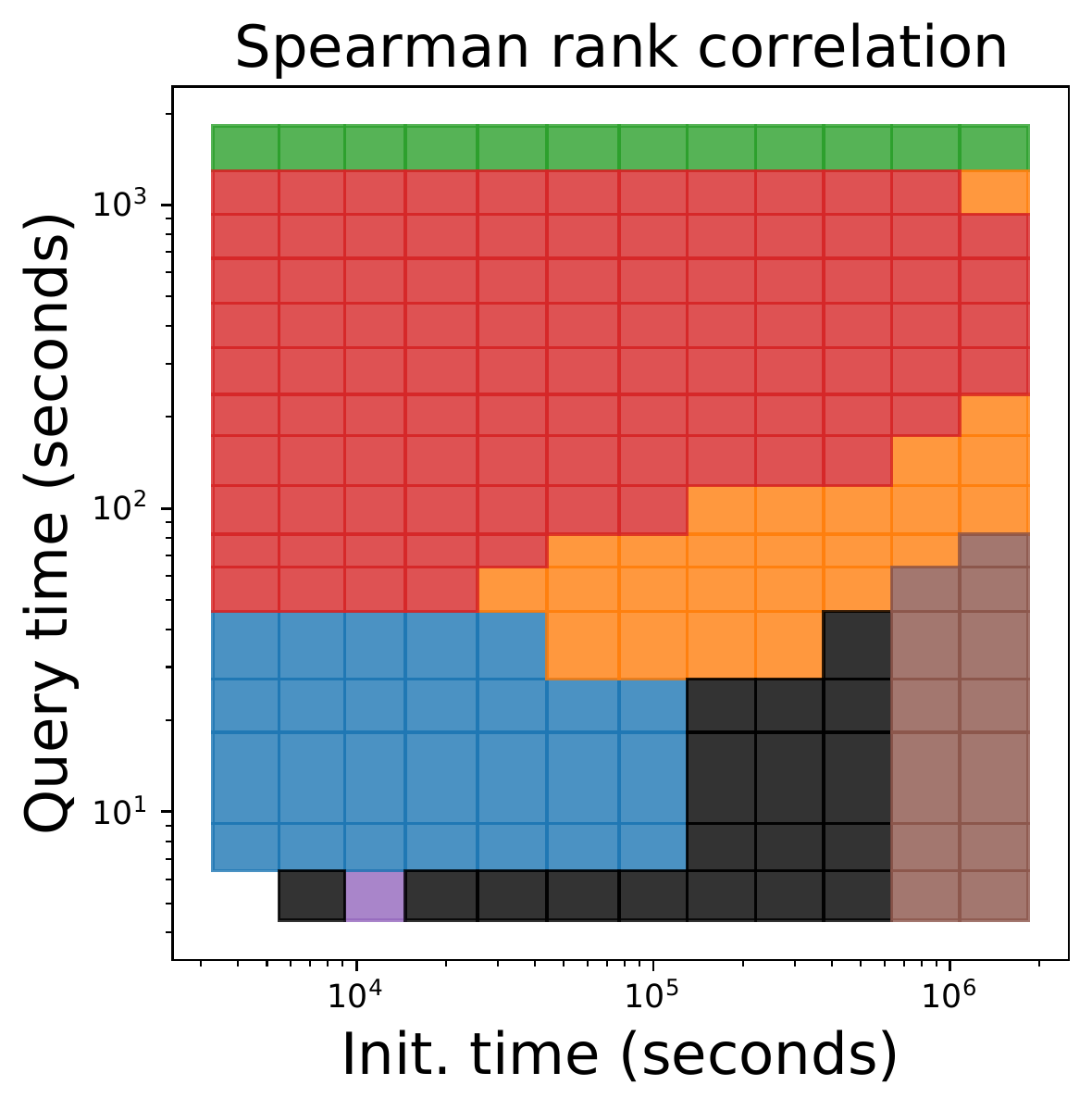}
\raisebox{0.0\height}{\includegraphics[width=.32\columnwidth]{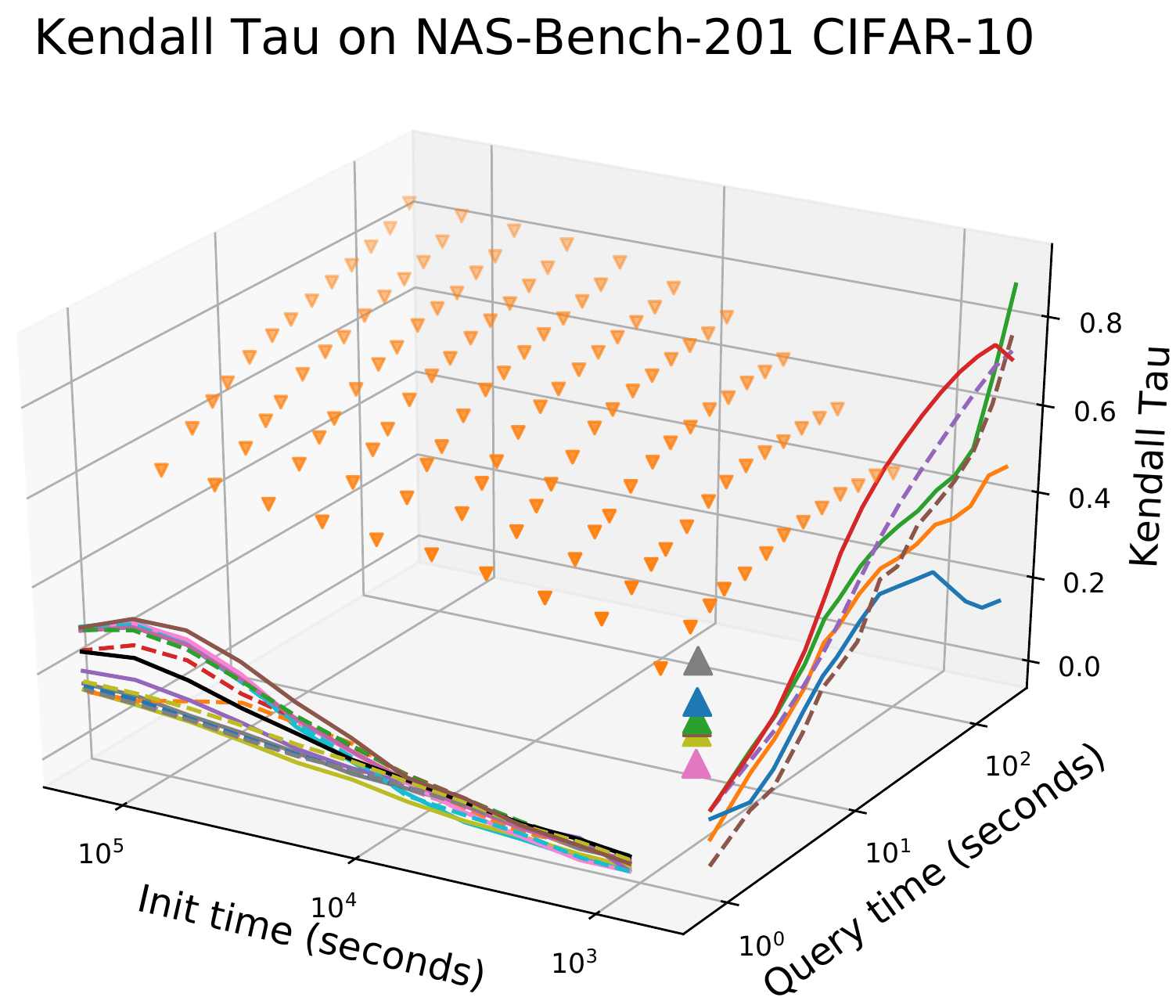}}
\raisebox{0.0\height}{\includegraphics[width=.94\columnwidth]{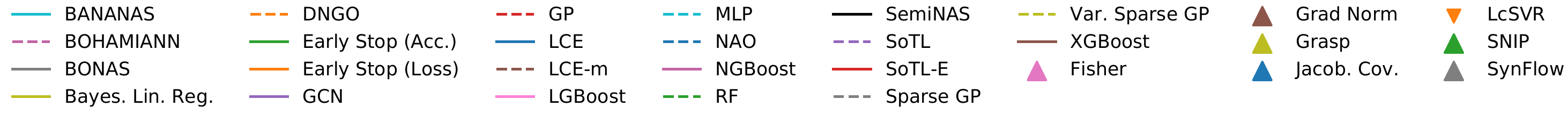}}
\caption{
The best predictors on NAS-Bench-201 CIFAR-10 with respect to Pearson 
(left) and Spearman (middle).
Kendall Tau values from a mutation-based training and test set (right).
}
\label{fig:mutate}
\end{figure}

Figure~\ref{fig:3d} shows a full three-dimensional plot for NAS-Bench-201 on CIFAR-10 
over initialization time, query time, and Kendall Tau rank correlation.
Of the 31 predictors we tested, we found that just seven of them are
Pareto-optimal with respect to Kendall Tau, initialization time, and query time.
That is, only seven algorithms have the highest Kendall Tau value for at least
one of the 154 query time/initialization time budgets on NAS-Bench-201 CIFAR-10.
This can be seen more clearly in Figure~\ref{fig:search_spaces} (left), 
which is a view from above Figure~\ref{fig:3d}: each lattice point displays the
predictor with the highest Kendall Tau value for the corresponding budget.
In Figure~\ref{fig:search_spaces} (right), 
we plot the Pareto-optimal predictors for five different
dataset/search space combinations.
In Section~\ref{subsec:experiments}, we give the full 3D plots and report the variance
across trials for each method.
In Figure~\ref{fig:mutate} (left), we also plot the Pearson and Spearman correlation
coefficients for NAS-Bench-201 CIFAR-10. The trends between these measures are largely the 
same, although we see that SemiNAS performs better on the rank-based metrics. 
For the rest of this section, we focus on the popular Kendall Tau metric, 
giving the full results for the other metrics in Section~\ref{subsec:experiments}.

We see similar trends across DARTS and the two NAS-Bench-201 datasets.
NAS-Bench-NLP also has fairly similar trends, although early stopping performs 
comparatively stronger.
NAS-Bench-101 is different from the other search spaces both in terms of the
topology and the benchmark itself, which we discuss later in this section.

In the low initialization time, low query time region, Jacobian covariance
or SynFlow perform well
across NAS-Bench-101 and NAS-Bench-201. However, none of the six zero-cost
methods perform well on the larger DARTS search space.
Weight sharing (which also has low initialization and low query time, as seen in Figure~\ref{fig:3d}),
did not yield high Kendall Tau values for these search spaces, either, which is consistent with recent 
work~\citep{sciuto2019evaluating, zela2020bench, zhang2020deeper}. However, rank 
correlation is not as crucial to one-shot NAS algorithms as it is for black box
predictor-based methods, as demonstrated by prior 
one-shot NAS methods that do perform well~\citep{li2020geometry, zela2020understanding,darts,darts+,gdas}.
In the low initialization time, high query time region, sum of training losses 
(SoTL-E) consistently 
performed the best, outperforming other learning-curve based methods. 

The high initialization time, low query time region
(especially the bottom row of the plots, corresponding to a query 
time of 1 second) is by far the most competitive region in the recent NAS literature. 
Sixteen of the 31 predictors had query times under one second, 
because many NAS algorithms are designed to initialize (and continually
update) performance predictors that are used to quickly query thousands of candidate 
architectures. GCN and SemiNAS, the specialized GCN/semi-supervised methods, 
perform especially well in the first half of this critical region, when the initialization
time is relatively low. However, boosted tree methods actually performed best in the second 
half of the critical region where the initialization time is high,
which is consistent with prior work~\citep{luo2020neural, nasbench301}.
Recall that for model-based methods, the initialization time corresponds to training
architectures to be used as training data for the performance predictor.
Therefore, our results suggest that techniques which can extract better latent features of the architectures can make up for a 
small training dataset, but methods based purely on performance data work better when there is enough such data.

Perhaps the most interesting finding is that on NAS-Bench-101/201,
SynFlow and Jacobian covariance, which take three seconds each to compute, 
both outperform all model-based methods even after \emph{30 hours} of initialization.
Put another way, NAS algorithms that make use of model-based predictors may be able to see
substantial improvements by using Jacobian covariance instead of a model-based predictor 
in the early iterations.

\begin{figure}
\centering
\includegraphics[width=.24\columnwidth]{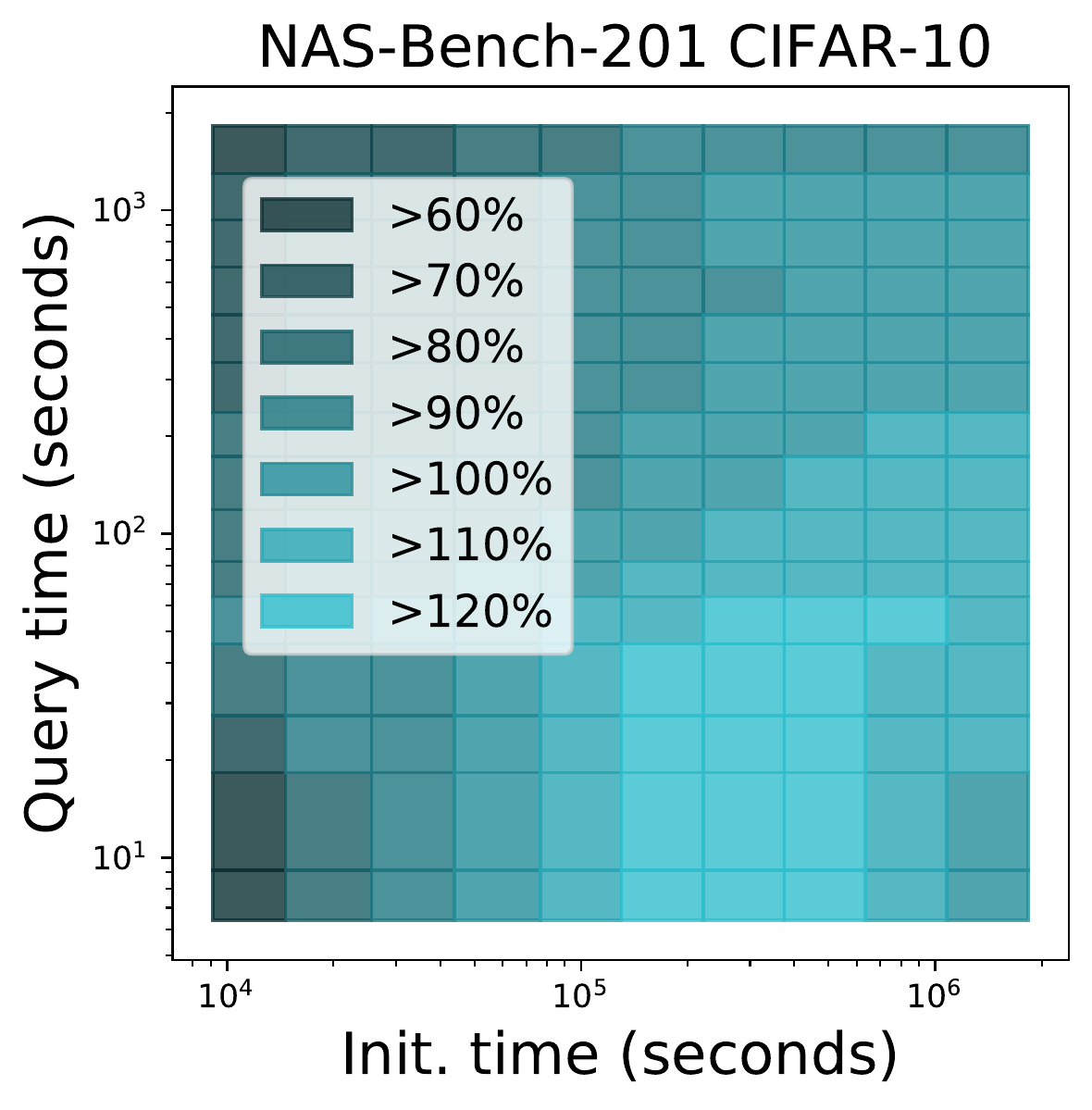}
\includegraphics[width=.24\columnwidth]{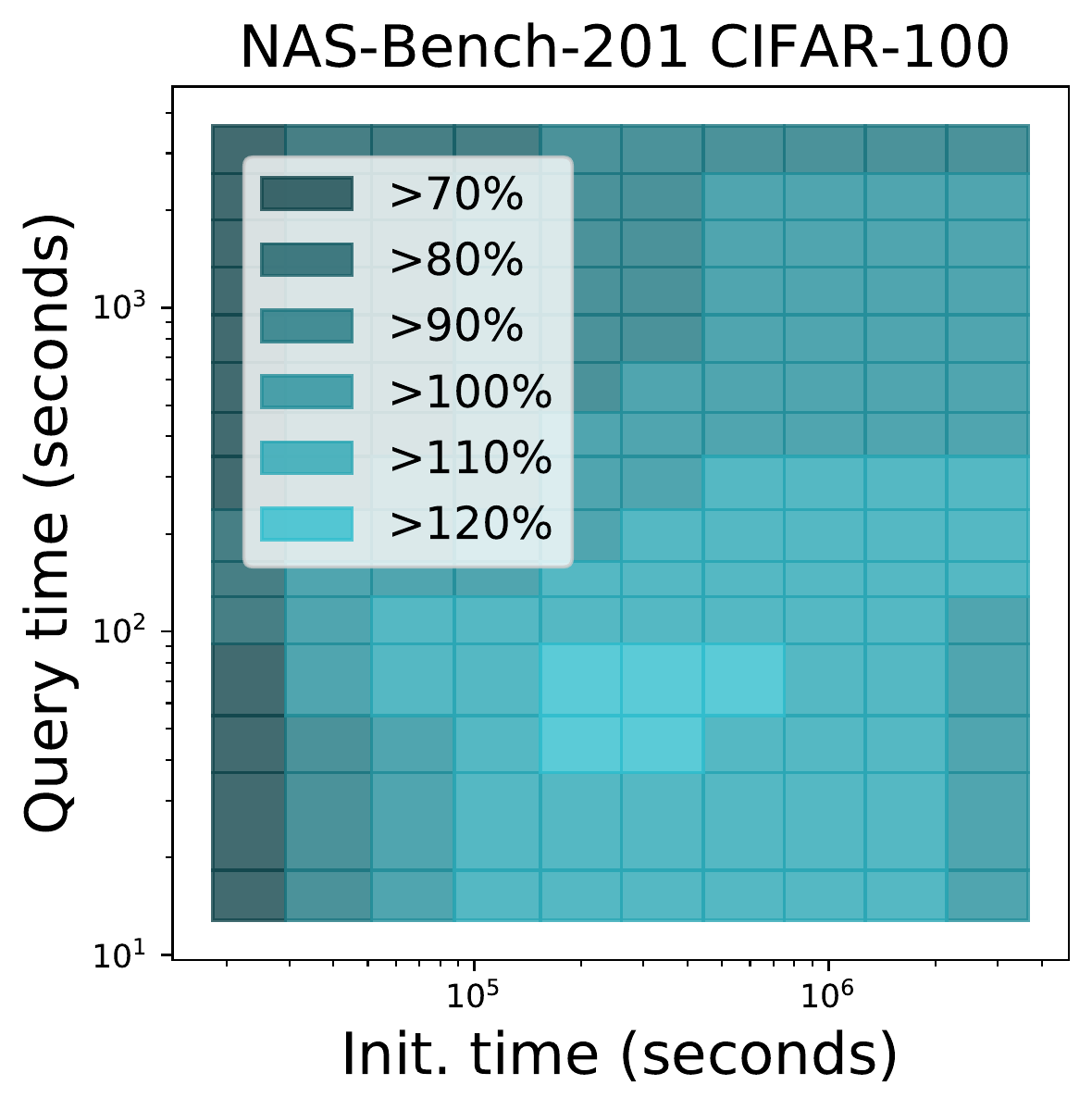}
\includegraphics[width=.24\columnwidth]{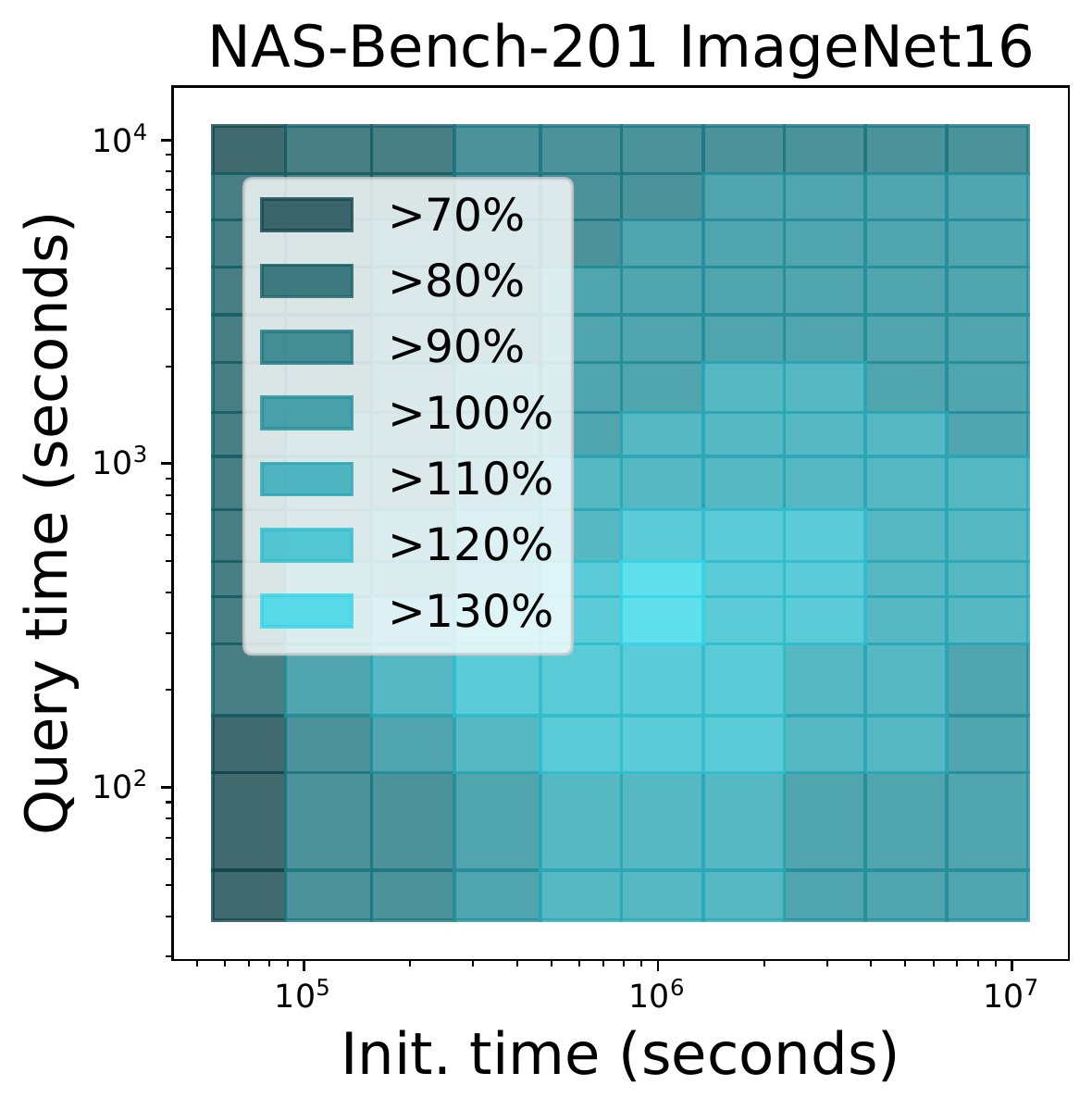}
\includegraphics[width=.24\columnwidth]{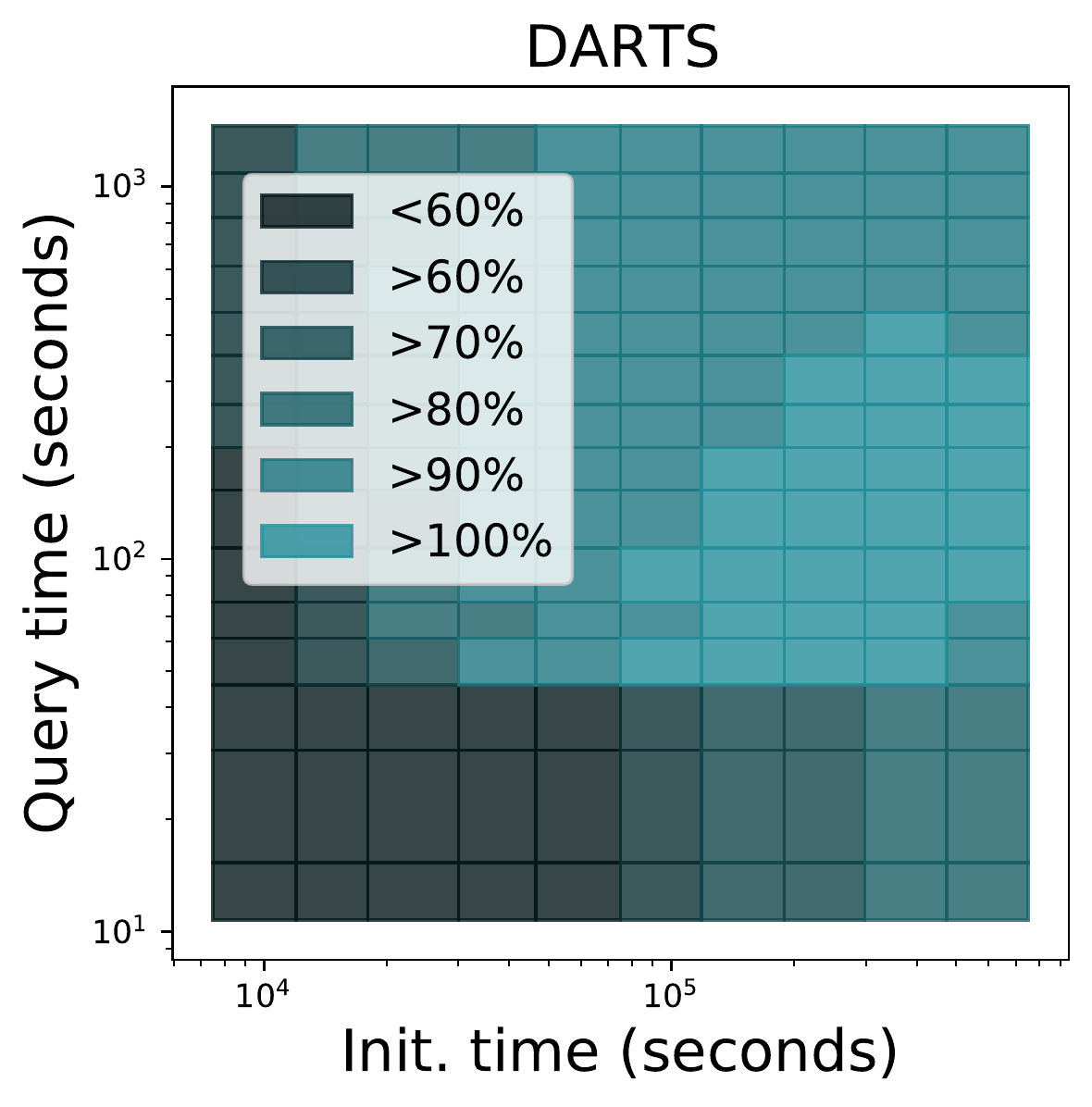}
\caption{Percentage of OMNI's Kendall Tau value compared to 
the next-best predictors for each budget constraint.}
\label{fig:omni}
\end{figure}

\paragraph{The Omnipotent Predictor.}
One conclusion from Figure~\ref{fig:search_spaces} is that different types
of predictors are specialized for specific initialization time and query time
constraints. A natural follow-up question is whether different families
are complementary and can be combined to achieve stronger performance.
In this section, we run a proof-of-concept to answer this question.
We combine the best-performing predictors from three different families in a simple way:
the best learning curve method (SoTL-E), and the best
zero-cost method (Jacobian covariance), are used as additional input 
features for a model-based predictor (we separately test SemiNAS and NGBoost).
We call this method OMNI, the \underline{omni}potent predictor.
We give results in Figure~\ref{fig:omni}
and pseudo-code as well as additional experiments in Section~\ref{subsec:omni}.
In contrast to all other predictors, the performance of OMNI
is strong across almost all budget constraints and search spaces.
In some settings, OMNI achieves a Kendall Tau value 30\% higher than the 
next-best predictors.

The success of OMNI verifies that the information learned by different families of predictors
are complementary: the information learned by extrapolating a learning curve, 
by computing a zero-cost proxy, and by encoding the architecture, all improve
performance. We further confirm this by running an ablation study for OMNI in 
Section~\ref{subsec:omni}.
We can hypothesize that each predictor type measures distinct quantities: SOTL-E measures the training speed, zero-cost predictors measure the covariance between activations on different datapoints, and model-based predictors simply learn patterns between the architecture encodings and the validation accuracies.
Finally, while we showed a proof-of-concept, 
there are several promising
areas for future work such as creating ensembles of the model-based approaches,
combining zero-cost methods with 
model-based methods in more sophisticated ways, and giving a full quantification of the correlation among different families of predictors.

\paragraph{NAS-Bench-101: a more complex search space.}
In Figure~\ref{fig:search_spaces}, the plot for NAS-Bench-101 looks significantly
different than the plots for the other search spaces, for two reasons.
The first reason is a technical one: the NAS-Bench-101 API only gives the validation
accuracy at four epochs, and does not give the training loss for any epochs.
Therefore, we could not implement SoTL or any learning curve extrapolation method.
However, all sixteen of the model-based predictors were implemented on NAS-Bench-101. In this case,
BANANAS significantly outperformed the next-best predictors (GCN and SemiNAS)
across every initialization time.
One explanation is due to the complexity of the NAS-Bench-101 search space:
while all NAS-Bench-201 architectures have the same graph topology 
and DARTS architectures' nodes have exactly two incoming edges,
the NAS-Bench-101 search space is much more diverse with architectures 
ranging from a single node and no edges, to five nodes with nine 
connecting edges. In fact, the architecture encoding used in BANANAS,
the path encoding, was designed specifically to deal with the complexity of the
NAS-Bench-101 search space (replacing the standard adjacency matrix encoding).
To test this explanation, in Appendix~\ref{app:experiments}
we run several of the simpler tree-based and GP-based 
predictors using the path encoding, and we see that these methods now surpass BANANAS 
in performance.

\paragraph{A mutation-based test set.}
The results from Figure~\ref{fig:search_spaces} used a test set drawn uniformly at random
from the search space (and the training set used by model-based predictors was also drawn
uniformly at random). However, neighborhood-based NAS algorithms such as local search,
regularized evolution, and some versions of Bayesian optimization consider architectures
which are local perturbations of the architectures encountered so far. Therefore, the
predictors used in these NAS algorithms must be able to distinguish architectures
which are local mutations of a small set of seed architectures.

We run an experiment in which the test set is created by mutating architectures from an
initial set of seed architectures. Specifically, we draw a set of 50 random architectures
and choose the five with the highest validation accuracy as seed architectures.
Then we create a set of 200 test architectures by randomly mutating up to three attributes
of the seed architectures. Therefore, all architectures in the test set are at most
an edit distance of three from a seed architecture, where two architectures are a single edit distance away if they differ by one operation or edge.

We create the training set by randomly choosing architectures
from the test set and mutating one random attribute.
As in all of our experiments, we ensure that the training set and test
set are disjoint. In Figure~\ref{fig:mutate} (right), 
we plot the correlation results for NAS-Bench-201 CIFAR-10. While the zero-cost and learning curve-based approaches have similar
performance, the model-based approaches have significantly worse performance compared to 
the uniform random setting. This is because the average edit distance between 
architectures in the test set is low, making it significantly harder for model-based
predictors to distinguish the performance of these architectures, even when using a training
set that is based on mutations of the test set.
In fact, interestingly, the performance of many model-based approaches starts to perform worse after $10^6$ seconds. SemiNAS in particular performs much worse in this setting, 
and boosted trees have comparatively stronger performance
in this setting.


\subsection{Predictor-Based NAS Experiments}\label{sec:nas}

Now we evaluate the ability of each model-based performance predictor to 
speed up 
NAS. We use two popular predictor-based NAS methods: the predictor-guided evolution
framework~\citep{npenas, sun2020new}, and
the Bayesian optimization + predictor framework~\citep{nasbot, ma2019deep, bonas}.
The predictor-guided evolution framework is an iterative procedure in which 
the best architectures in the current population are mutated to create a set of candidate
architectures. A predictor (trained on the entire population) chooses $k$ 
architectures which are then evaluated. 
In our experiments, the candidate pool is created by mutating the
top five architectures 40 times each, and we set $k=20$.
For each predictor, we run predictor-guided evolution for 25 iterations and
average the results over 100 trials.
The BO + predictor framework is similar to the evolution framework,
but an ensemble of three performance predictors are used so that uncertainty
estimates for each prediction can be computed.
In each iteration, the candidate architectures whose predictions maximize 
an acquisition function are then evaluated. 
Similar to prior work, we use independent Thompson sampling~\citep{bananas},
as the acquisition function, and an ensemble is created by using a different
ordering of the training set and different random weight initializations
(if applicable) of the same predictor. In each iteration, the top 20 architectures are chosen
from a randomly sampled pool of 200 architectures. 

\begin{figure}
    \centering
    \begin{tabular}{c@{}c@{}}
      \begin{tabular}{l}
        \raisebox{0.0\height}{\includegraphics[width=.39\columnwidth]{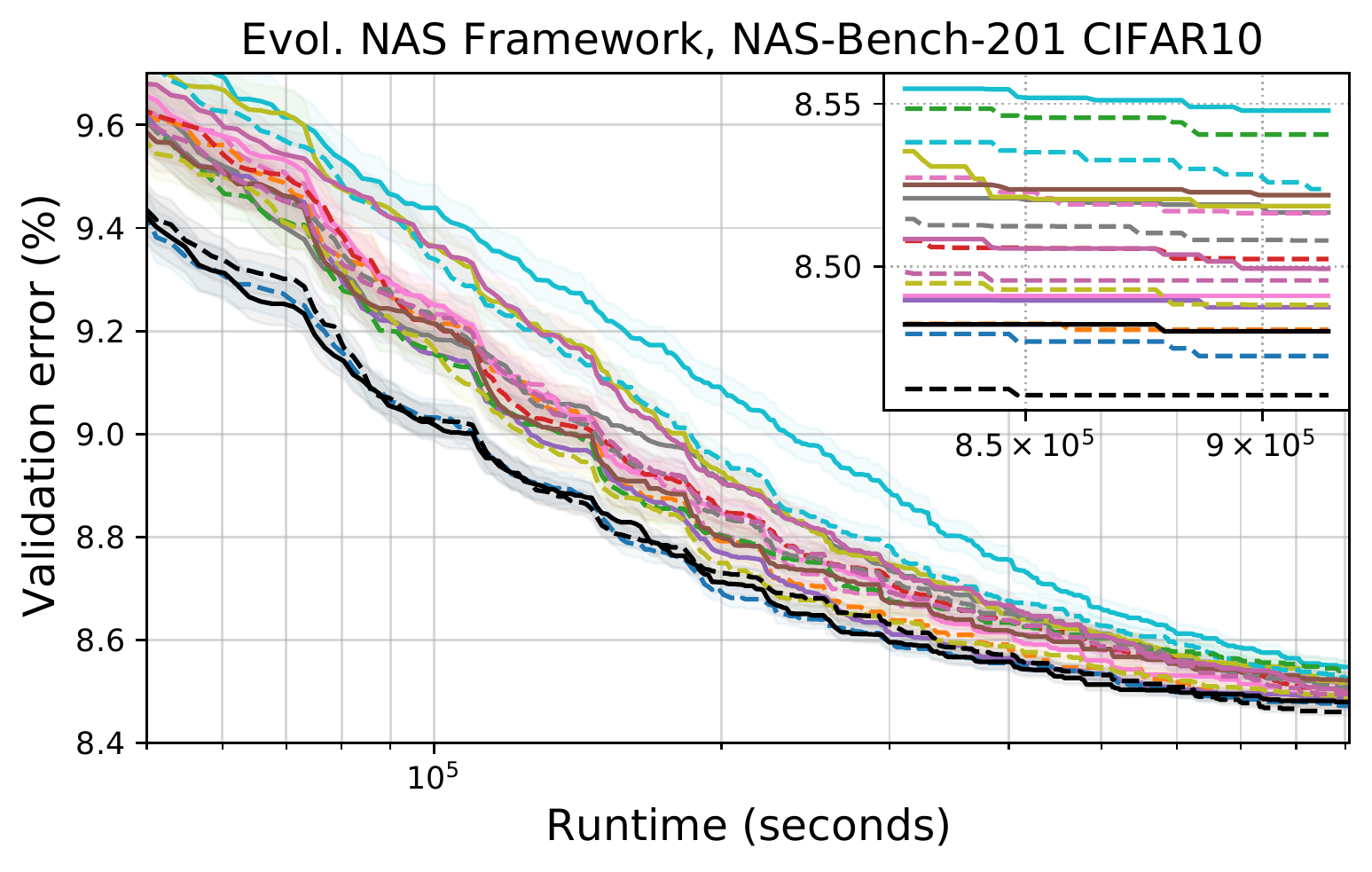}}
        \raisebox{0.0\height}{\includegraphics[width=.395\columnwidth]{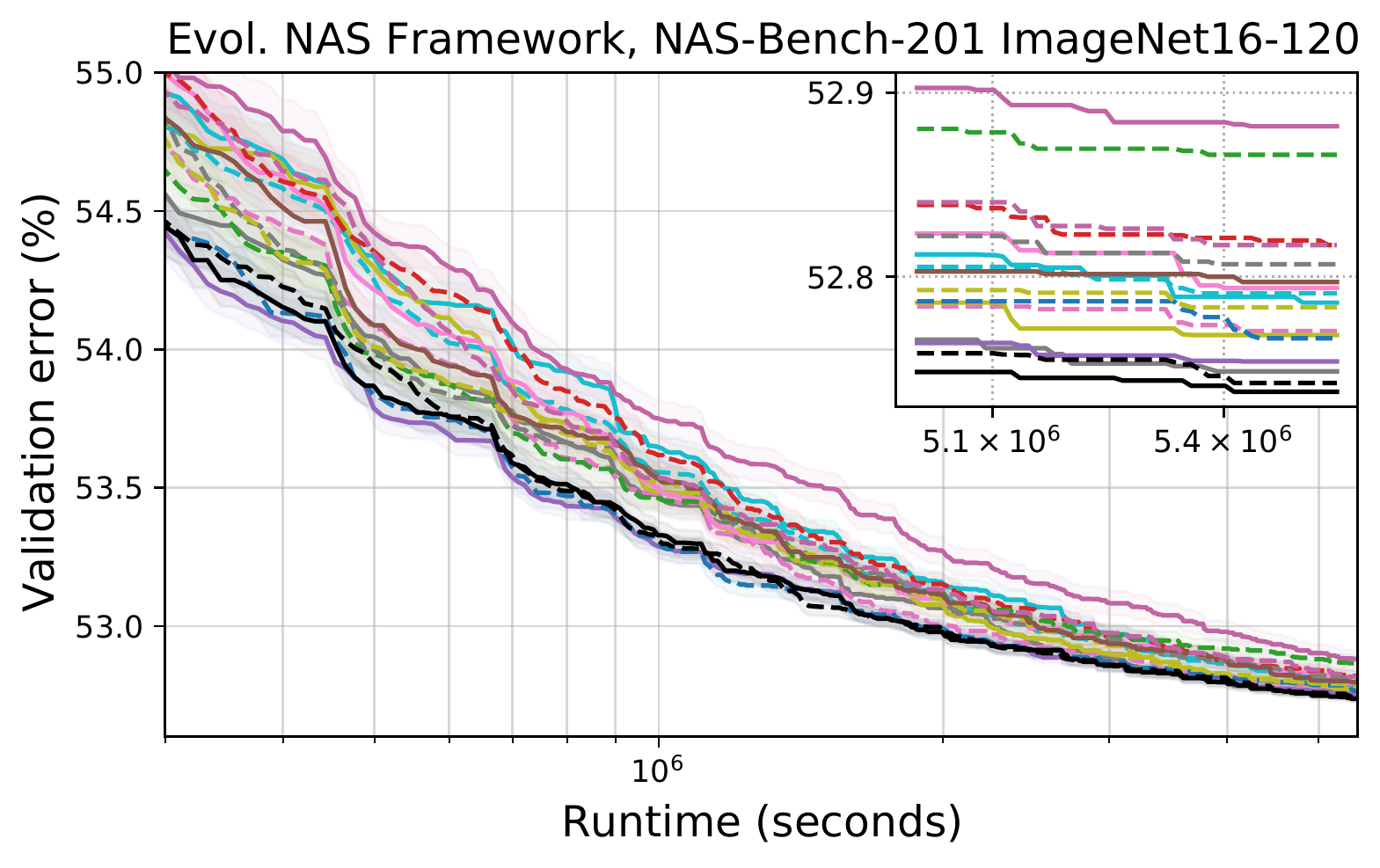}}
        \tabularnewline
        \raisebox{0.0\height}{\includegraphics[width=.39\columnwidth]{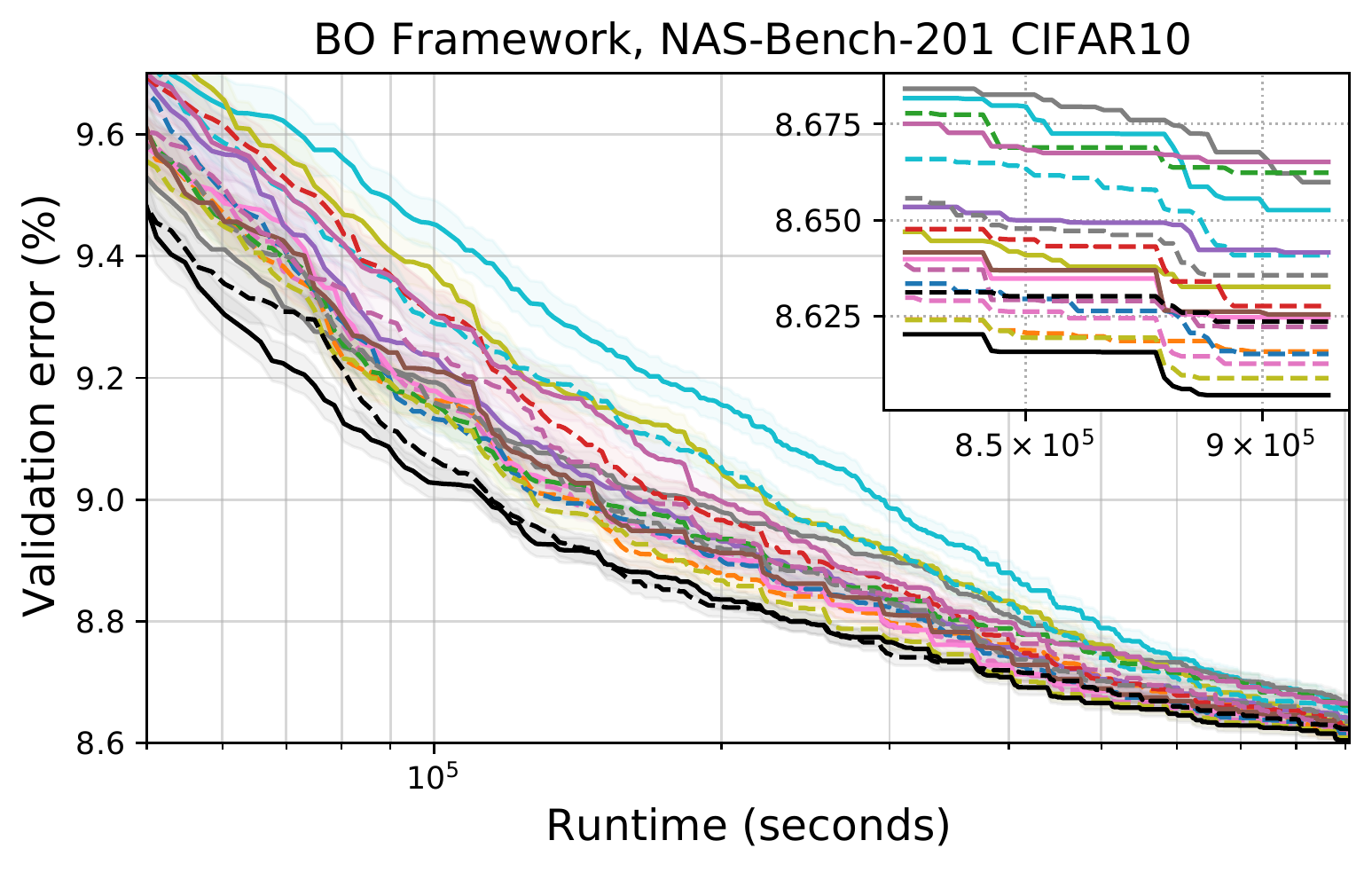}}
        \raisebox{0.0\height}{\includegraphics[width=.40\columnwidth]{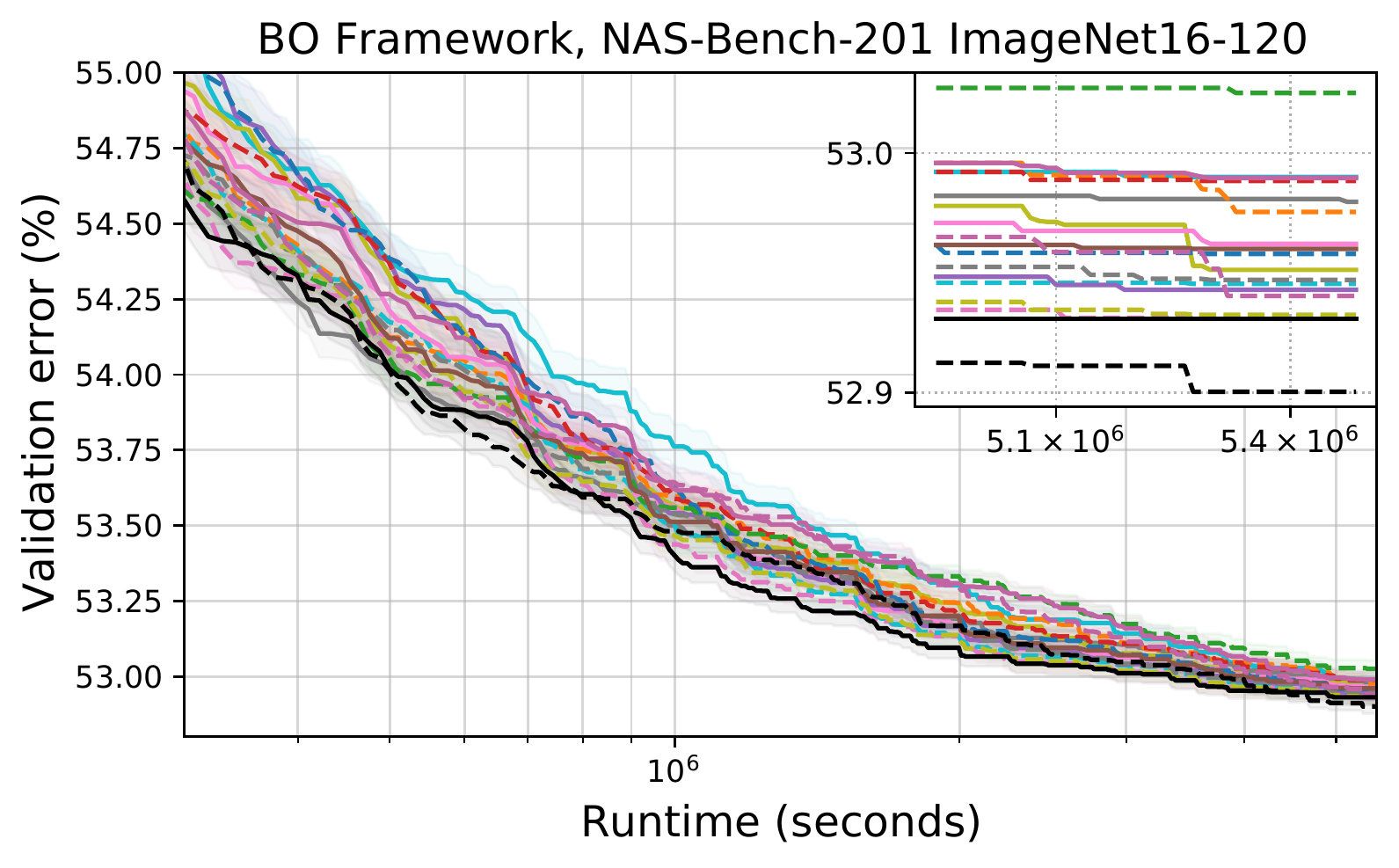}}
        \tabularnewline
      \end{tabular}
        &
      \multirow{-8}{*}{\includegraphics[width=.13\columnwidth]{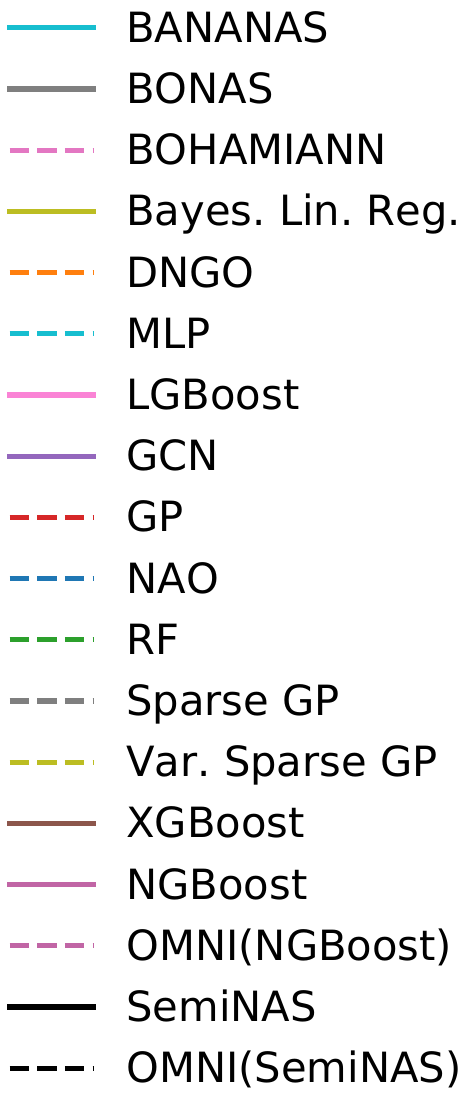}}
     \tabularnewline
    \end{tabular}
\caption{Validation error vs.\ runtime for the predictor-guided evolution framework
and the Bayesian optimization + predictor framework
using different predictors. 
}
\label{fig:nas}
\end{figure}

In Figure~\ref{fig:nas}, we present results for both NAS 
frameworks on NAS-Bench-201
CIFAR-10 and ImageNet16-120 for the 16 model-based predictors.
We also test OMNI, using its lowest query time setting (consisting of
NGBoost + Jacobian covariance), and another version of OMNI that replaces
NGBoost with SemiNAS.
Our results show that the model-based predictors with the top Kendall Tau 
rank correlations in the 
low query time region from Figure~\ref{fig:search_spaces} 
also roughly achieve the best performance when applied for NAS: 
SemiNAS and NAO perform the best for shorter runtime, and boosted trees 
perform best for longer runtime. OMNI(NGBoost) consistently outperforms NGBoost, and OMNI(SemiNAS) often achieves top performance.
This suggests that using zero-cost methods in conjunction with model-based methods is a promising direction for future study.

\subsection{So, how powerful are performance predictors?} 
Throughout Section~\ref{sec:experiments}, 
we tested performance predictors in a variety of settings, by varying
the search spaces, datasets, runtime budgets, and training/test distributions. 
We saw largely the same trends among all of our experiments. Interesting findings
included the success of zero-cost predictors even when compared to model-based
predictors and learning curve extrapolation predictors with longer runtime budgets,
and the fact that information from different families of predictors are complementary.
When choosing a performance predictor for new applications, we recommend deciding on
a target initialization time and query time budget, consulting 
Figures~\ref{fig:search_spaces} and~\ref{fig:app}, and then combining the best
predictors from the desired runtime setting, similar to OMNI. For example, if a
performance predictor with medium initialization time and low runtime is desired
for a search space similar to NAS-Bench-201 or DARTS, we recommend using NGBoost
with Jacobian covariance and SynFlow as additional features.

\section{Societal Impact} \label{sec:impact}
Our hope is that our work will have a positive impact on the AutoML community
by making it quicker and easier to develop and fairly compare performance predictors.
For example, AutoML practitioners can consult our experiments to more easily
decide on the performance prediction methods best suited to their application,
rather than conducting computationally intensive experiments of their 
own~\citep{patterson2021carbon}.
Furthermore, AutoML researchers can use our library to develop new performance
prediction techniques and compare new methods to 31 other algorithms across
four search spaces.
Since the topic of this work is AutoML, it is a level of abstraction away from real
applications. This work may be used to improve deep learning applications, both beneficial
(e.g.\ reducing $\text{CO}_2$ emissions), or harmful (e.g.\ creating language models
with heavy bias) to society.

\section{Conclusions and Limitations} \label{sec:conclusion}
In this work, we gave the first large-scale study of performance predictors for
neural architecture search. We compared 31 different performance predictors,
including learning curve extrapolation methods, weight sharing methods, zero-cost methods,
and model-based methods. We tested the performance of the predictors in a variety of
settings and with respect to different metrics. 
Although we ran experiments on four different search spaces, it will be interesting to
extend our experiments to even more machine learning tasks beyond image classification and
language modeling. 

Our new predictor, OMNI, is the first predictor to combine complementary information from three
families of performance preditors, leading to substantially improved performance. While the
simplicity of OMNI is appealing, it also opens up new directions for future work by combining
different predictors in more sophisticated ways.
To facilitate follow-up work, we release our code featuring a library of 
performance predictors.
Our goal is for our repository to grow over time as it is used by the
community, so that experiments in our library can be even more comprehensive.

\begin{ack}
This work was done while CW and YL were employed at Abacus.AI.
AZ and FH acknowledge support by the European Research Council (ERC) under the European Union Horizon 
2020 research and innovation programme through grant no. 716721, and by BMBF grant DeToL.
BR was supported by the Clarendon Fund of University of Oxford.
\end{ack}

\newpage
\bibliography{main}
\bibliographystyle{plain}


\newpage
\appendix

\section{NAS Research checklist}\label{app:nas_checklist}
There have been a few recent works which have called for improving the reproducibility and 
fairness in experimental comparisons in NAS research~\citep{randomnas, nasbench, yang2019evaluation}.
This led to the release of a NAS best practices checklist~\citep{lindauer2019best}.
We address each part of the checklist.

\begin{enumerate}

\item \textbf{Best Practices for Releasing Code}\\[0.2cm]
For all experiments you report: 
\begin{enumerate}
  \item Did you release code for the training pipeline used to evaluate the final architectures?
    \answerYes{ We used publicly available NAS benchmarks: NAS-Bench-101, NAS-Bench-201, DARTS/NAS-Bench-301, and
    NAS-Bench-NLP, so we did not train the architectures ourselves.}
  \item Did you release code for the search space
    \answerYes{Since we used NAS benchmarks, this code is already
    publicly available.}
\item Did you release the hyperparameters used for the final evaluation pipeline, as well as random seeds?
  \answerYes{Since we used NAS benchmarks, the final training pipeline was fixed.
  We released our code, which includes the seeds used.}
  \item Did you release code for your NAS method?
  \answerYes{All of our code is available at
  \url{https://github.com/automl/naslib}.}
  \item Did you release hyperparameters for your NAS method, as well as random seeds?
  \answerYes{The hyperparameters are given in Appendix Table \ref{tab:hpo}. 
  We ran 100 trials (random seeds) for each experiment, and we released the code to launch these 
  experiments is in our repository.}
\end{enumerate}

\item \textbf{Best practices for comparing NAS methods}
\begin{enumerate}
  \item For all NAS methods you compare, did you use exactly the same NAS benchmark, including the same dataset (with the same training-test split), search space and code for training the architectures and hyperparameters for that code?
    \answerYes{We only used NAS benchmarks, which means the training details are 
    fixed.}
  \item Did you control for confounding factors (different hardware, versions of DL libraries, different runtimes for the different methods)?
    \answerYes{We only used NAS Benchmarks, which keep these details fixed.}	
    \item Did you run ablation studies?
    \answerYes{Our abalation study for OMNI is in Appendix \ref{subsec:omni}.}
	\item Did you use the same evaluation protocol for the methods being compared?
    \answerYes{We only used NAS Benchmarks, which keep this fixed.}
	\item Did you compare performance over time?
    \answerYes{Our experiments compare performance over initialization time and query time.}
	\item Did you compare to random search?
    \answerNo{We did not compare to random search, although we used other baselines for performance predictors such as random forests and simple MLPs.}
	\item Did you perform multiple runs of your experiments and report seeds?
    \answerYes{We ran 100 trials of each experiment, and we released the code to launch these 
  experiments is in our repository.}
	\item Did you use tabular or surrogate benchmarks for in-depth evaluations?
    \answerYes{We only used tabular and surrogate benchmarks.}

\end{enumerate}

\item \textbf{Best practices for reporting important details}
\begin{enumerate}
  \item Did you report how you tuned hyperparameters, and what time and resources
this required?
    \answerYes{We reported this information in Section \ref{sec:experiments}.}
  \item Did you report the time for the entire end-to-end NAS method
    (rather than, e.g., only for the search phase)?
    \answerYes{We present results along three axes: performance, initialization time, and query time.}
  \item Did you report all the details of your experimental setup?
    \answerYes{We did include all details of the setup in Section \ref{sec:experiments} and Appendix \ref{app:experiments}.}

\end{enumerate}

\end{enumerate}

\section{Details from Section~\ref{sec:experiments} (Experiments)}\label{app:experiments}

In this section, we give more details from Section~\ref{sec:experiments}, 
and we present more experiments.
In Section~\ref{subsec:descriptions},
we give a short description and implementation details for all 31 predictors.
We also mention the licenses for the datasets we use.
Next, in Section~\ref{subsec:hpo}, we give the details for hyperparameter tuning.
Then in Section~\ref{subsec:experiments}, we give detailed experimental results for all search spaces.
After that, in Section~\ref{subsec:omni}, we give pseudo-code and an ablation study for OMNI.

\subsection{Descriptions and Implementation Details}\label{subsec:descriptions}
We describe all 31 methods that we used.

\begin{itemize}
    \item \textbf{BANANAS.} The BANANAS~\citep{bananas} predictor is a model-based
    predictor consisting of an ensemble of three MLPs. Architectures are
    encoded using the \emph{path encoding}, which encodes each possible path from the input
    to the output of the cell as a bit.  We use the code from the original repository, but with PyTorch for the MLPs instead of Tensorflow.
    \item \textbf{Bayesian Linear Regression}
    Bayesian Linear Regression~\citep{bishop2006pattern} is one of the simplest 
    Bayesian methods, which assumes that the samples are normally distributed,
    and assumes that the output labels are a linear function of the input
    features (the architecture encoding in our case). 
    Therefore, the predictions are sampled from a distribution computed in closed 
    form given the observations, the posterior of the linear model parameters and 
    the observation noise.
    We use the implementation from the \texttt{pybnn}~\footnote{\url{https://github.com/automl/pybnn}} 
    package
    and the one-hot adjacency matrix encoding.
    \item \textbf{BOHAMIANN.}
    BOHAMIANN~\citep{springenberg2016bayesian} utilizes Bayesian inference via stochastic gradient Hamiltonian Monte Carlo (SGHMC) in order to sample from the estimated posterior of a Bayesian Neural Network. We use the original implementation from the \texttt{pybnn} package
    and the one-hot adjacency matrix encoding.
    \item \textbf{BONAS.} Bayesian optimization for NAS~\citep{bonas} is a NAS algorithm which 
    makes use of a graph convolutional network (GCN) as a model-based predictor within an outer loop of
    Bayesian optimization. In this work, we refer to ``BONAS'' as the GCN predictor.
    The implementation from the original paper works on a constrained version of the DARTS search space where the normal cell and the reduction cell have the same architecture. We adapted BONAS for the three search spaces using the original encoding style. Specifically, given that the DARTS 
    search space includes both normal and reduction cells, we encoded both in one adjacency matrix by 
    arranging the two cells' adjacency matrices diagonally and zero-padding the rest. 
    %
    \item \textbf{DNGO.}
    Deep Networks for Global Optimization (DNGO) is an implementation of Bayesian optimization
    using adaptive basis regression using neural networks instead of
    Gaussian processes to avoid the cubic scaling~\cite{snoek2015scalable}.
    We use the adaptive basis regressor as a model-based predictor, using the
    original code from the \texttt{pybnn} package
    and the one-hot adjacency matrix encoding.
    \item \textbf{Early Stopping with Val.\ Acc.}
    Early Stopping using validation accuracy has been considered in NAS many 
    times (e.g.~\citep{zhou2020econas, hyperband, bohb, zoph2018learning}). 
    It uses the validation accuracy from the most recent epoch trained so far as a proxy for architecture performance.
    \item \textbf{Early Stopping with Val.\ Loss} Early stopping with validation
    loss uses the validation loss instead of the validation accuracy~\citep{ru2020revisiting}.
    \item \textbf{Fisher.}
    Fisher~\citep{abdelfattah2021zerocost} is a zero-cost predictor which computes
    the sum over all gradients of the activations in a neural network.
    Fisher builds off of prior work on channel pruning at 
    initialization~\citep{theis2018faster}.
    We used the implementation from Abdelfattah et al.\ \citep{abdelfattah2021zerocost}.
    \item \textbf{GCN.} The GCN approach for NAS has also been studied by Wen et al.\ \citep{wen2019neural}. 
    Although the code was never released, we used an unofficial implementation online
    from Zhang \citep{zhang2020neural}. The encoding strategy follows that of BONAS.
    \item \textbf{GP.}
    Gaussian Process (GP)~\citep{rasmussen2003gaussian} is a simple model where every finite number of random variables have a joint Gaussian distribution. It is commonly used as the default model choice in Bayesian optimization, and it is fully specified by its mean and covariance functions. The runtime of GPs is
    cubic in the number of datapoints, because the covariance matrix must be 
    inverted to compute the predictive distribution is computed. We use the \texttt{Pyro} implementation~\citep{bingham2019pyro} and the one-hot adjacency matrix encoding.
    \item \textbf{Grad Norm.}
    Grad Norm~\citep{abdelfattah2021zerocost} is a zero-cost predictor 
    which sums the Euclidean norm of the gradients of one minibatch of training
    data. It was used by Abdelfattah et al.\ \citep{abdelfattah2021zerocost} as a baseline when comparing
    other zero-cost methods.
    We used the implementation from Abdelfattah et al.\ \citep{abdelfattah2021zerocost}.
    \item \textbf{Grasp.}
    Grasp~\citep{wang2019picking} was introduced as a technique to prune network 
    weights based on a saliency metric at initialisation. It improved over
    SNIP~\citep{lee2018snip} by approximating the change in gradient norm.
    Later, it was used as a zero-cost predictor in NAS
    by Abdelfattah et al.\ \citep{abdelfattah2021zerocost}. 
    We used the implementation from Abdelfattah et al.\ \citep{abdelfattah2021zerocost}.
    \item \textbf{Jacobian covariance.}
    Jacobian covariance~\citep{mellor2020neural} is a zero-cost method which
    measures the modelling flexibility of a network based on the covariance of its prediction Jacobians with respect to different image inputs. 
    Mellor et al.\ \citep{mellor2020neural} claims that architectures with more flexible prediction at initialization tend to have better test performance after training. We use the original code.
    \item \textbf{LCE.}
    Learning curve extrapolation (LCE)~\citep{domhan2015speeding} takes in a
    partial learning curve as input, and then extrapolates the curve to a chosen
    epoch. It works by fitting the curve to several parametric models, and
    choosing the best model using MCMC. We use the original code, but we used
    a subset of the original parametric models which we found to improve
    performance.
    \item \textbf{LCE-m.}
    LCE-m, introduced as a baseline by Klein et al.\ \citep{lcnet} is an extrapolation method similar to LCE but with a modified loss function that drops LCE's original mechanism of biasing the search to never underestimate the accuracy at the asymptote of the
    curve. The list of parametric models was also partially changed.
    Note that this is not to be confused with the learning curve \emph{prediction}
    method from Klein et al.\ \citep{lcnet}.
    We used the original code, but we used a subset of the parametric models which
    we found to improve performance.
    \item \textbf{LcSVR.}
    This is a hybrid predictor, which extrapolates the learning curves using a trained $\nu$-SVR (LcSVR)~\citep{baker2017accelerating}. Specifically, the $\nu$-SVR model takes in a partial learning curve as well as their 
    first and second derivatives, and the training hyperparameters as the inputs. 
    Like other model-based predictors, the $\nu$-SVR model must be trained by
    using fully evaluated architectures as training data, and like other 
    learning-curve-based predictors, each new query requires partially training
    the architecture. We use the original code.
    \item \textbf{LGBoost.} Light Gradient Boosting Machine (LightGBM or 
    LGBoost)~\citep{ke2017lightgbm}
    was designed to be a more lightweight gradient boosting implementation.
    LGBoost has been
    used as a model-based predictor both in NAS algorithms~\citep{luo2020neural} 
    and in the creation of NAS-Bench-301~\cite{nasbench301}. We followed the implementation 
    of Luo et al.\ \citep{luo2020neural} and used the one-hot adjacency matrix encoding. 
    We reduced the minimum number of data points in a leaf to 5 so that LGBoost could run
    with smaller training set sizes.
    \item \textbf{MLP.} The multilayer perceptron (MLP) predictor is a model-based predictor 
    consisting of a fully-connected neural network, which has been used by prior work as a
    baseline for comparisons~\citep{bananas, alphax}. We use the implementation in BANANAS,
    in which the default network has width 10 and depth 20.
    To encode the architecture, we used a one-hot encoding of the adjacecy matrix and list of
    operations, as in most prior work in NAS~\citep{nasbench}.
    \item \textbf{NAO.}
    Neural Architecture Optimization makes use of an encoder-decoder~\citep{luo2018neural}. The original neural architecture is mapped to a continuous representation via an LSTM encoder network. Performance prediction is powered by a feedfoward neural network, and the decoder is built with
    an LSTM and attention mechanisms. We used the implementation from SemiNAS~\citep{seminas}.
    \item \textbf{NGBoost.} Natural gradient boosting (NGBoost)~\citep{duan2020ngboost} is 
    another gradient-boosted method that uses natural gradients in order to enable uncertainty 
    estimates of the predictions. It has been studied as a model-based predictor in the creation of
    NAS-Bench-301~\citep{nasbench301}. We used the original code and the one-hot adjacency matrix
    encoding.
    \item \textbf{OneShot.}
    We derive the OneShot predictor following a similar procedure as in prior work~\citep{zela2020understanding}. 
    We first train the weight-sharing model (supernetwork) with normal SGD training. The number of epochs and number of training examples are picked based on a grid search optimizing the Spearman rank correlation on a validation set. After the weight-sharing network is trained, 
    we use it as a predictor for the performance of an architecture by computing the performance on the validation set using only the subpath in the supernetwork corresponding to that architecture. 
    The rest of the supernetwork is zeroed out.
    \item \textbf{Random Forest.} Random forests~\citep{liaw2002classification}
    consist of ensembles of decision trees. Random forests have been studied as 
    model-based predictors in the creation of NAS-Bench-301~\citep{nasbench301}.
    We use the Scikit-learn implementation~\citep{pedregosa2011scikit} and the one-hot
    adjacency matrix encoding.
    \item \textbf{RSWS.}
    Random Search with Weight Sharing (RSWS)~\citep{randomnas} is used exactly the same as the OneShot predictor when predicting the performance of an architecture. The only difference is the way RSWS optimizes the weight-sharing model. Instead of training it as a single network, at each mini-batch iteration, RSWS uniformly samples one architecture from the search space and updates the weights of \emph{only} that operations corresponding to that architecture in the supernetwork.
    \item \textbf{SemiNAS.}
    Semi-supervised NAS (SemiNAS) uses semi-supervised learning with the NAO architecture~\citep{seminas}.
    Specifically, additional synthetic training data is generated and used to train the architecture. The performance of synthetic training data is predicted by the NAO predictor.
    We use the original implementation, reducing the ratio of real vs.\ synthetic data to 1:1 and
    decreasing the number of epochs to 100 to decrease its extreme training time.
    SemiNAS was originally implemented for NAS-Bench-101 and ProxylessNAS~\citep{proxylessnas} search spaces. We adapted it for NAS-Bench-201 and DARTS using the original encoding style. Encoder/decoder lengths and vocabulary sizes of the NAO predictor were changed accordingly to fit the new search spaces.
    \item \textbf{SNIP.}
    Single-shot network pruning (SNIP) was first proposed in~\citep{lee2018snip} 
    as a technique to prune network weights at initialisation. SNIP was later 
    adapted by Abdelfattah et al.\ \citep{abdelfattah2021zerocost} to become a zero-cost predictor for ranking architecture performance. We used the implementation 
    from Mikler \citep{mikler2019snip}.
    \item \textbf{SoTL.}
    Sum of training losses (SoTL)~\citep{ru2020revisiting} is a learning curve-based predictor, which estimates the generalization performance of architectures based on the sum of their training losses over the epochs trained so far. SoTL does not attempt to predict the final validation accuracy as in model-based
    predictors, but does output a score with high rank correlation with respect to the final validation accuracy. We use the original implementation of SoTL~\citep{ru2020revisiting}.
    \item \textbf{SoTL-E.}
    Sum of training losses at last epoch E (SoTL-E)~\citep{ru2020revisiting}
    is very similar to SoTL, but it only considers the sum over the training batches in the most recent epoch trained so far.
    \item \textbf{Sparse GP.}
    Compared to classical GPs, Sparse Gaussian Processes (Sparse GPs)~\citep{candela05} scale better to large amounts of data by introducing the so-called inducing variables to summarize the training data. To bypass the expensive marginal likelihood estimation, variational inference is used in order to approximate the posterior distribution. We use the \texttt{Pyro} implementation~\citep{bingham2019pyro} and the one-hot adjacency matrix encoding.
    \item \textbf{SynFlow.}
    Synaptic Flow (SynFlow)~\citep{tanaka2020pruning} was introduced as a technique 
    to prune network weights at initialisation based on a saliency metric. 
    It improved over SNIP~\citep{lee2018snip} and Grasp~\citep{wang2019picking} 
    by avoiding layer collapse when performing parameter pruning by taking a product
    of all parameters in the network. It was used as a zero-cost predictor in NAS
    by Abdelfattah et al.\ \citep{abdelfattah2021zerocost}. 
    We used the implementation from Abdelfattah et al.\ \citep{abdelfattah2021zerocost}.
    \item \textbf{Variational Sparse GP.}
    Variational Sparse Gaussian Process (Var.\ Sparse GP)~\citep{titsias2009variational} is similar to the Sparse GP, but it can handle non-Gaussian likelihoods. We use the \texttt{Pyro} implementation~\citep{bingham2019pyro} and the one-hot adjacency matrix encoding.
    \item \textbf{XGBoost.} eXtreme Gradient Boosting (XGBoost)~\citep{chen2016xgboost}
    is the first of three gradient-boosted decision tree implementations that we used. 
    XGBoost has been used as a model-based predictor in the creation of
    NAS-Bench-301~\citep{nasbench301}. We used the original code and the
    one-hot adjacency matrix encoding.
\end{itemize}

In Table~\ref{tab:licenses}, we discuss the licenses for the NAS datasets
we used.

\begin{table}[t]
\caption{Licenses for the datasets that we use.}
\centering
\begin{tabular}{@{}l|c|c@{}}
\toprule
\multicolumn{1}{l}{\textbf{Dataset}} & \multicolumn{1}{c}{\textbf{License}} & \multicolumn{1}{c}{\textbf{URL}} \\
\midrule 
NAS-Bench-101 & Apache 2.0 & \url{https://github.com/google-research/nasbench} \\
NAS-Bench-201 & MIT & \url{https://github.com/D-X-Y/NAS-Bench-201} \\
NAS-Bench-301 & Apache 2.0 & \url{https://github.com/automl/nasbench301} \\
NAS-Bench-NLP & None & \url{https://github.com/fmsnew/nas-bench-nlp-release} \\
\bottomrule
\end{tabular}
\label{tab:licenses}
\end{table}

\subsection{Hyperparameter Tuning}\label{subsec:hpo}
Now we give the details of hyperparameter tuning. 
Recall from Section~\ref{sec:experiments} that we run cross-validation on all model-based 
predictors that we studied, for two reasons. First, we compared 16 model-based predictors directly
from the original repositories when possible, but the published hyperparameters from different
methods have significantly different levels of hyperparameter tuning. Therefore, running
more hyperparameter tuning for all predictors will help to level the playing field.
Second, most predictor-based NAS algorithms can naturally utilize cross-validation during the search
to improve performance. This is because the bottleneck for predictor-based 
NAS algorithms is typically the training of architectures, not fitting
the predictor~\citep{snoek2015scalable, lindauer2017smac, bohb}. 
Therefore, adding cross-validation to model-based predictors is a more realistic setting.
Note that the same is not true for other families of performance predictors. For example, 
LCE methods are typically used to replace fully training architectures during NAS, therefore
we would not know the final validation accuracy of these architectures and would not be able to run
cross-validation.

Overall, our goal was to run lightweight cross-validation to level the playing field. For each
model-based predictor, we chose 3-5 hyperparameters and chose ranges based on their default values
from their original repositories.  See Table~\ref{tab:hpo}.
Note that some methods such as the three GP-based methods and the three methods from 
\texttt{pybnn} (Bayes.\ Lin.\ Reg., BOHAMIANN, DNGO) already had cross-validation built in,
so we excluded these.
In all of our experiments, we ran random search on each performance predictor for 5000 iterations,
with a maximum total runtime of 15 minutes.

\begin{table}[ht]
\caption{Hyperparameters of the model-based methods and their default values from their original repositories.}
\label{tab:hpo}
\centering
\resizebox{0.85\linewidth}{!}{\begin{tabular}{@{}lllll@{}}
\toprule
Model & Hyperparameter & Range & Log-transform & Default Value \\
\midrule
\multicolumn{1}{c}{\multirow{3}{*}{BANANAS}} & Num.\ layers & {[}5, 25{]} & false & 20 \\
\multicolumn{1}{c}{} & Layer width & {[}5, 25{]} & false & 20 \\
\multicolumn{1}{c}{} & Learning rate & {[}0.0001, 0.1{]} & true & 0.001 \\
\midrule
\multirow{3}{*}{BONAS} & Num.\ layers & {[}16, 128{]} & true & 64 \\
& Batch size & {[}32, 256{]} & true & 128 \\
& Learning rate & {[}0.00001, 0.1{]} & true & 0.0001 \\
\midrule
\multirow{4}{*}{GCN} & Num.\ layers & {[}64, 200{]} & true & 144 \\
& Batch size & {[}5, 32{]} & true & 7 \\
& Learning rate & {[}0.00001, 0.1{]} & true & 0.0001 \\
& Weight decay & {[}0.00001, 0.1{]} & true & 0.0003 \\
\midrule
\multirow{3}{*}{LCSVR} & Penalty param.\ & {[}0.00001, 10{]} & true & - \\
& Kernel coefficient & {[}0.00001, 10{]} & false & - \\
& Frac.\ support vectors & {[}0, 1{]} & false & - \\
\midrule
\multirow{3}{*}{LGBoost} & Num.\ leaves & {[}10, 100{]} & false & 31 \\
& Learning rate & {[}0.001, 0.1{]} & true & 0.05 \\
& Feature fraction & {[}0.1, 1{]} & false & 0.9 \\
\midrule
\multirow{3}{*}{MLP} & Num.\ layers & {[}5, 25{]} & false & 20 \\
& Layer width & {[}5, 25{]} & false & 20 \\
& Learning rate & {[}0.0001, 0.1{]} & true & 0.001 \\
\midrule
\multirow{3}{*}{NAO} & Num.\ layers & {[}16, 128{]} & true & 64 \\
& Batch size & {[}32, 256{]} & true & 100 \\
& Learning rate & {[}0.00001, 0.1{]} & true & 0.001 \\
\midrule
\multirow{4}{*}{NGBoost} & Num.\ estimators & {[}128, 512{]} & true & 505 \\
& Learning rate & {[}0.001, 0.1{]} & true & 0.081 \\
& Max depth & {[}1, 25{]} & false & 6 \\
& Max features & {[}0.1, 1{]} & false & 0.79 \\
\midrule
\multirow{3}{*}{RF} & Num.\ estimators & {[}16, 128{]} & true & 116 \\
& Max features & {[}0.1, 0.9{]} & true & 0.17 \\
& Min samples (leaf) & {[}1, 20{]} & false & 2 \\
& Min samples (split) & {[}2, 20{]} & true & 2 \\
\midrule
\multirow{3}{*}{SemiNAS} & Num.\ layers & {[}16, 128{]} & true & 64 \\
& Batch size & {[}32, 256{]} & true & 100 \\
& Learning rate & {[}0.00001, 0.1{]} & true & 0.001 \\
\midrule
\multirow{5}{*}{XGBoost} & Max depth & {[}1, 15{]} & false & 6 \\
& Min child weight & {[}1, 10{]} & false & 1 \\
& Col sample (tree) & {[}0, 1{]} & false & 1 \\
& Learning rate & {[}0.001, 0.5{]} & true & 0.3 \\
& Col sample (level) & {[}0, 1{]} & false & 1 \\
\bottomrule
\end{tabular}}
\end{table}

\subsection{Additional Experiments}\label{subsec:experiments}
\begin{figure}
\centering
\raisebox{0.0\height}{\includegraphics[width=.29\columnwidth]{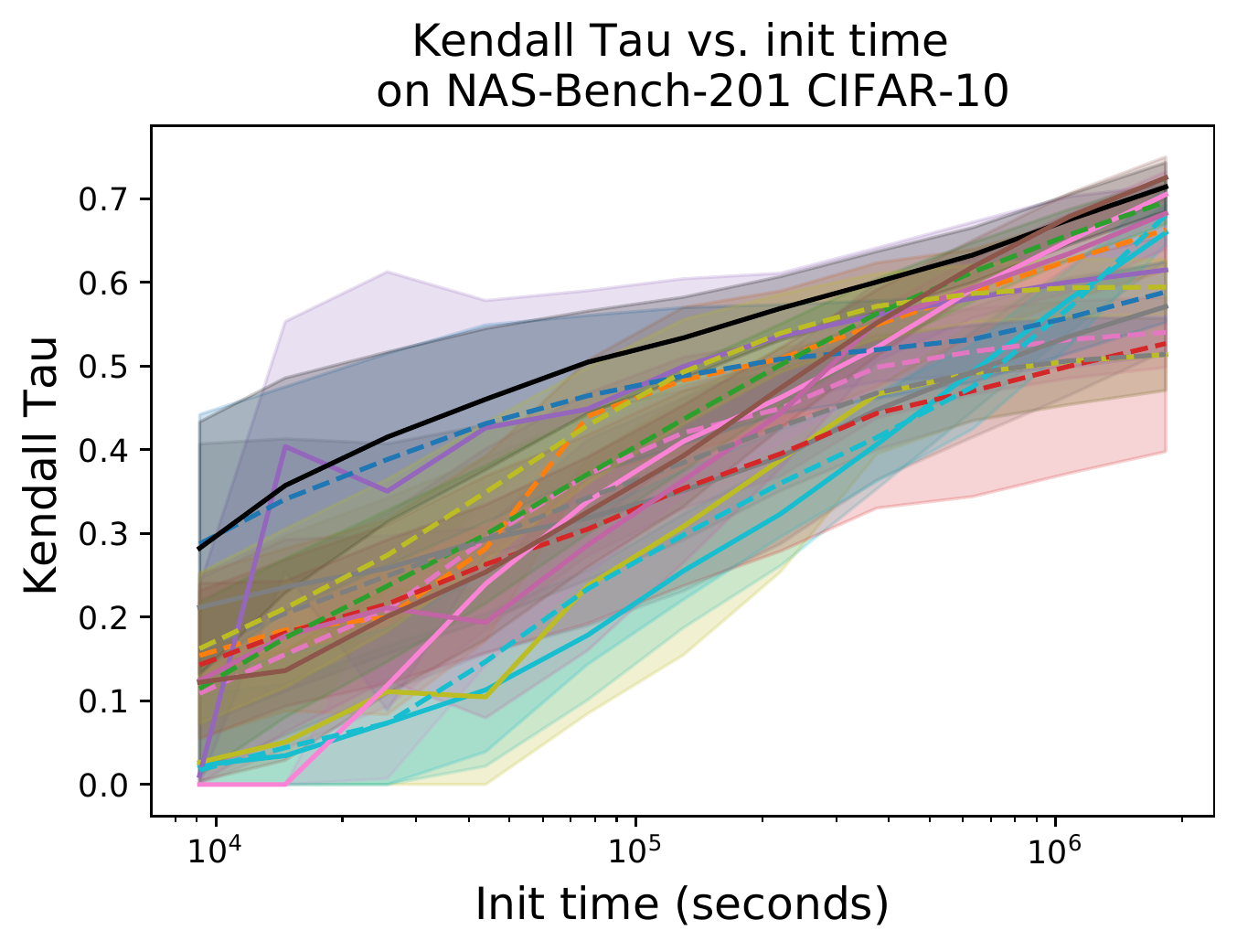}}
\raisebox{0.0\height}{\includegraphics[width=.29\columnwidth]{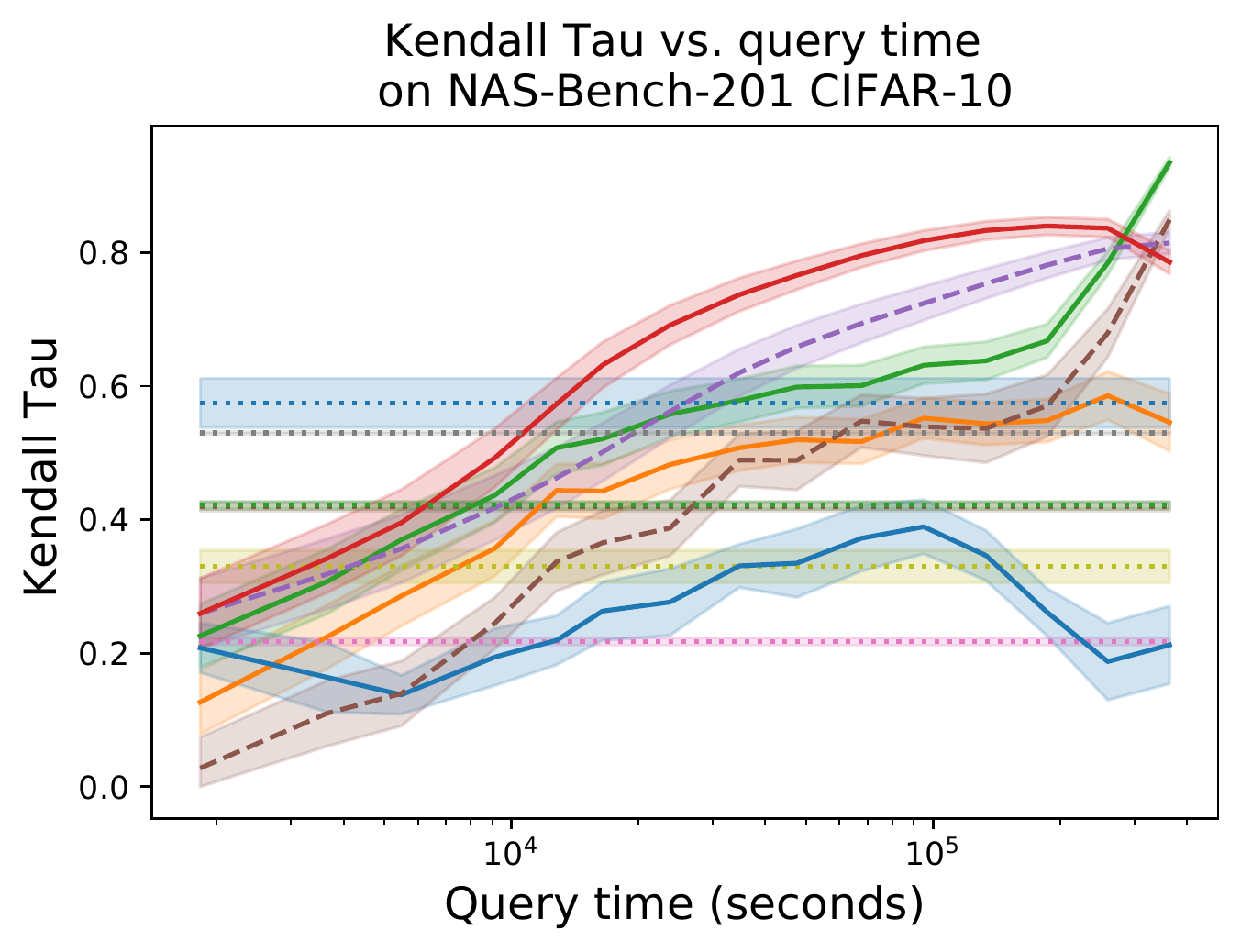}}
\raisebox{0.0\height}{\includegraphics[width=.28\columnwidth]{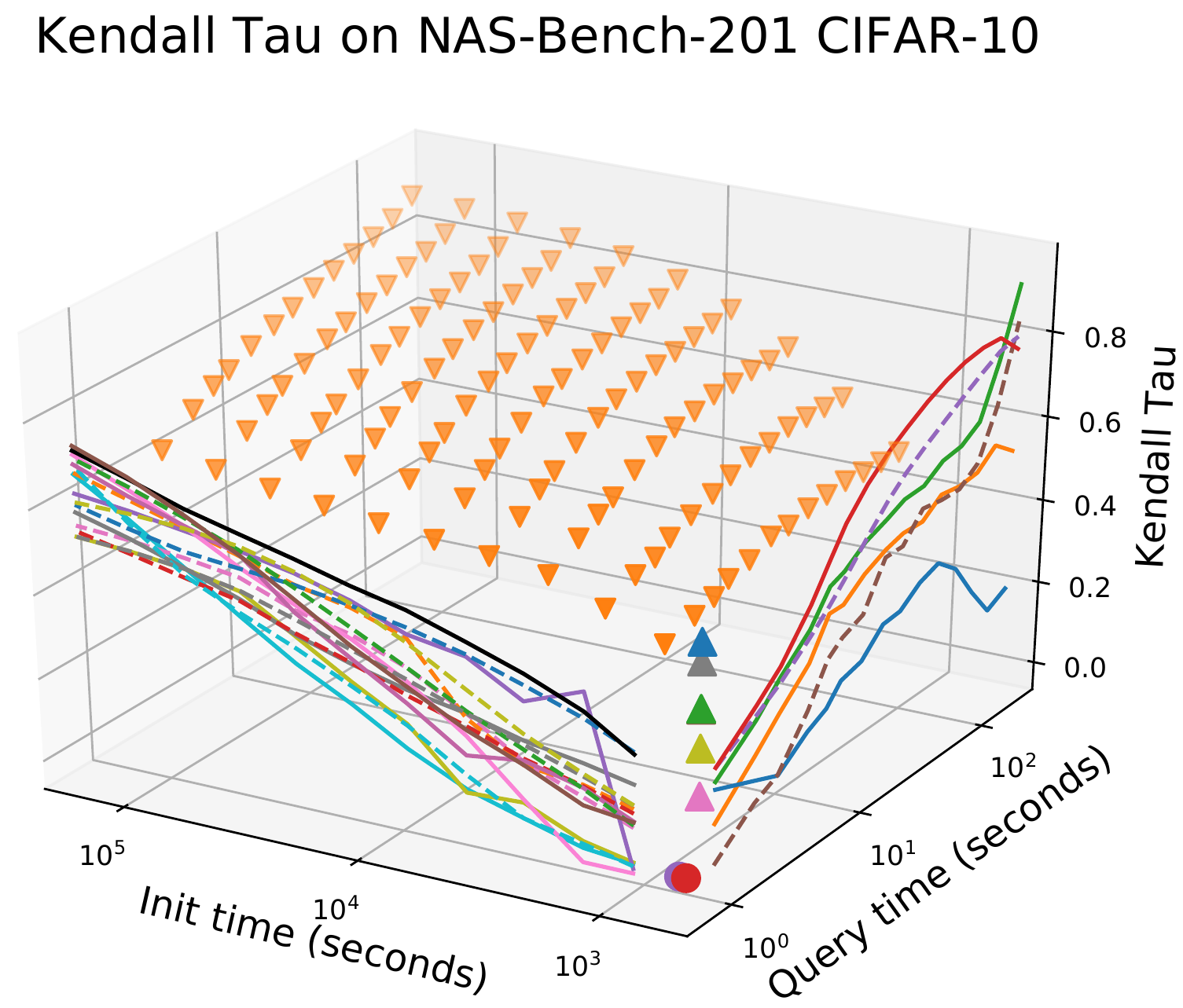}}
\raisebox{0.0\height}{\includegraphics[width=.29\columnwidth]{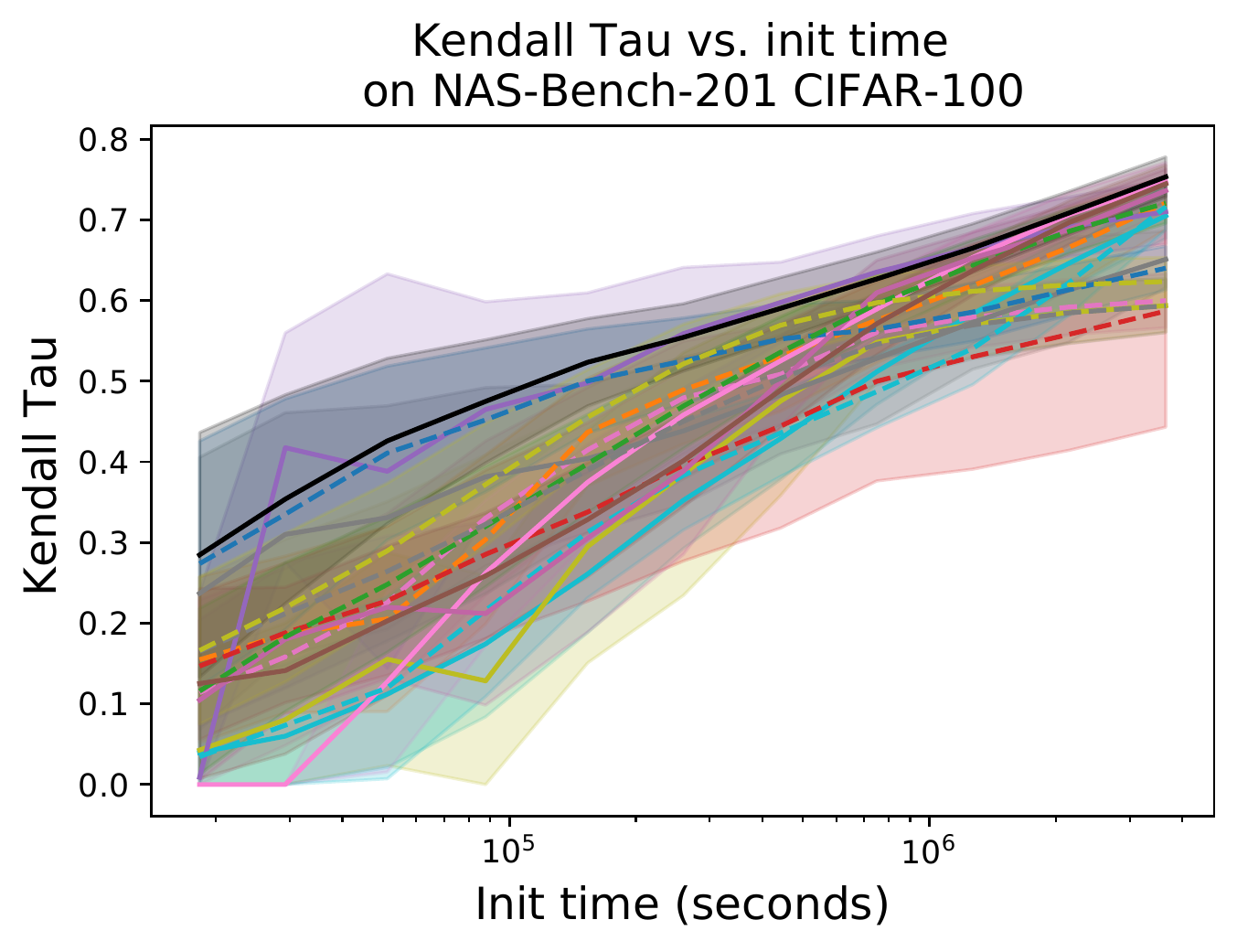}}
\raisebox{0.0\height}{\includegraphics[width=.29\columnwidth]{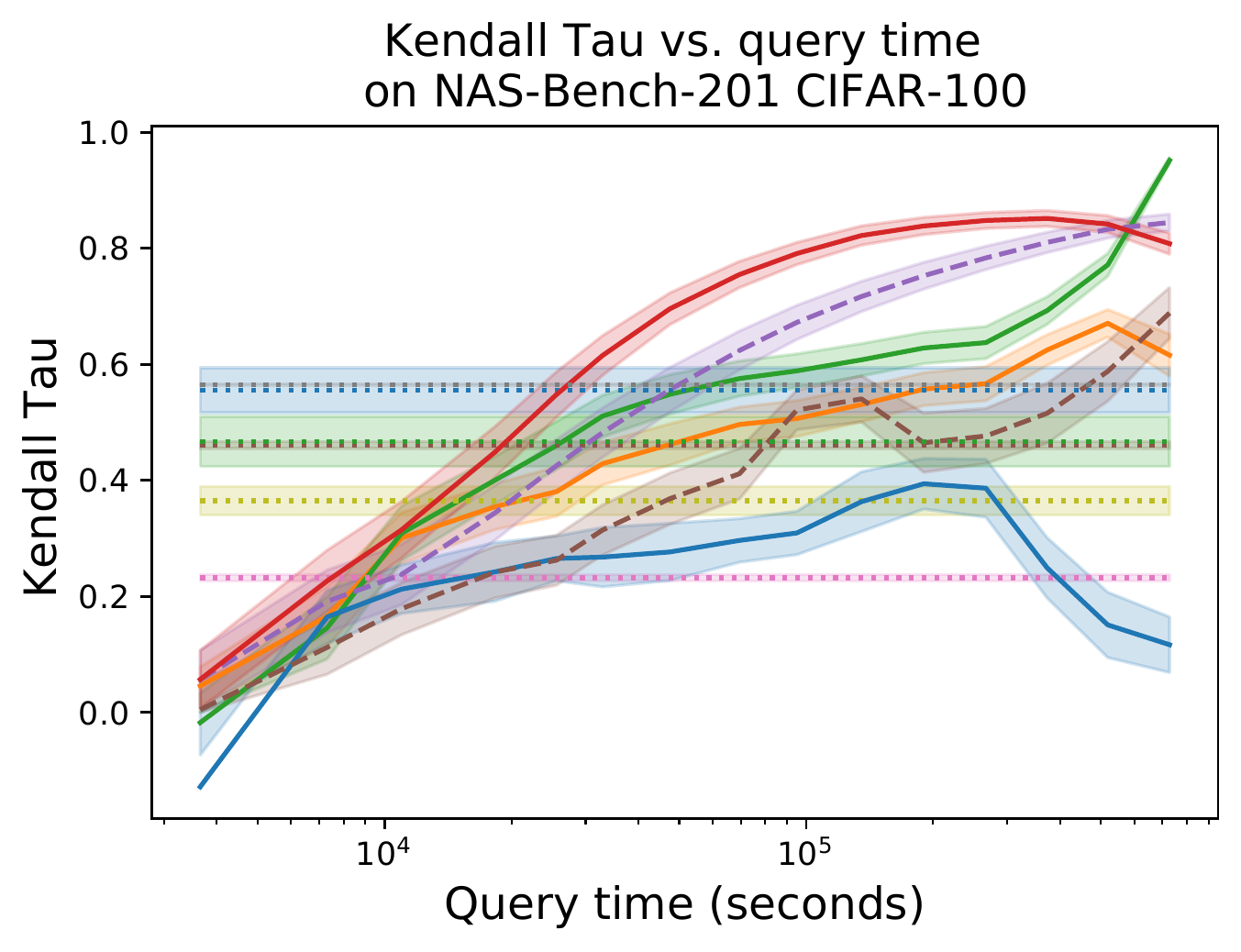}}
\raisebox{0.0\height}{\includegraphics[width=.28\columnwidth]{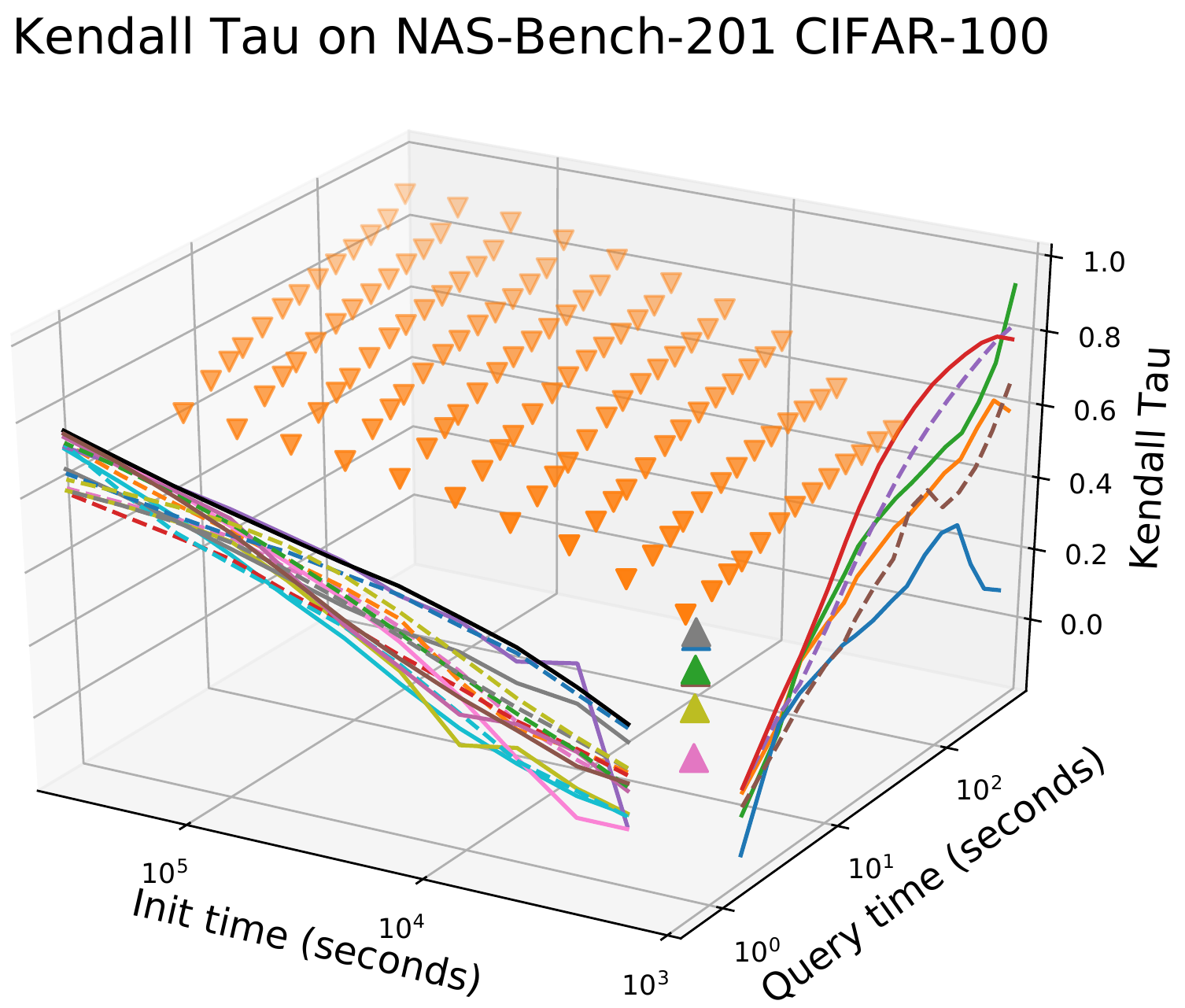}}
\raisebox{0.0\height}{\includegraphics[width=.29\columnwidth]{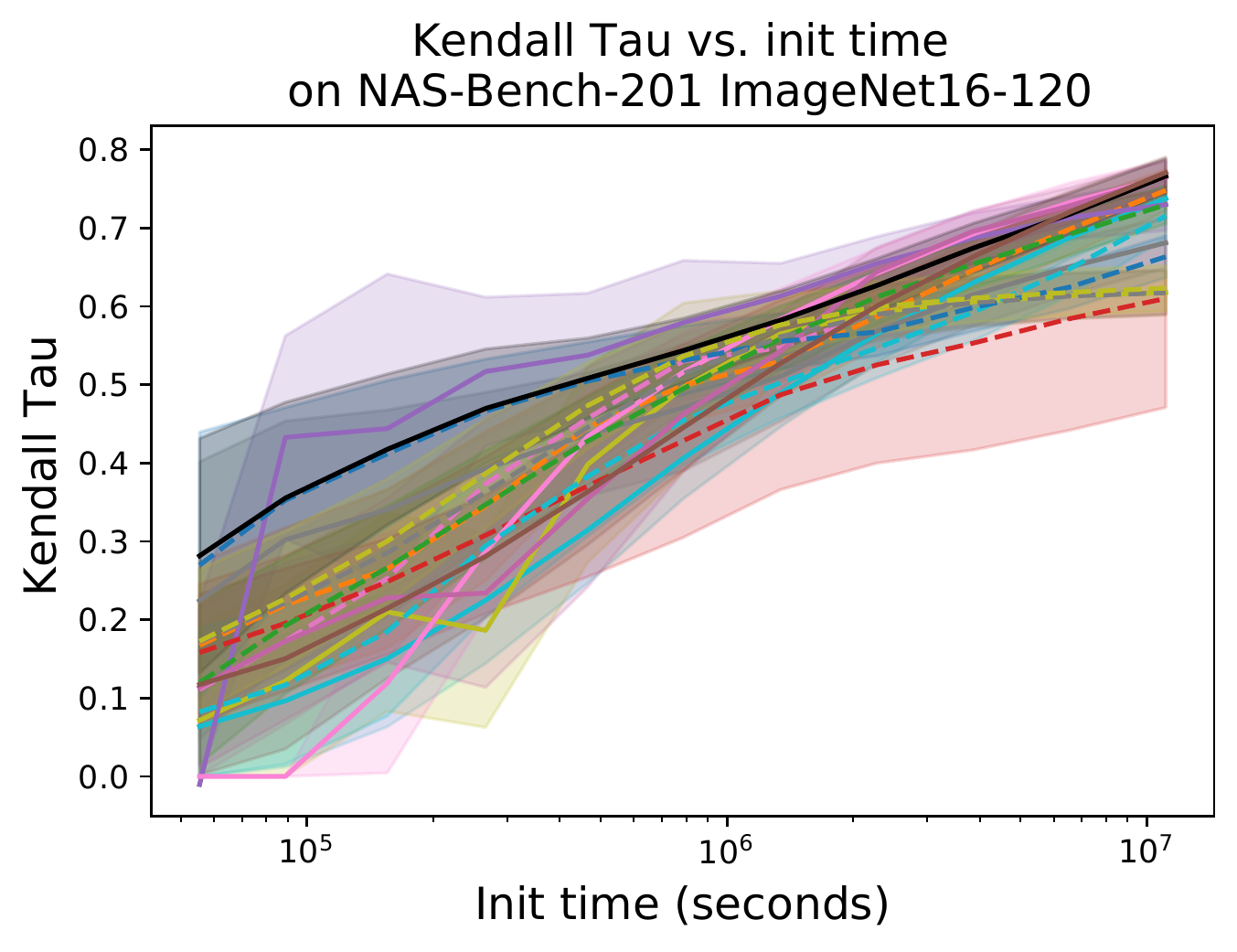}}
\raisebox{0.0\height}{\includegraphics[width=.29\columnwidth]{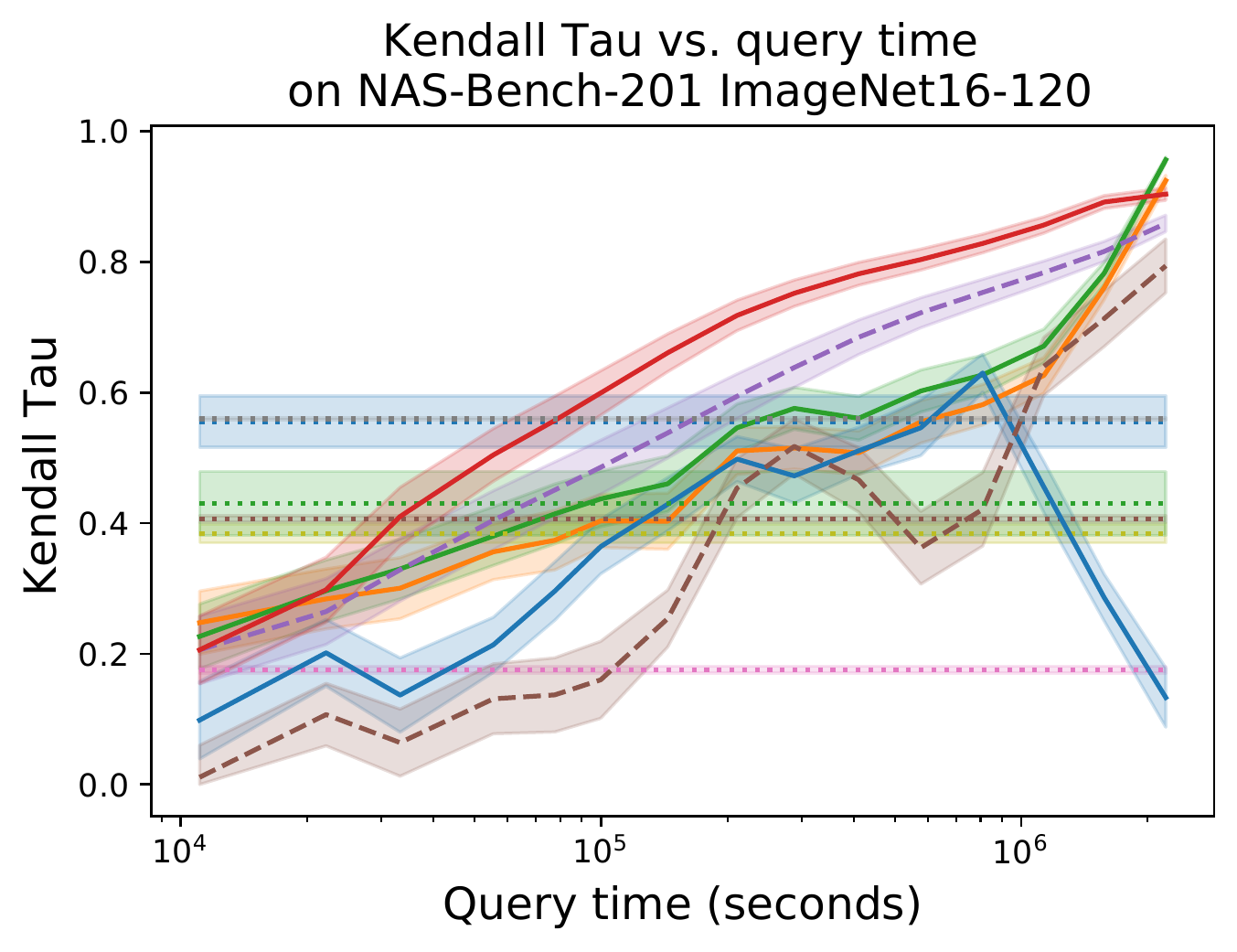}}
\raisebox{0.0\height}{\includegraphics[width=.28\columnwidth]{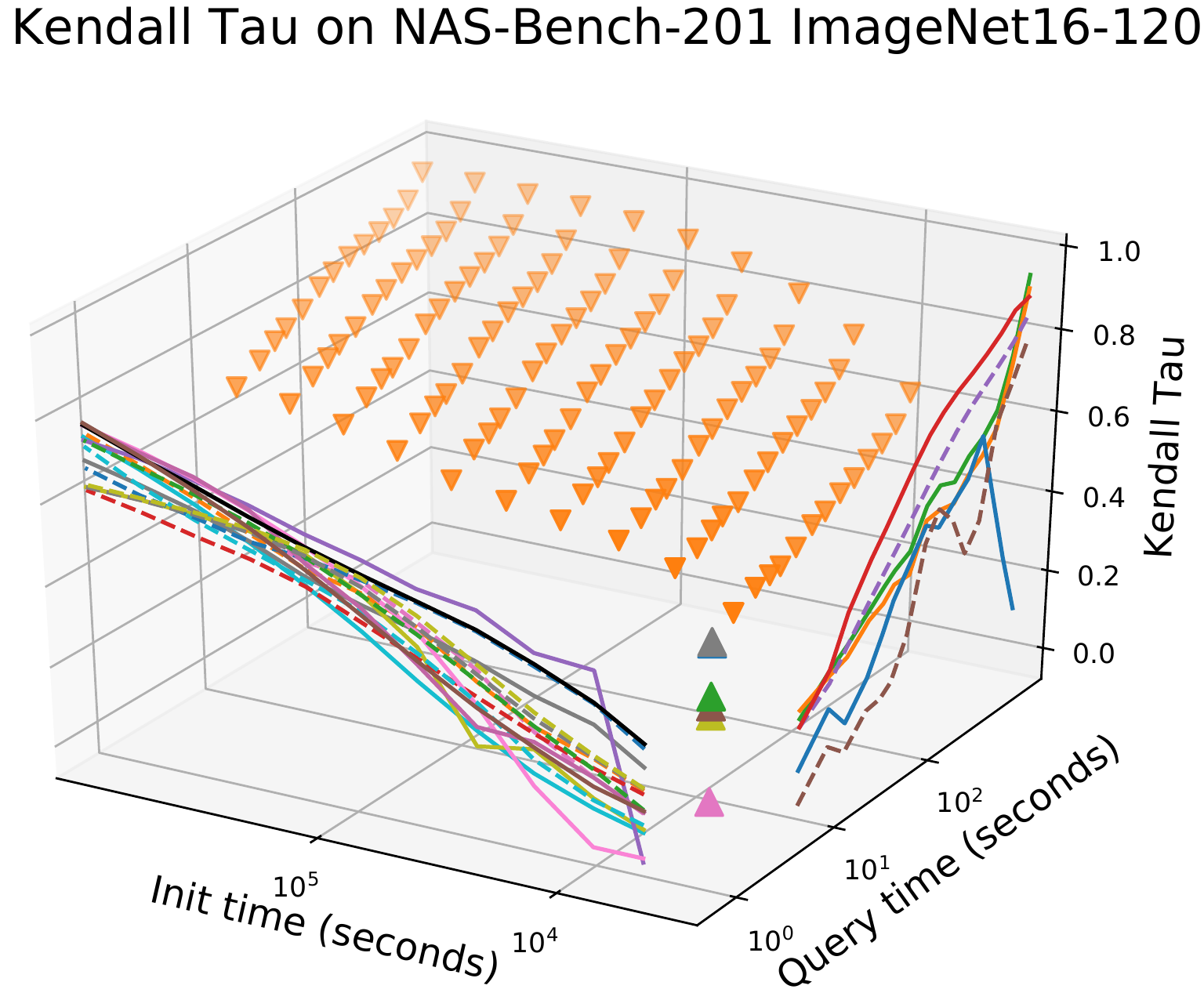}}
\raisebox{0.0\height}{\includegraphics[width=.29\columnwidth]{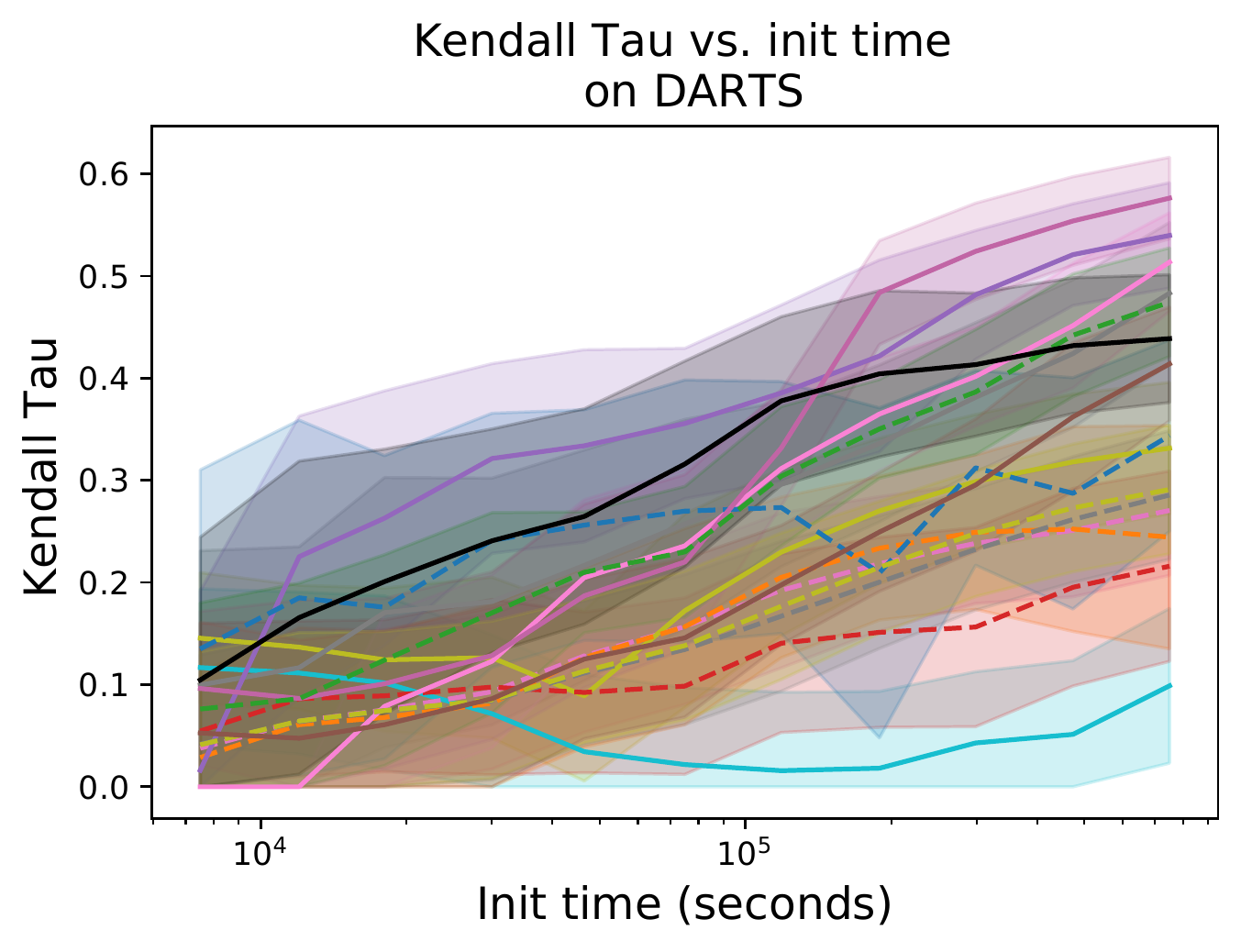}}
\raisebox{0.0\height}{\includegraphics[width=.29\columnwidth]{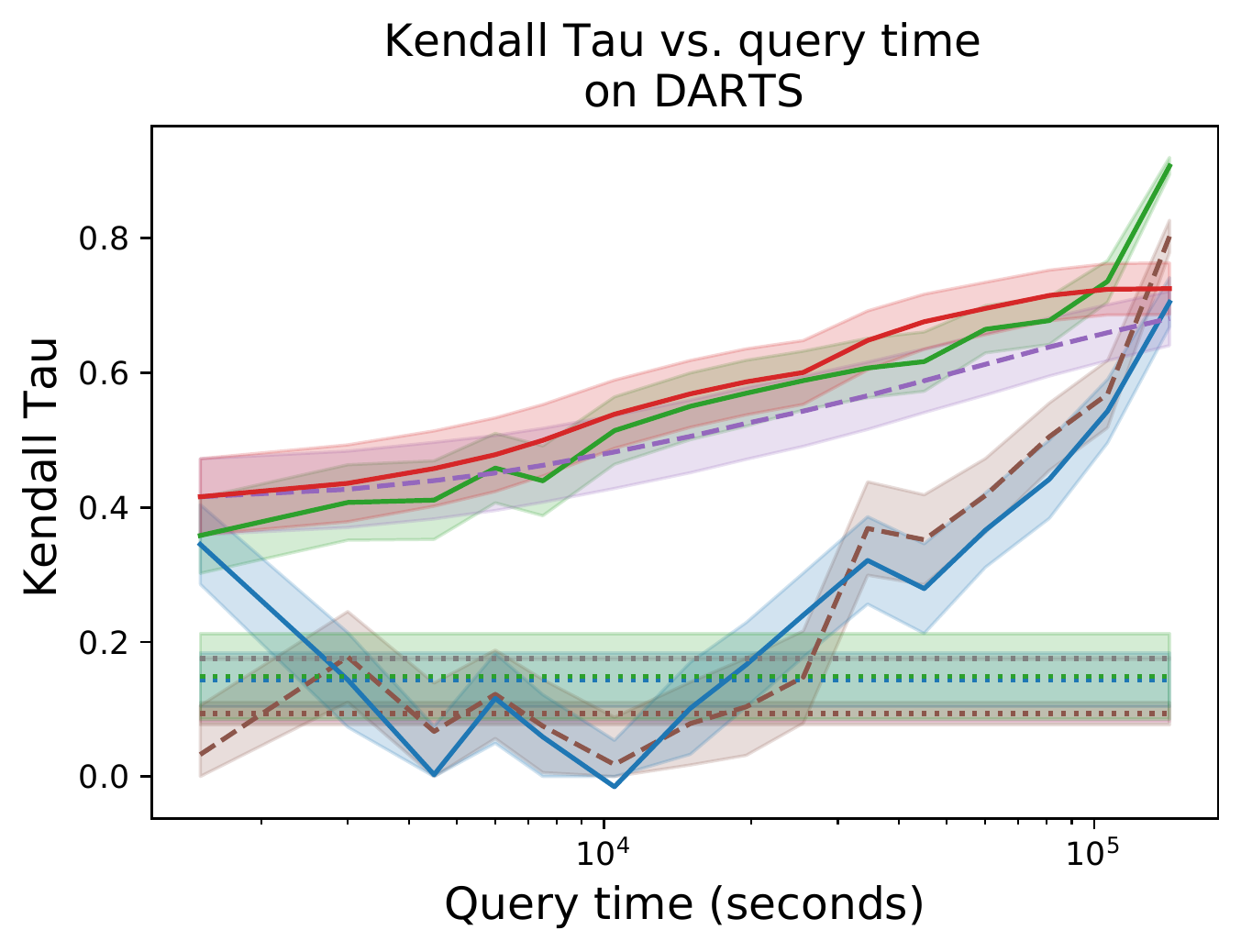}}
\raisebox{0.0\height}{\includegraphics[width=.28\columnwidth]{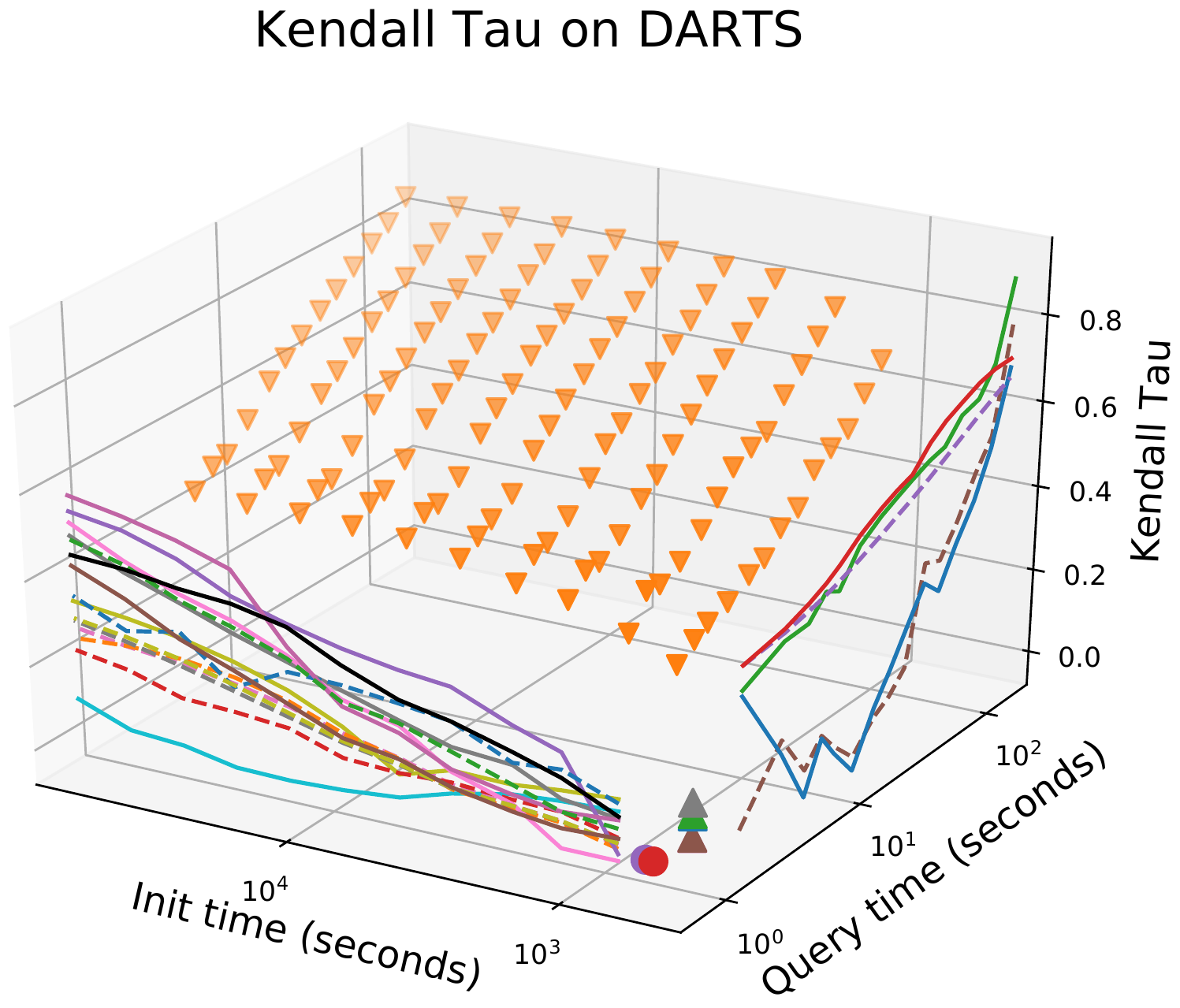}}
\raisebox{0.0\height}{\includegraphics[width=.29\columnwidth]{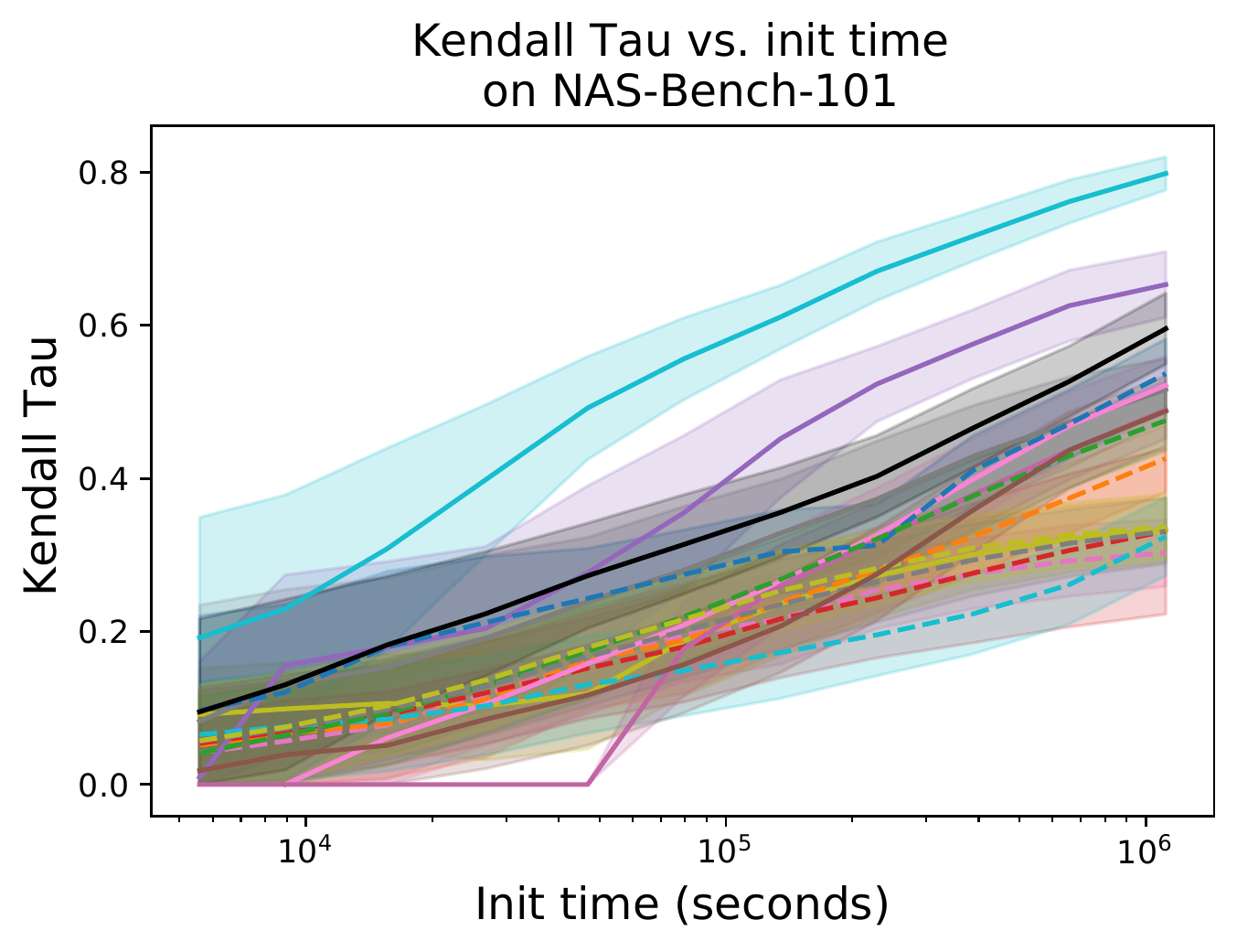}}
\raisebox{0.0\height}{\includegraphics[width=.29\columnwidth]{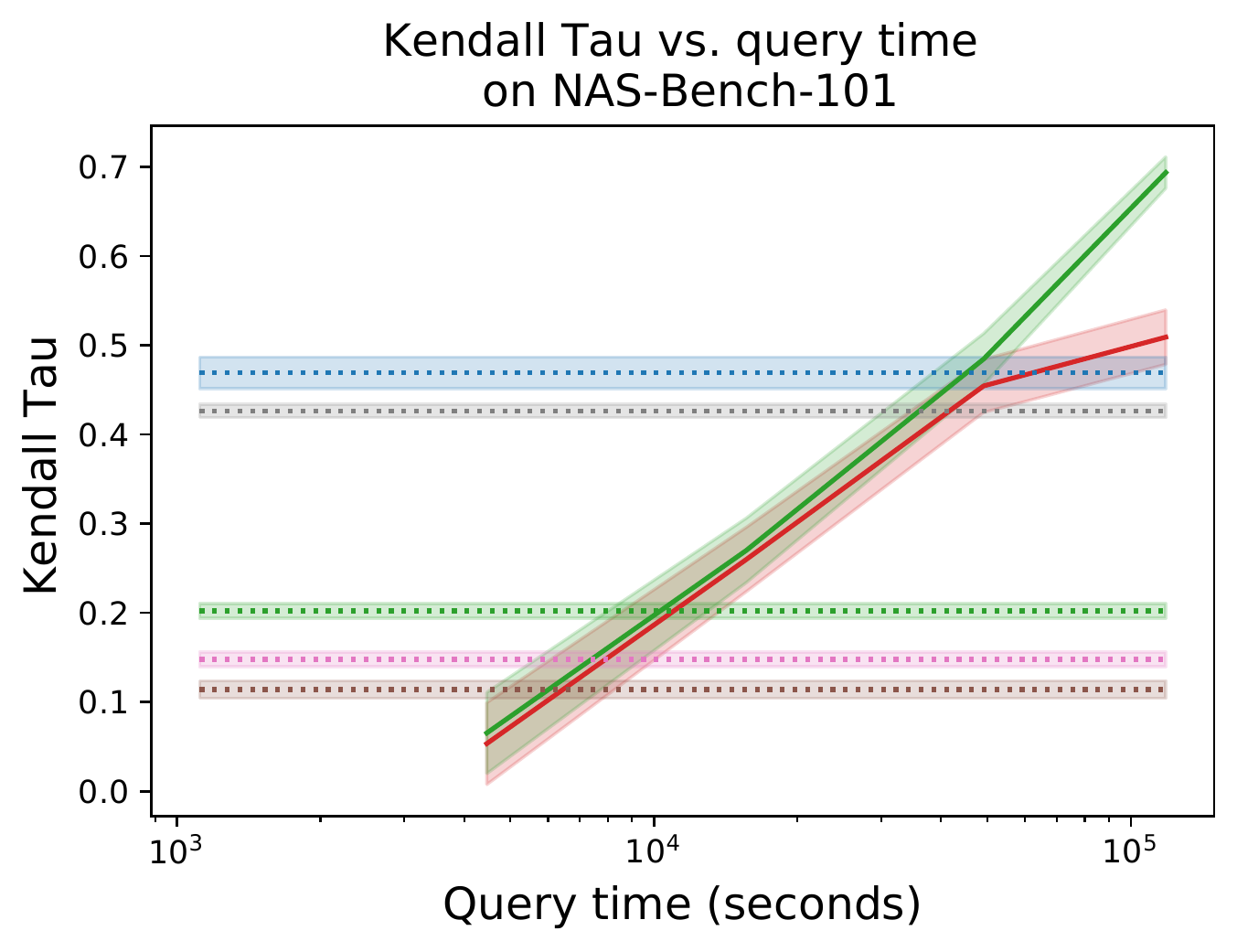}}
\raisebox{0.0\height}{\includegraphics[width=.28\columnwidth]{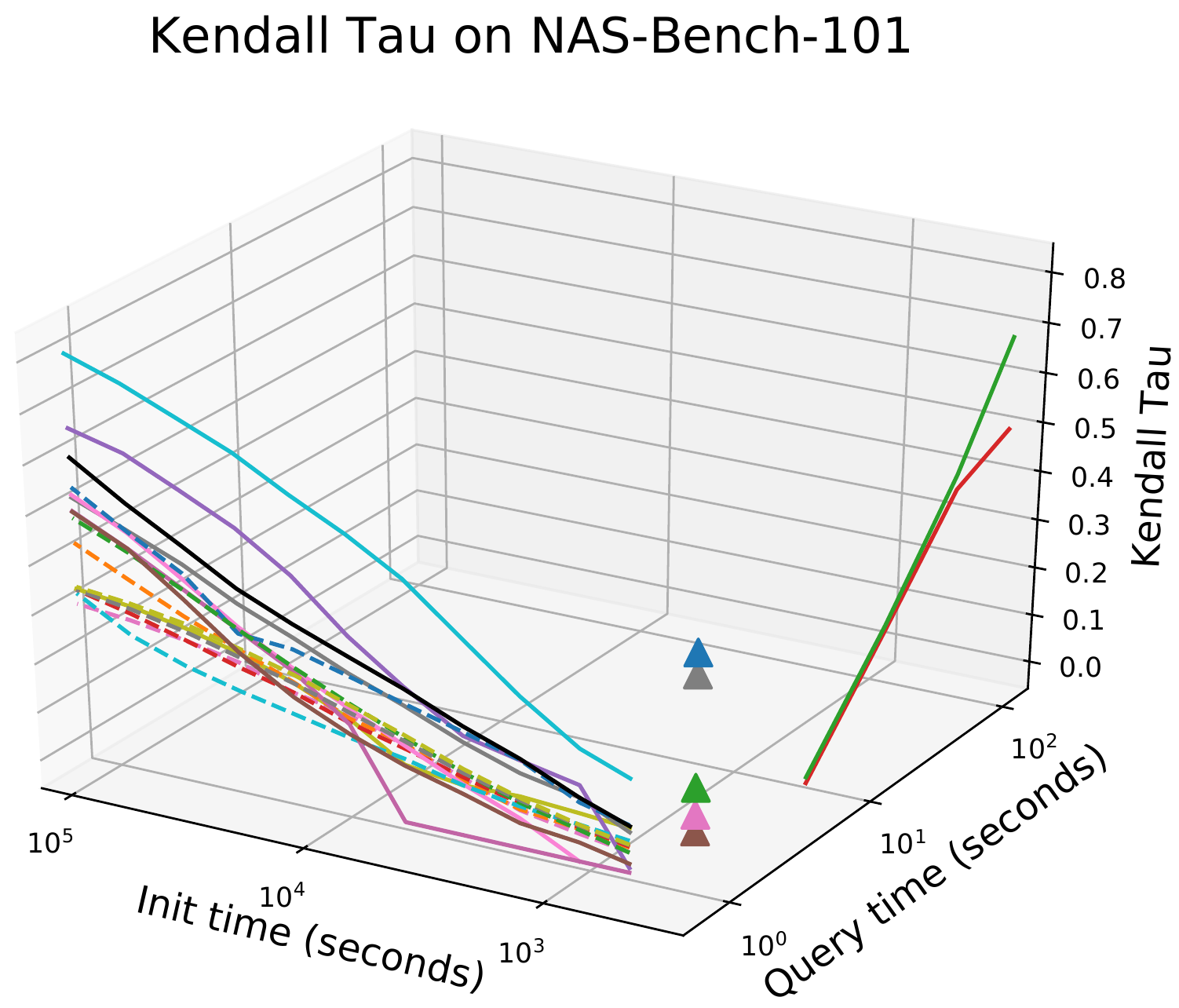}}
\raisebox{0.0\height}{\includegraphics[width=.9\columnwidth]{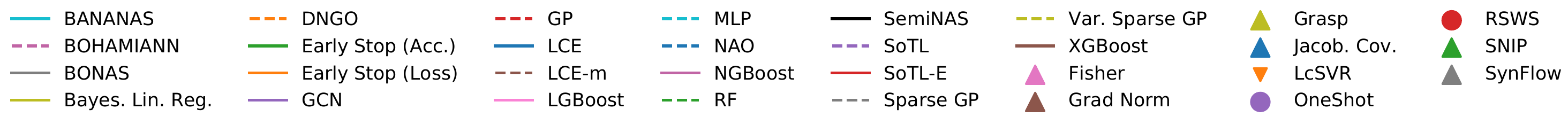}}
\caption{Full results for Kendall Tau with standard deviations shaded.}
\label{fig:app}
\end{figure}

In Figure~\ref{fig:search_spaces}, we plotted the predictors which are Pareto-optimal
for each initialization and query time budget. In this section, we present the complete results,
including 3D plots for all search spaces, as well as separate 
plots for initialization time and query time vs.\ Kendall Tau 
which include the standard deviation of the results from 100 trials. 
See Figure~\ref{fig:app}.
We put in a reasonable effort to implement all 31 predictors for all search spaces, however,
a few of these combinations are omitted. For example, running learning curve methods on
NAS-Bench-101 would require training the architectures from scratch, since NAS-Bench-101 does
not include full learning curve information. 

In general, in Figure~\ref{fig:app} 
we see that the relative performance of the predictors are largely
the same across the three datasets from NAS-Bench-201. However, there are clear
differences across NAS-Bench-201 CIFAR-10, NAS-Bench-101, and DARTS, even though
all three use CIFAR-10 as the dataset. For example, for high initialization time,
BANANAS exhibits the highest rank correlation on NAS-Bench-101 but the worst rank 
correlation on DARTS. As explained in Section~\ref{sec:experiments}, 
this is largely due to the strength of the path encoding
specifically on NAS-Bench-101. However, the path encoding does not scale as well
on larger search spaces such as DARTS.

\begin{figure}
\centering
\raisebox{0.0\height}{\includegraphics[width=.23\columnwidth]{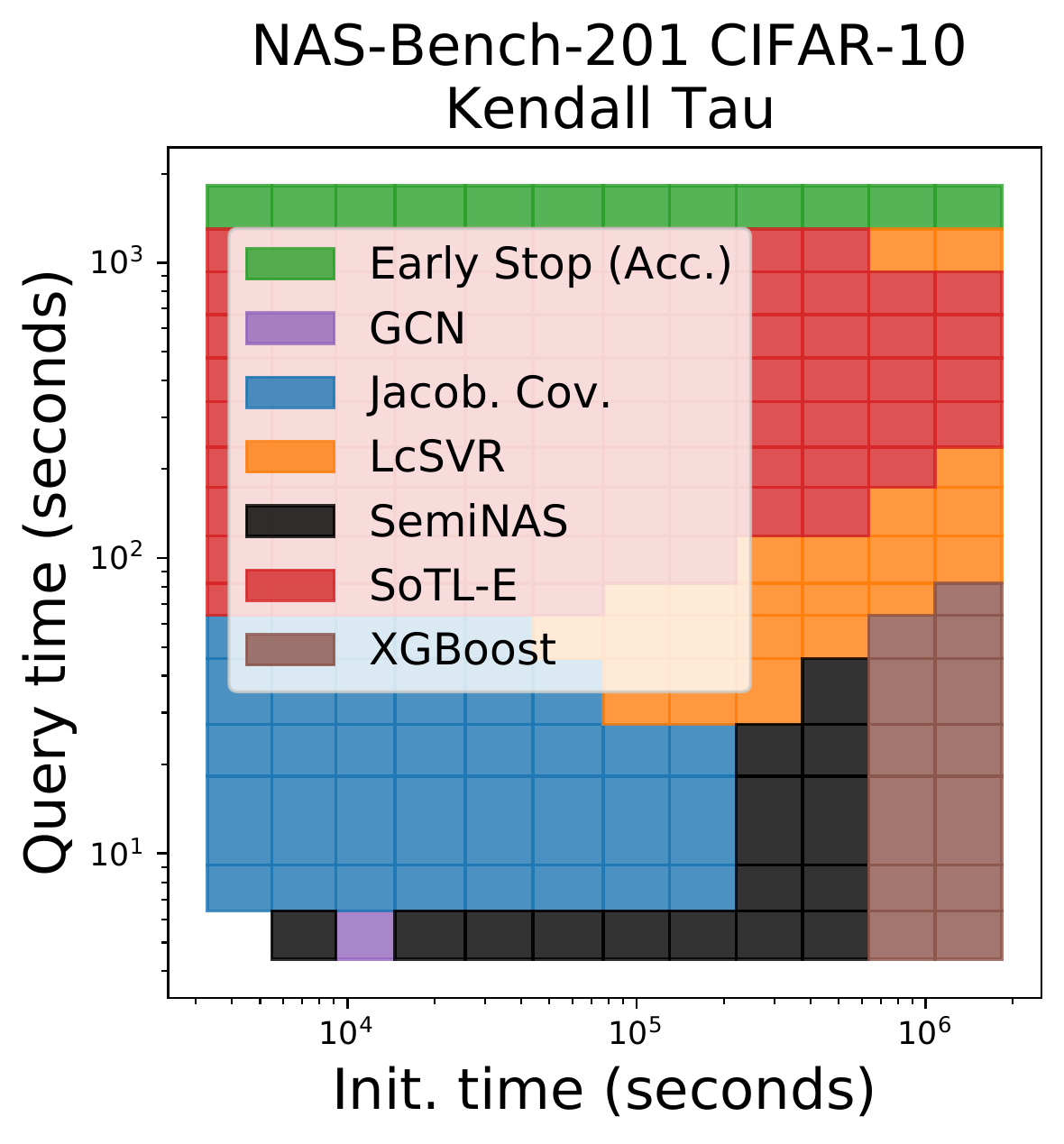}}
\raisebox{0.0\height}{\includegraphics[width=.23\columnwidth]{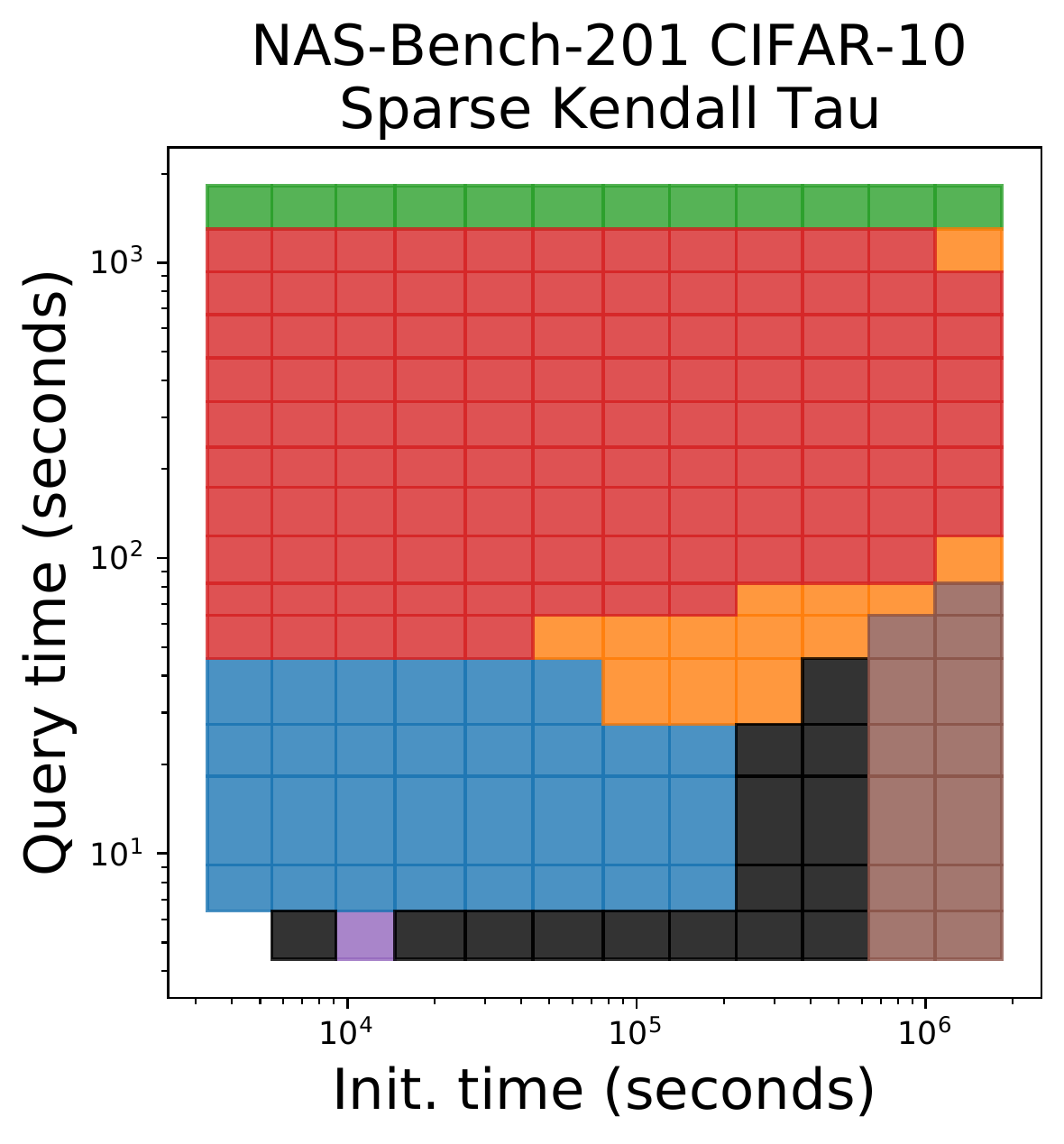}}
\raisebox{0.0\height}{\includegraphics[width=.23\columnwidth]{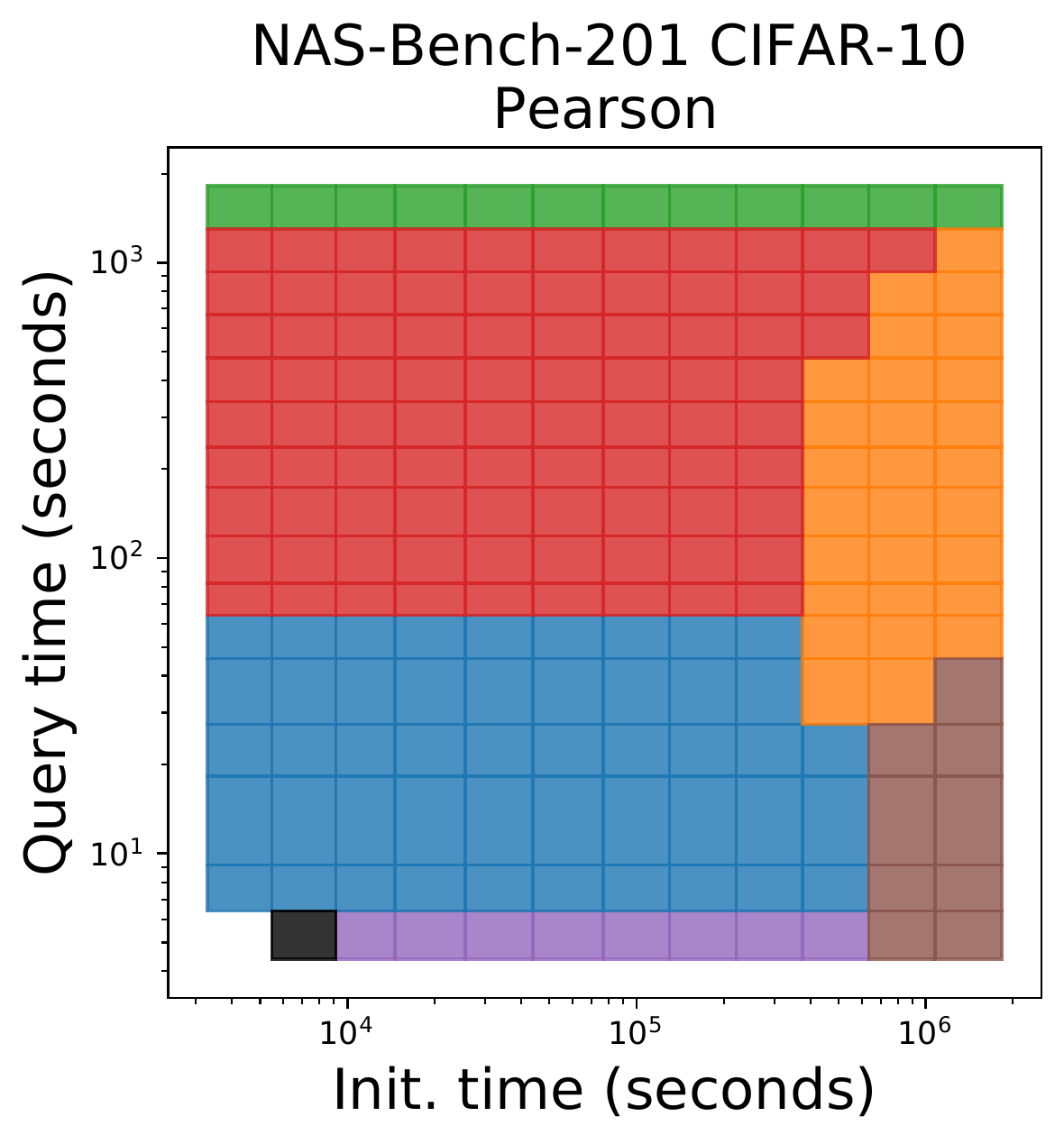}}
\raisebox{0.0\height}{\includegraphics[width=.23\columnwidth]{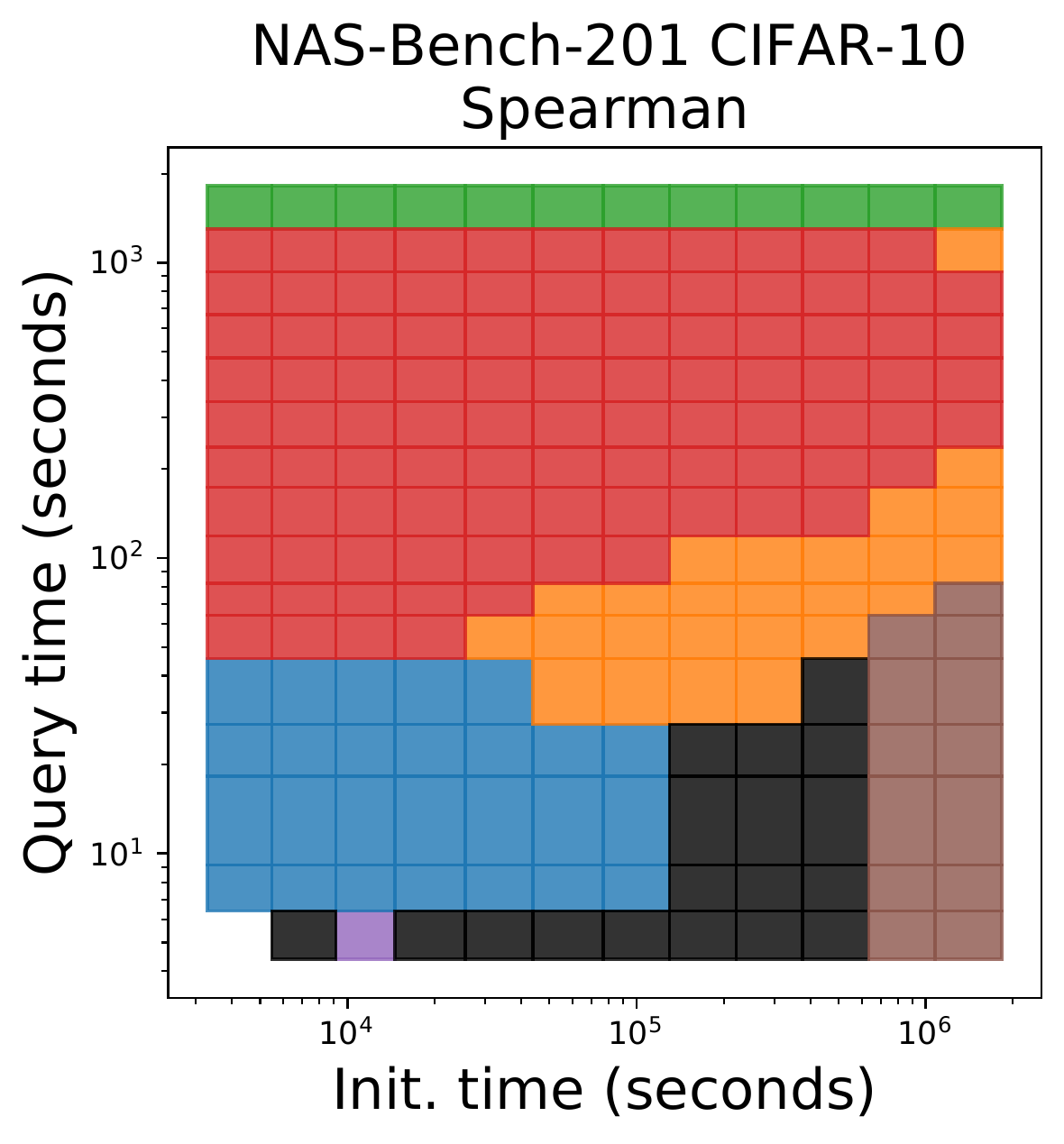}}
\raisebox{0.0\height}{\includegraphics[width=.23\columnwidth]{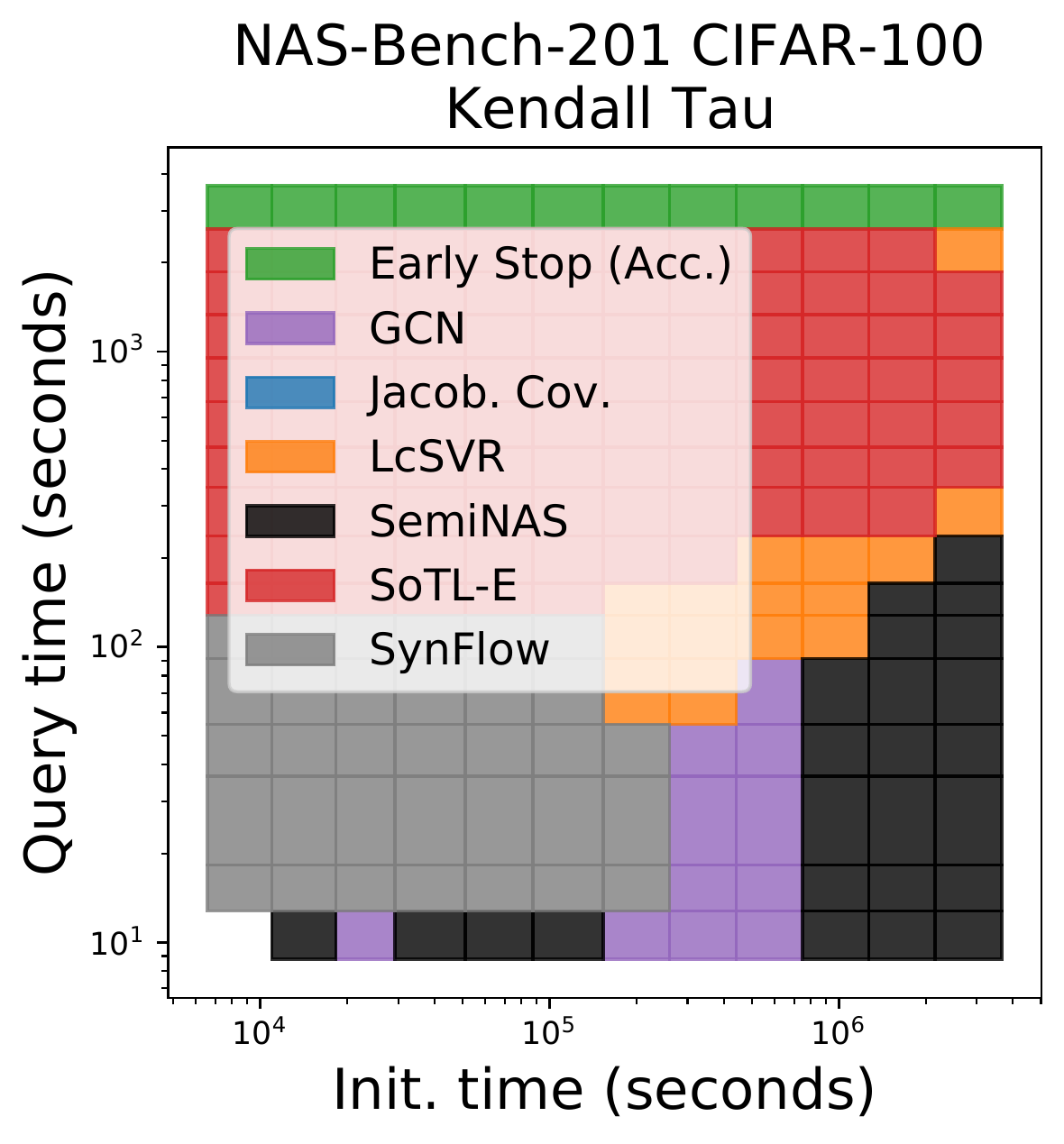}}
\raisebox{0.0\height}{\includegraphics[width=.23\columnwidth]{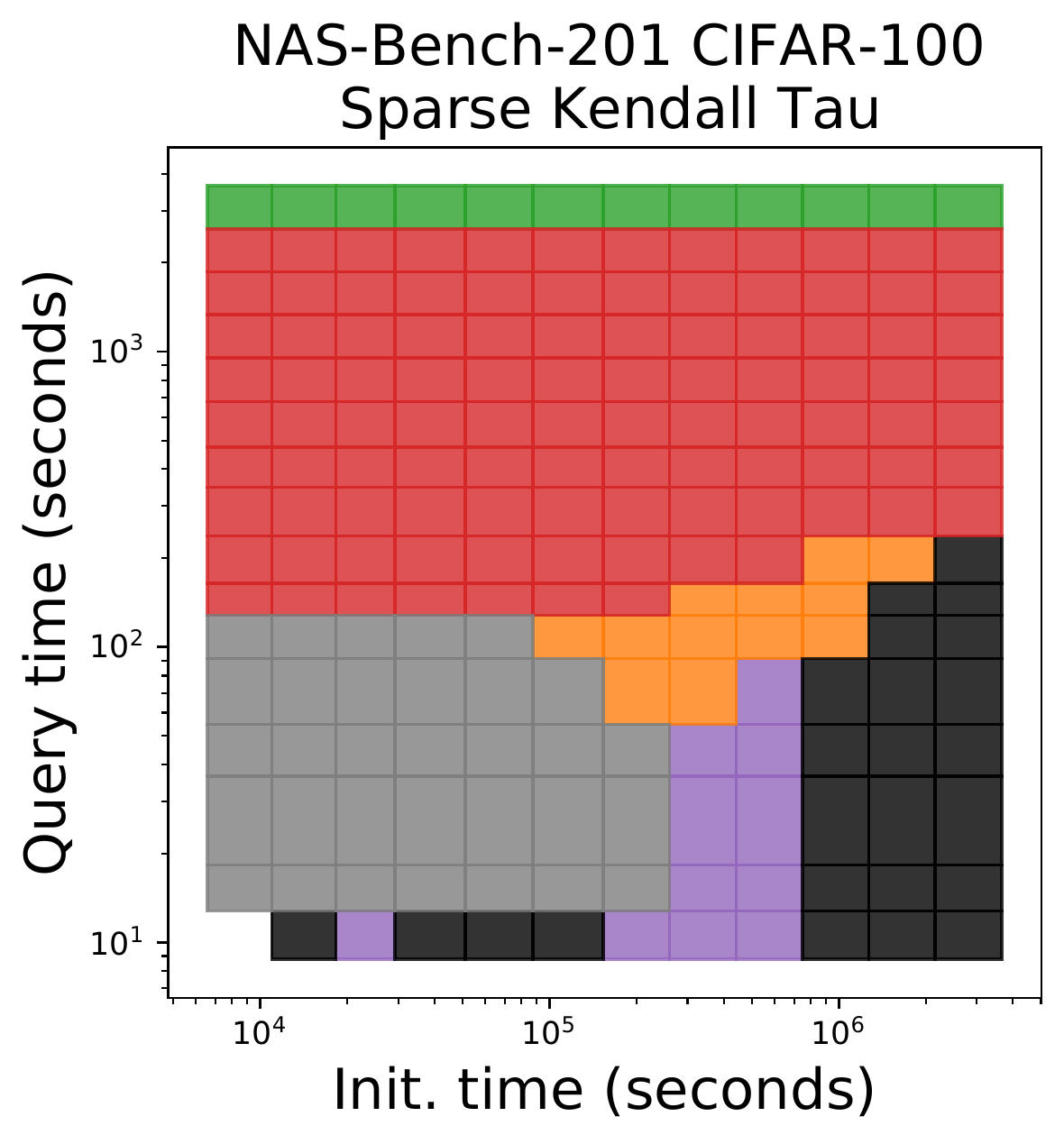}}
\raisebox{0.0\height}{\includegraphics[width=.23\columnwidth]{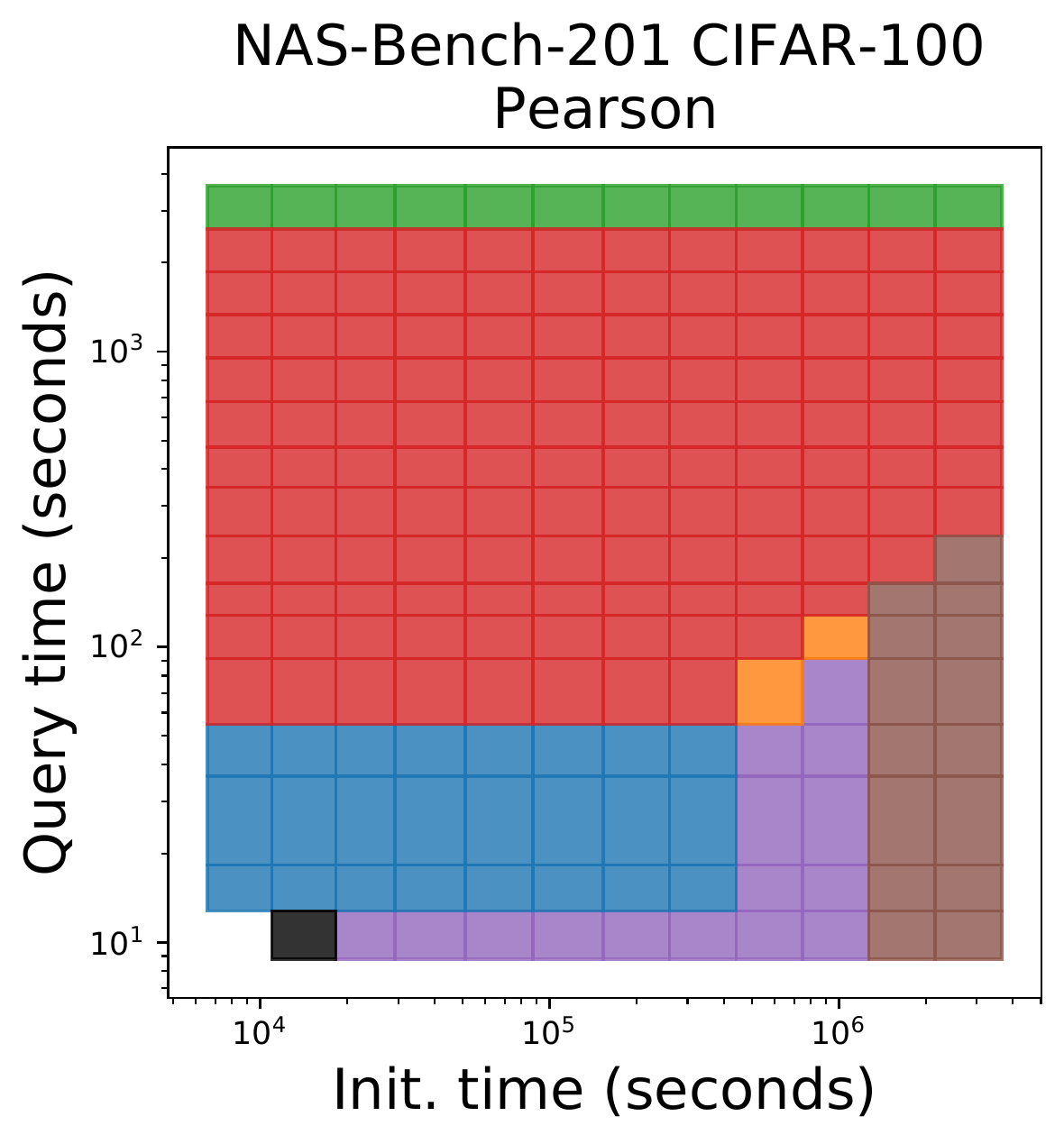}}
\raisebox{0.0\height}{\includegraphics[width=.23\columnwidth]{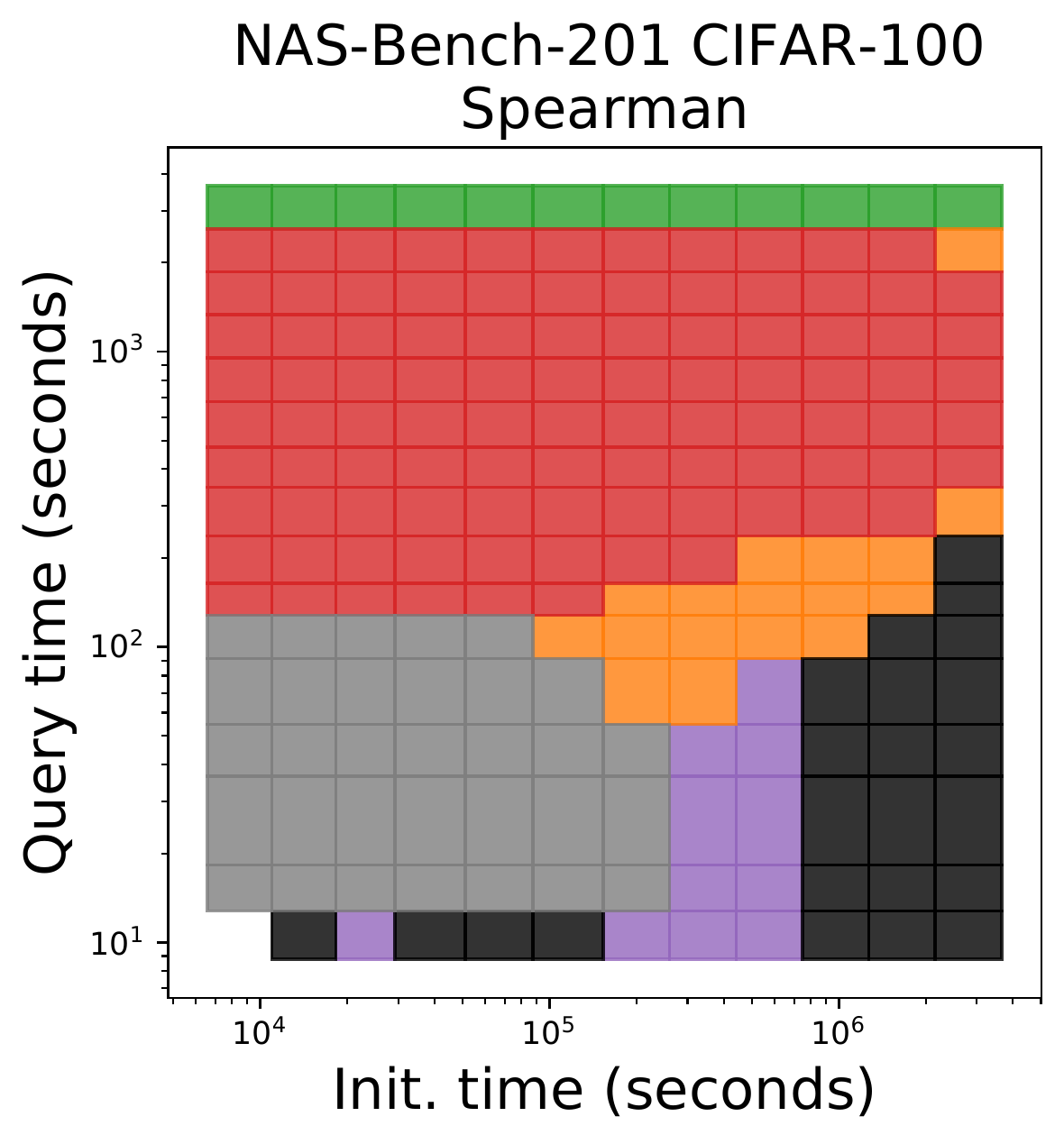}}
\raisebox{0.0\height}{\includegraphics[width=.23\columnwidth]{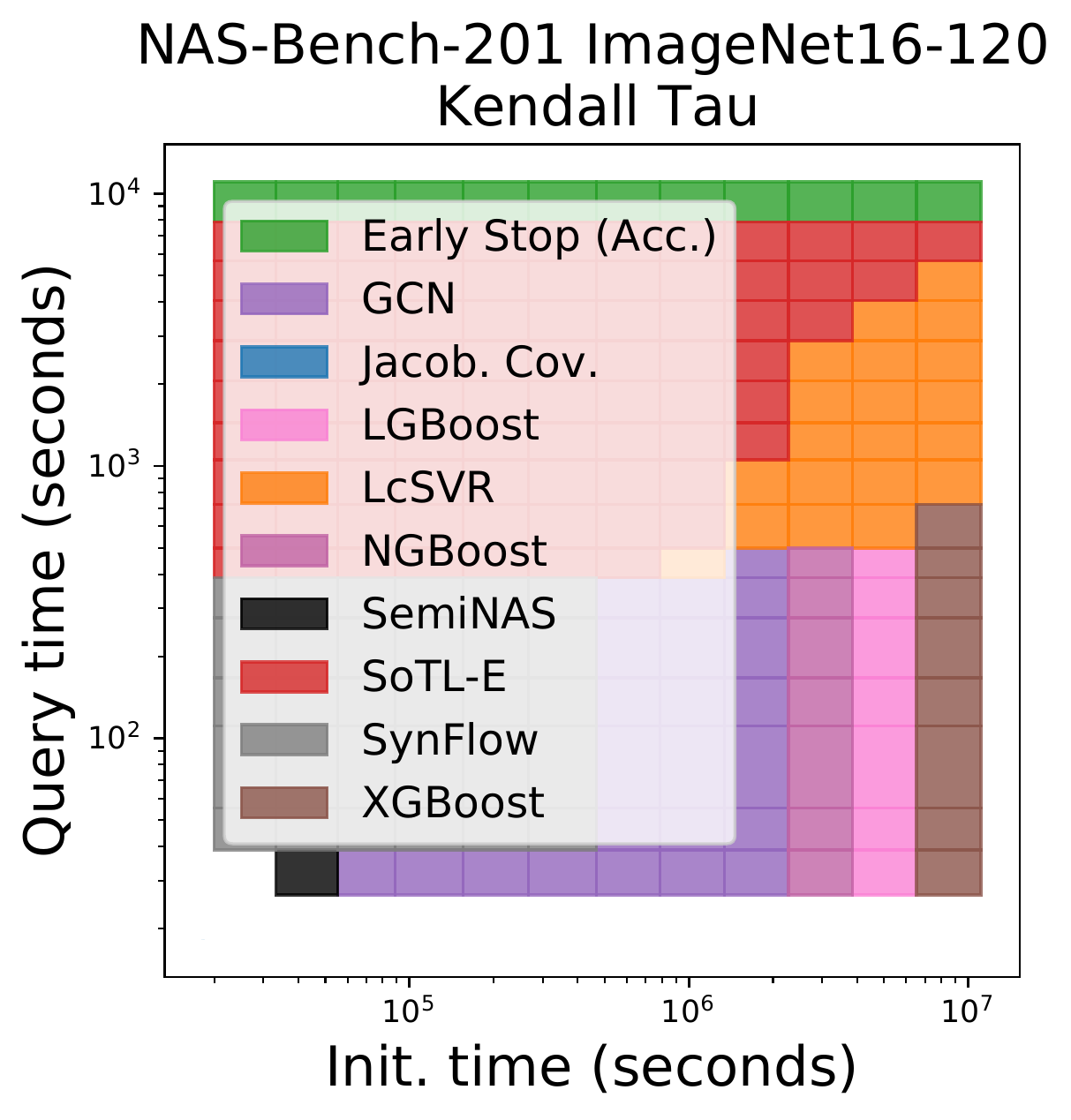}}
\raisebox{0.0\height}{\includegraphics[width=.23\columnwidth]{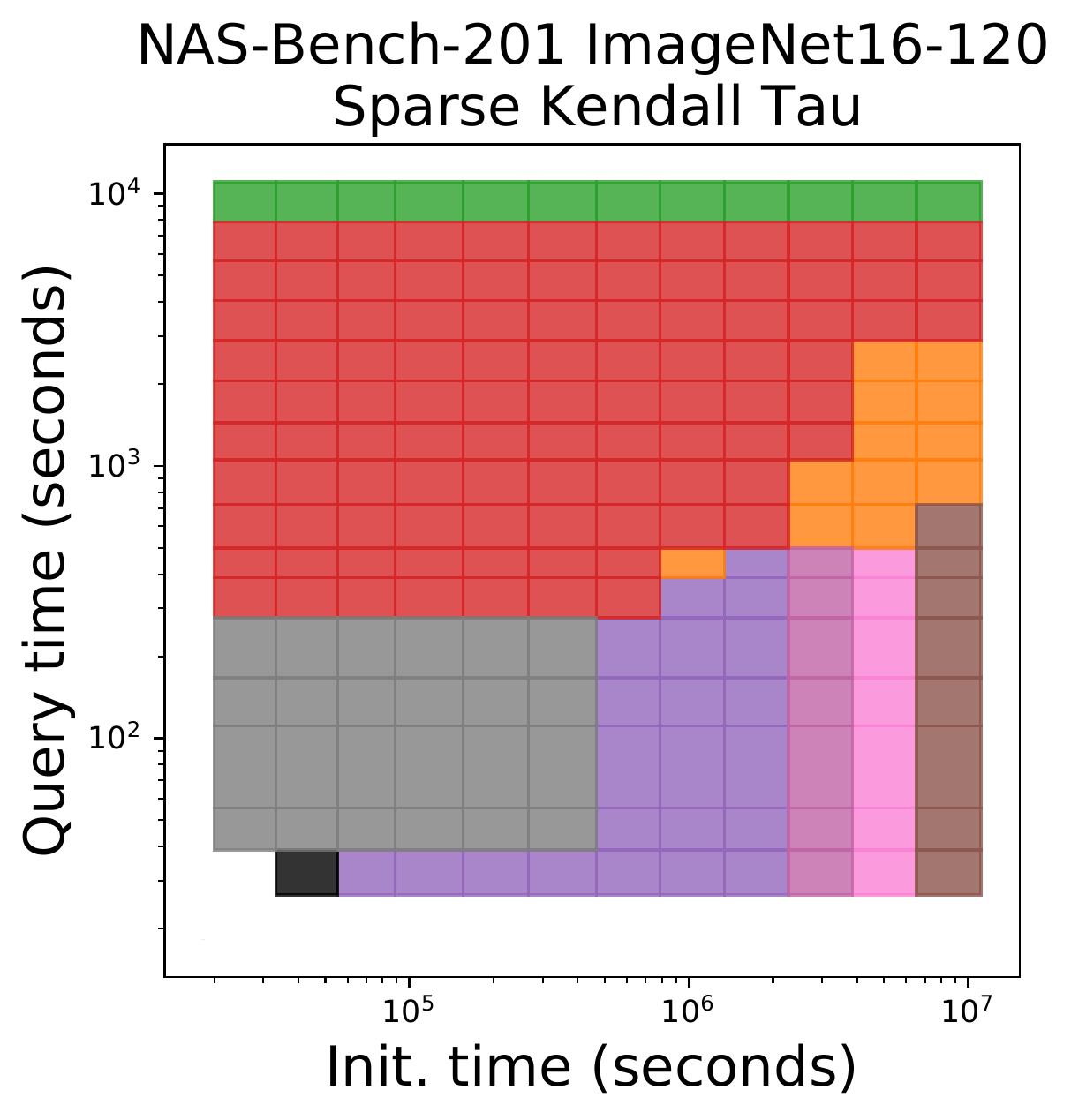}}
\raisebox{0.0\height}{\includegraphics[width=.23\columnwidth]{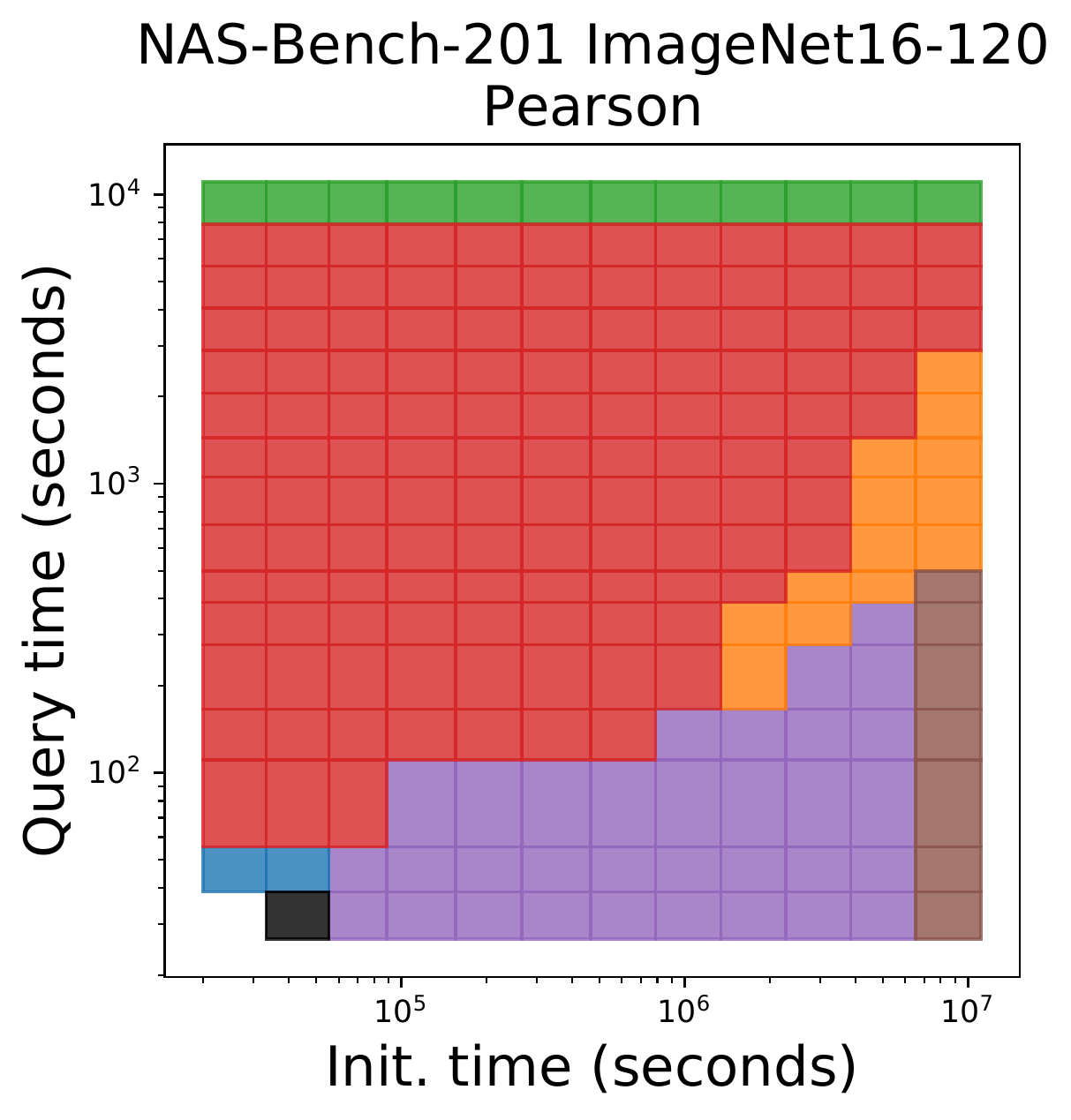}}
\raisebox{0.0\height}{\includegraphics[width=.23\columnwidth]{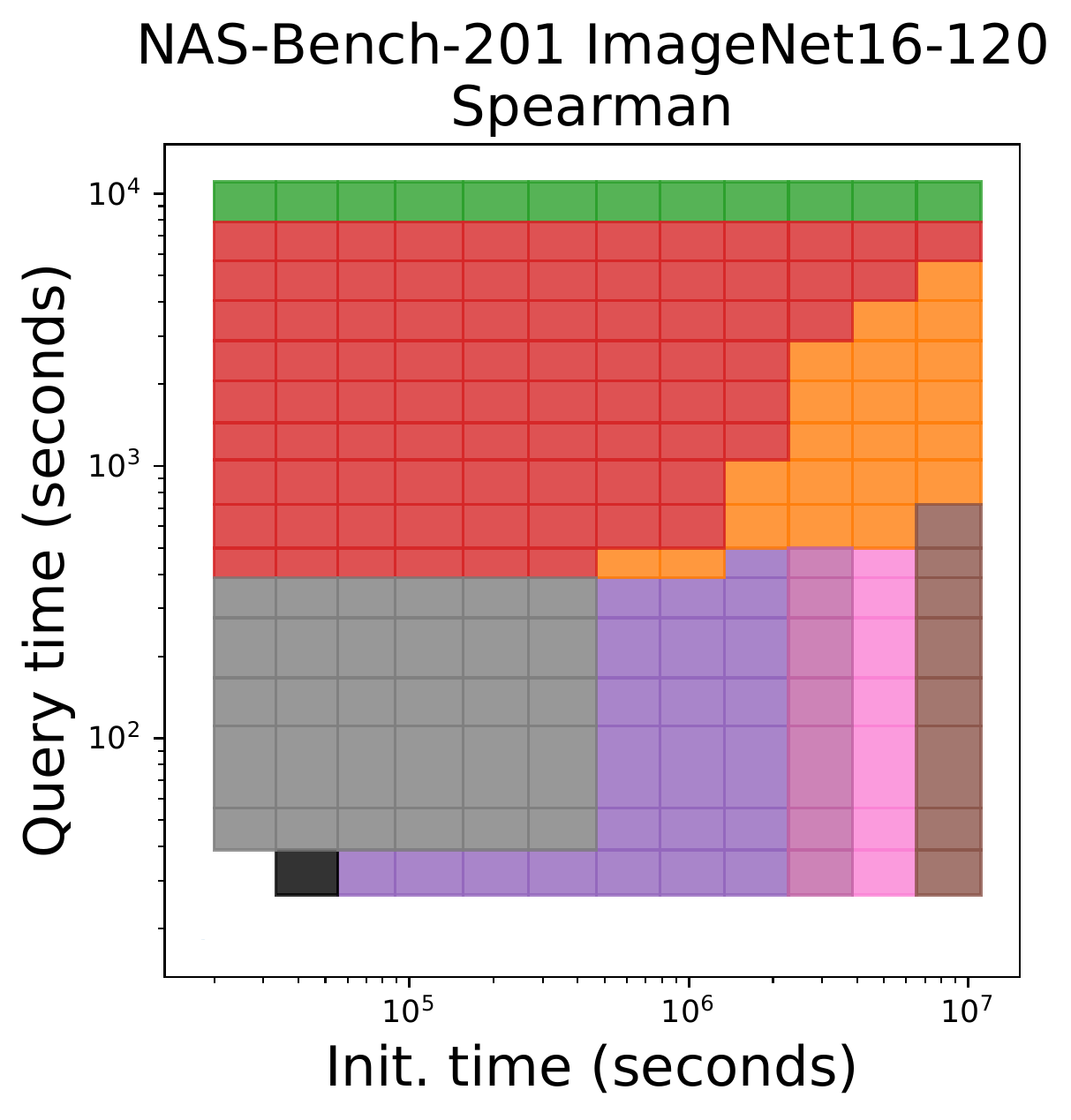}}
\raisebox{0.0\height}{\includegraphics[width=.23\columnwidth]{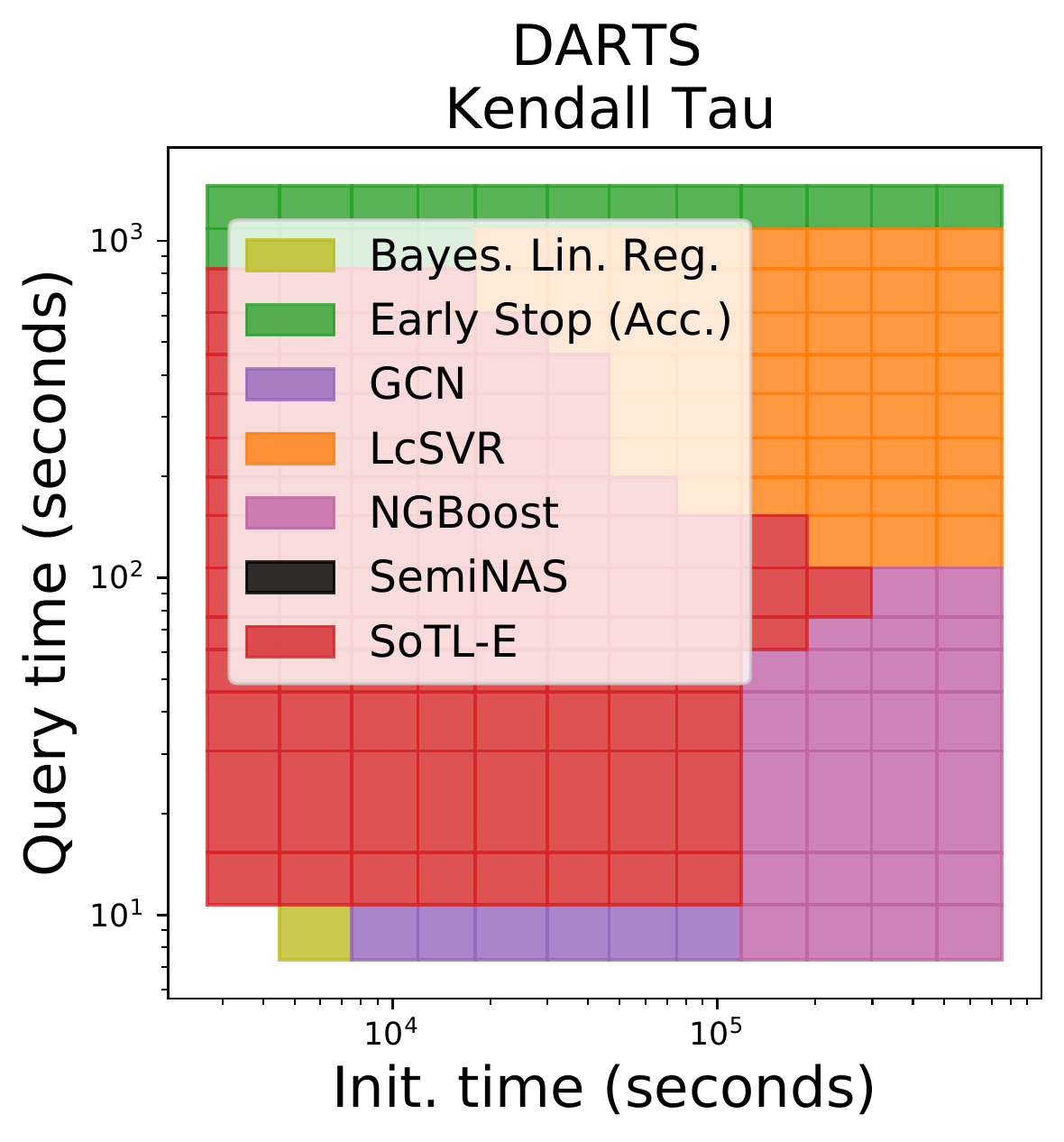}}
\raisebox{0.0\height}{\includegraphics[width=.23\columnwidth]{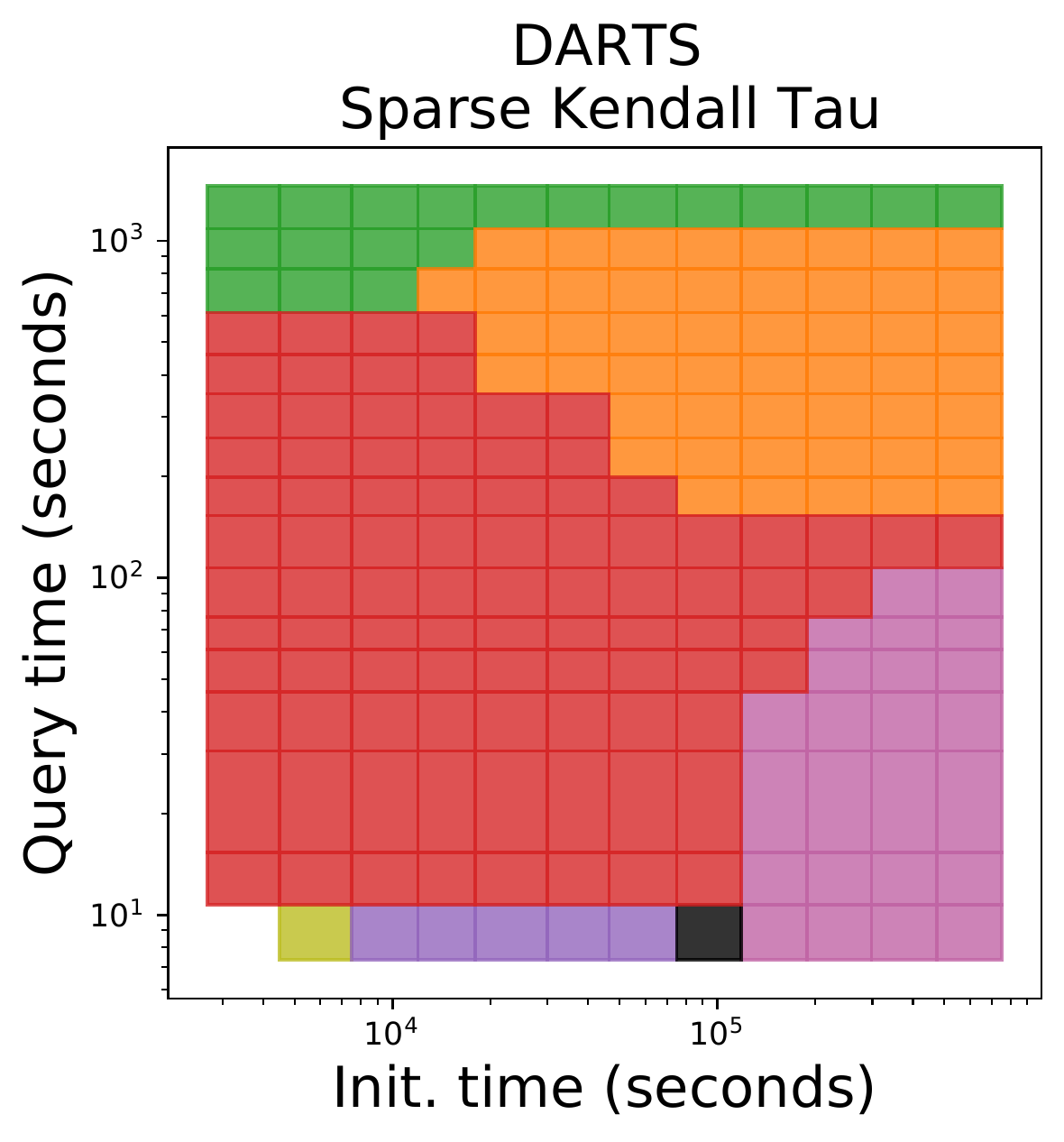}}
\raisebox{0.0\height}{\includegraphics[width=.23\columnwidth]{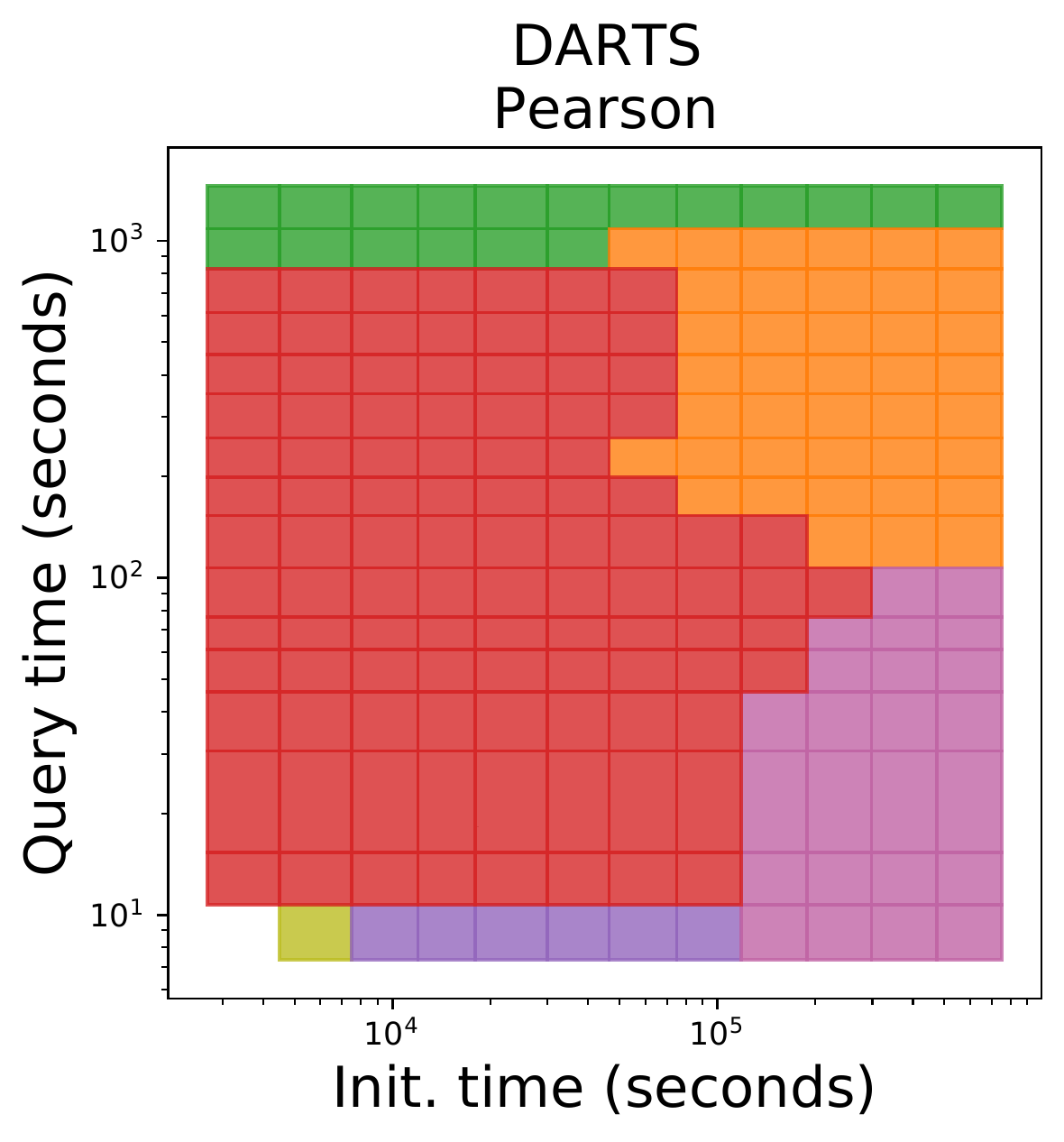}}
\raisebox{0.0\height}{\includegraphics[width=.23\columnwidth]{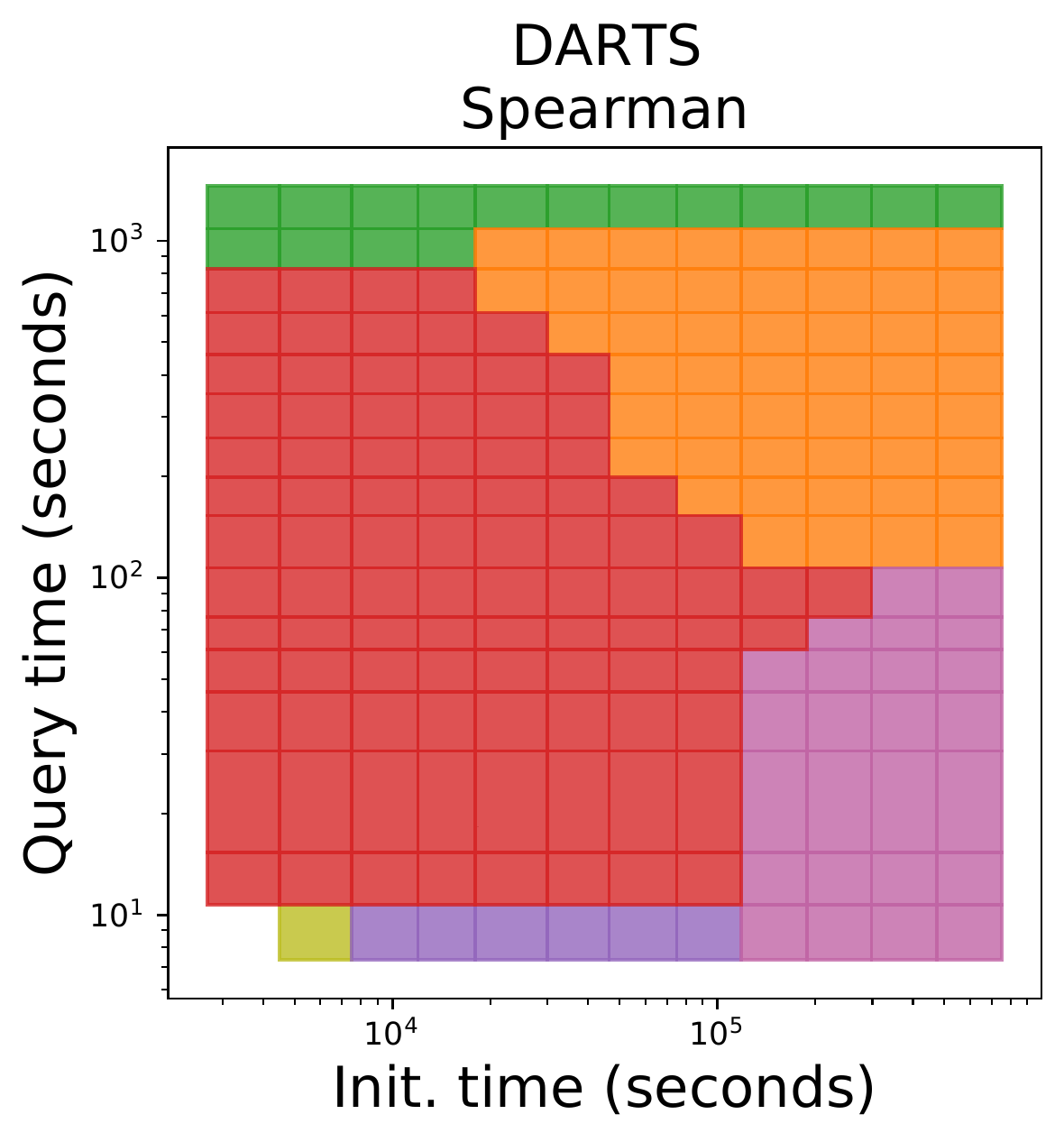}}
\raisebox{0.0\height}{\includegraphics[width=.23\columnwidth]{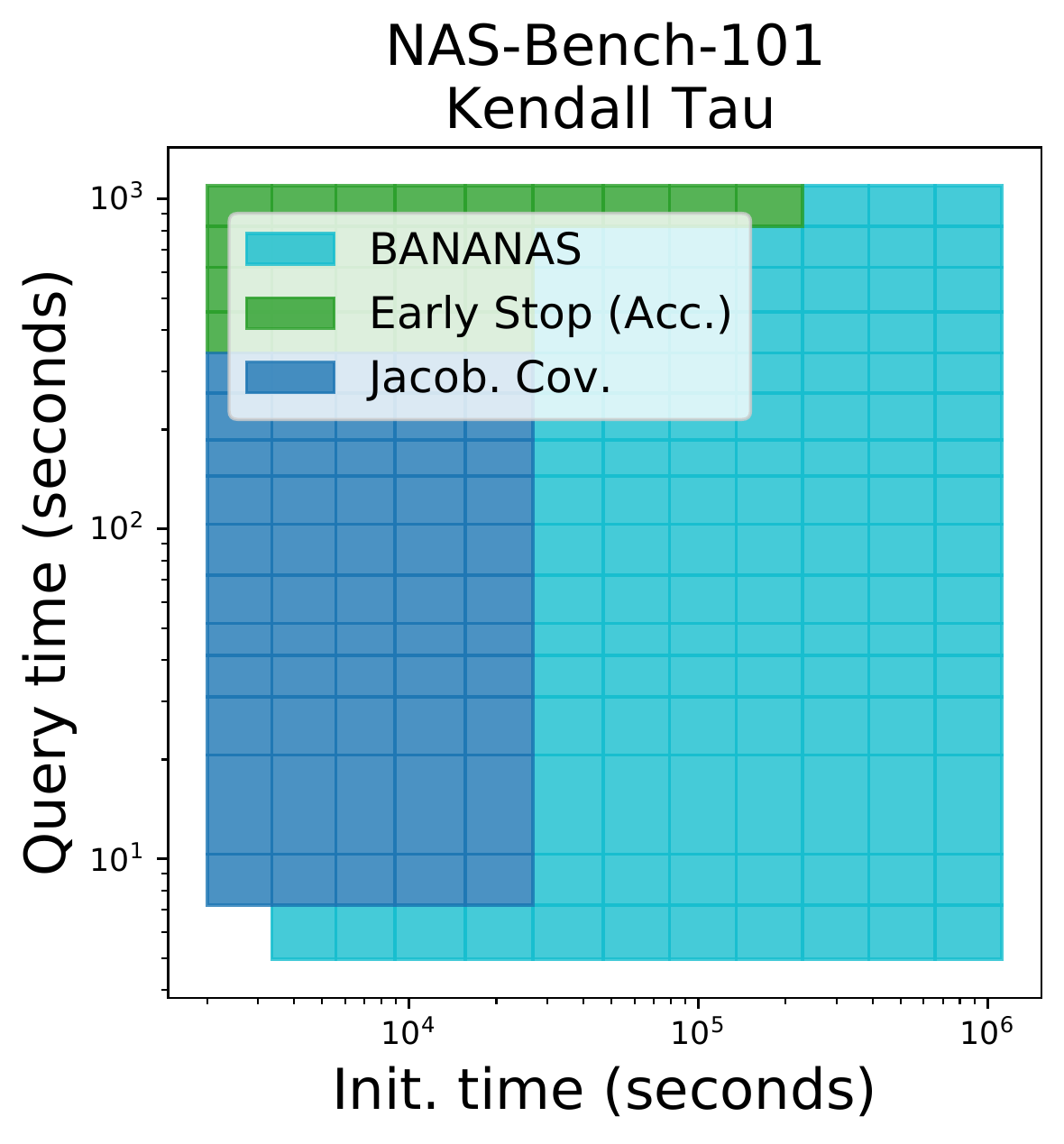}}
\raisebox{0.0\height}{\includegraphics[width=.23\columnwidth]{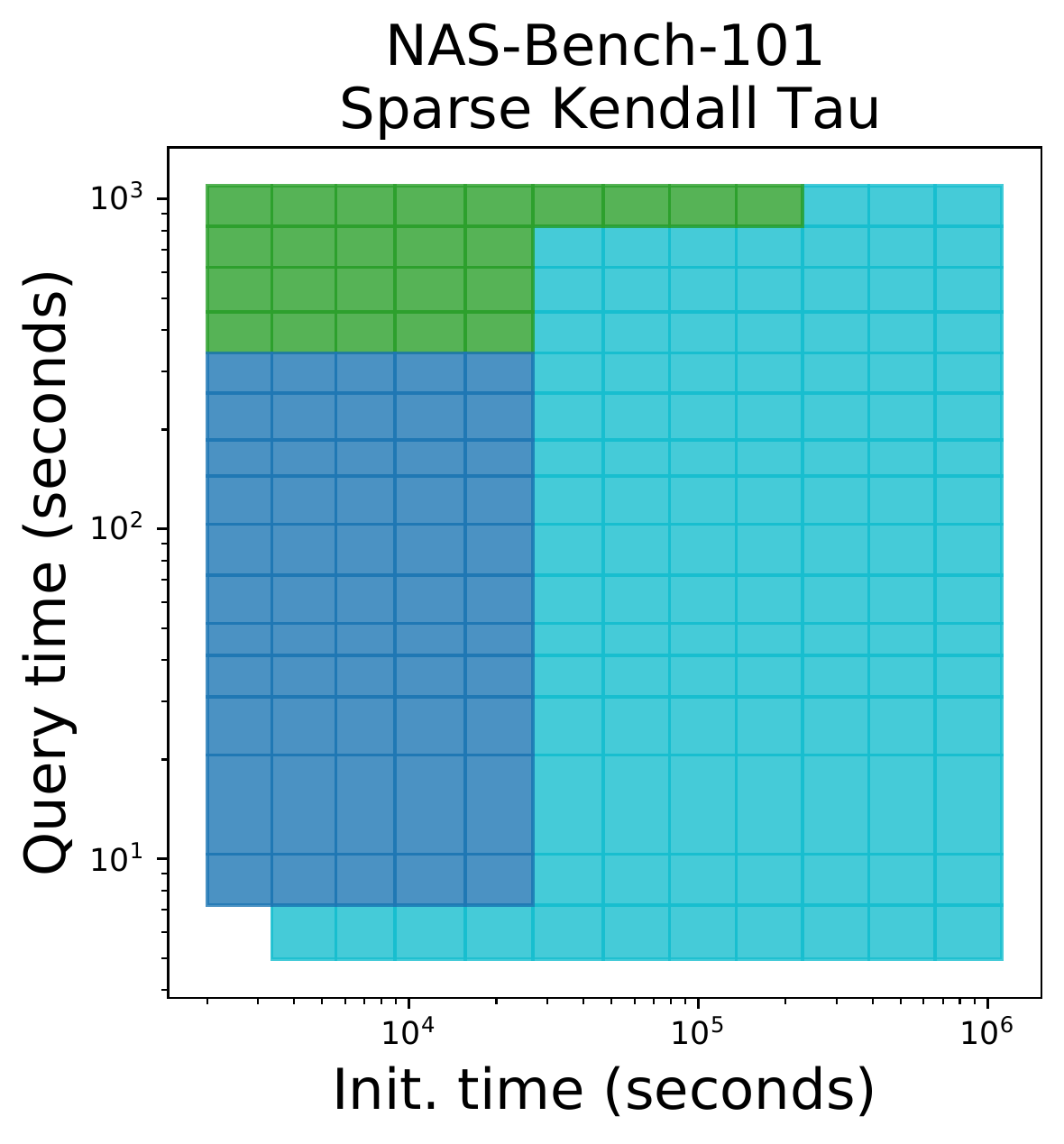}}
\raisebox{0.0\height}{\includegraphics[width=.23\columnwidth]{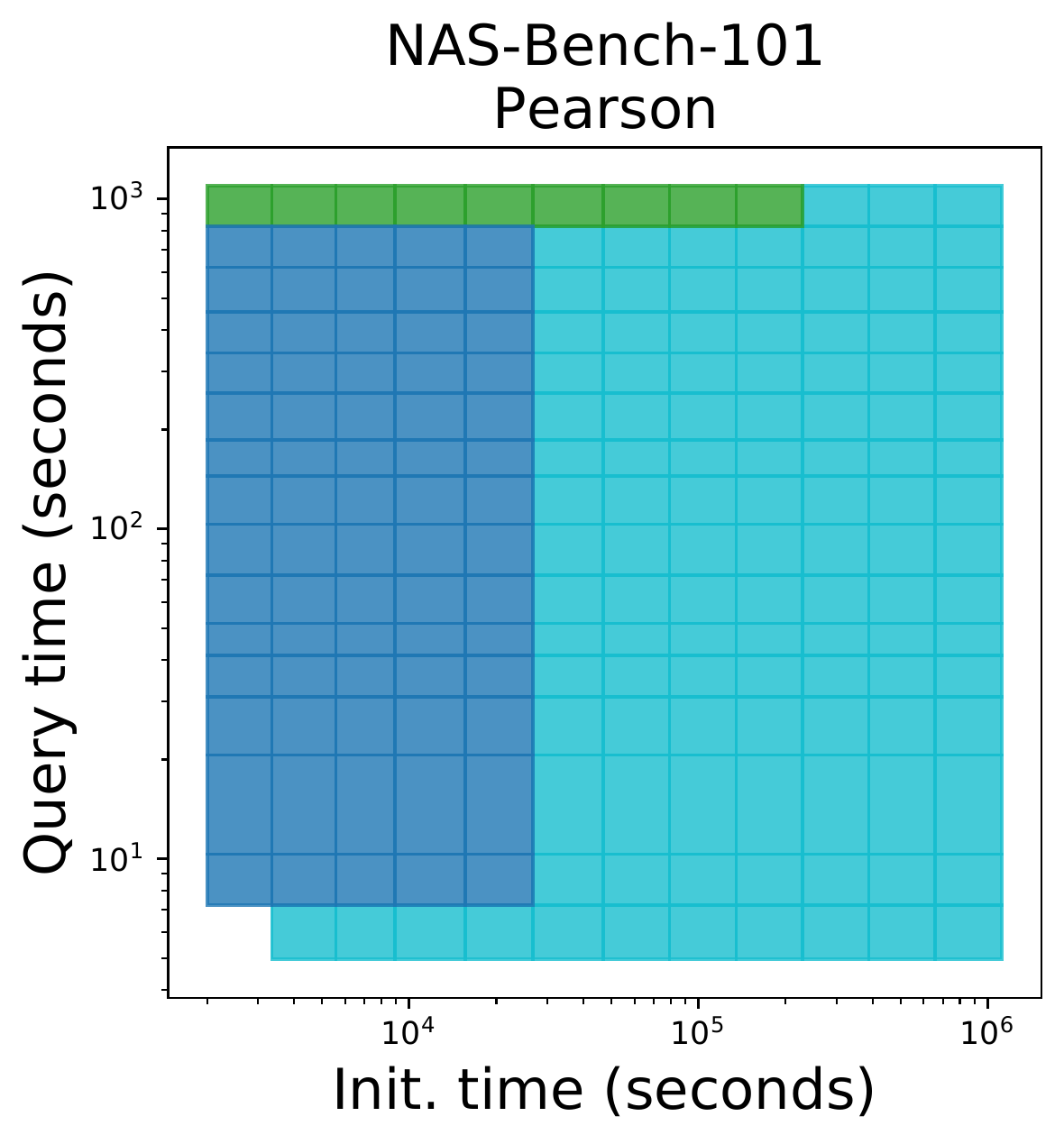}}
\raisebox{0.0\height}{\includegraphics[width=.23\columnwidth]{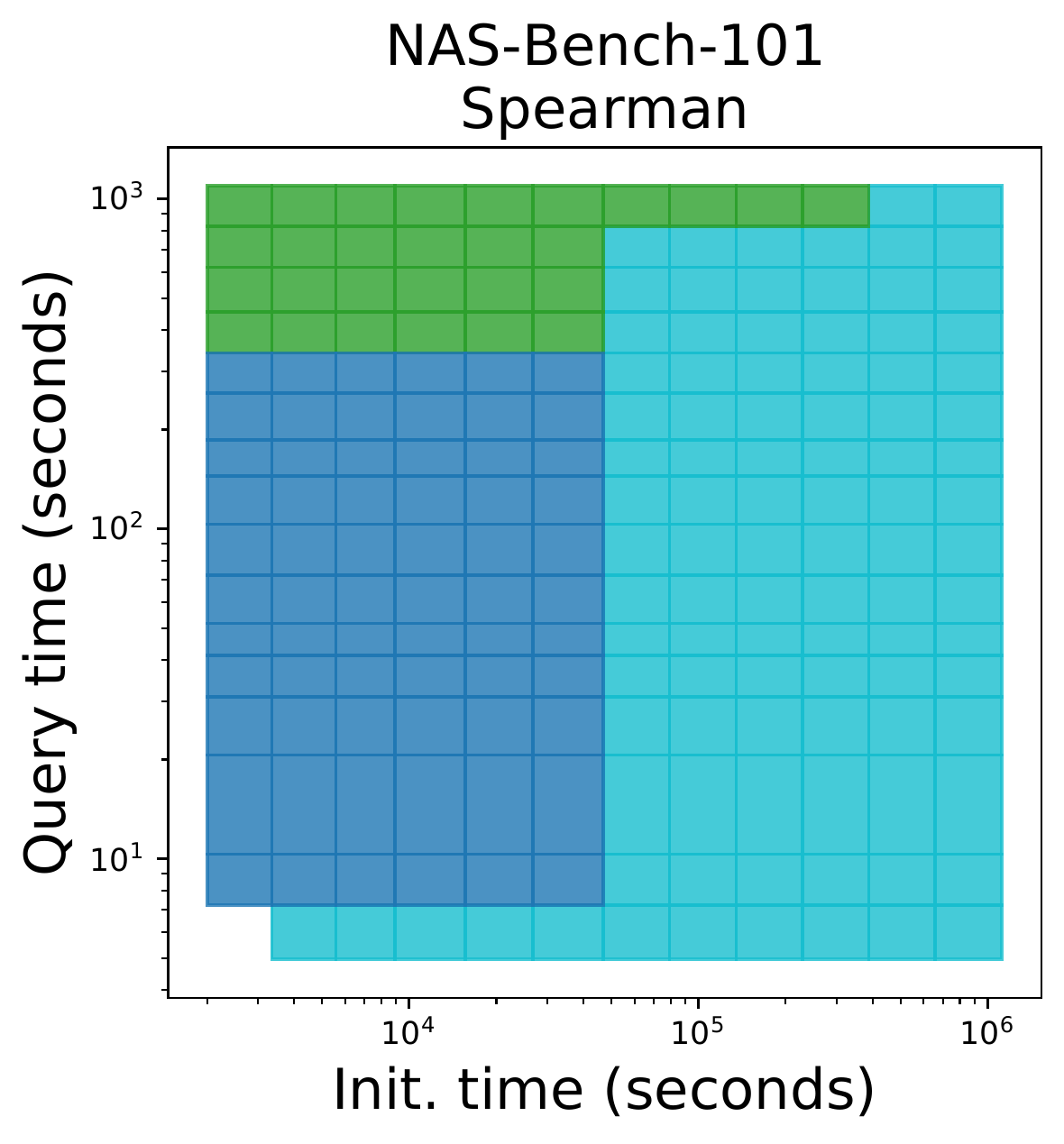}}
\caption{The performance predictors with the highest metrics for all initialization time and query
time budgets and search spaces. The first column is repeated from Figure~\ref{fig:search_spaces}.
}
\label{fig:paretos}
\end{figure}

In Figure~\ref{fig:paretos}, we plot the predictors which are Pareto-optimal for each
initialization and query time budget, for three metrics: Pearson correlation, 
Spearman rank correlation, and sparse Kendall Tau. We also include Kendall Tau 
for completeness (which we had already presented in Figure~\ref{fig:search_spaces}).

\begin{figure}
\centering
\raisebox{0.0\height}{\includegraphics[width=.29\columnwidth]{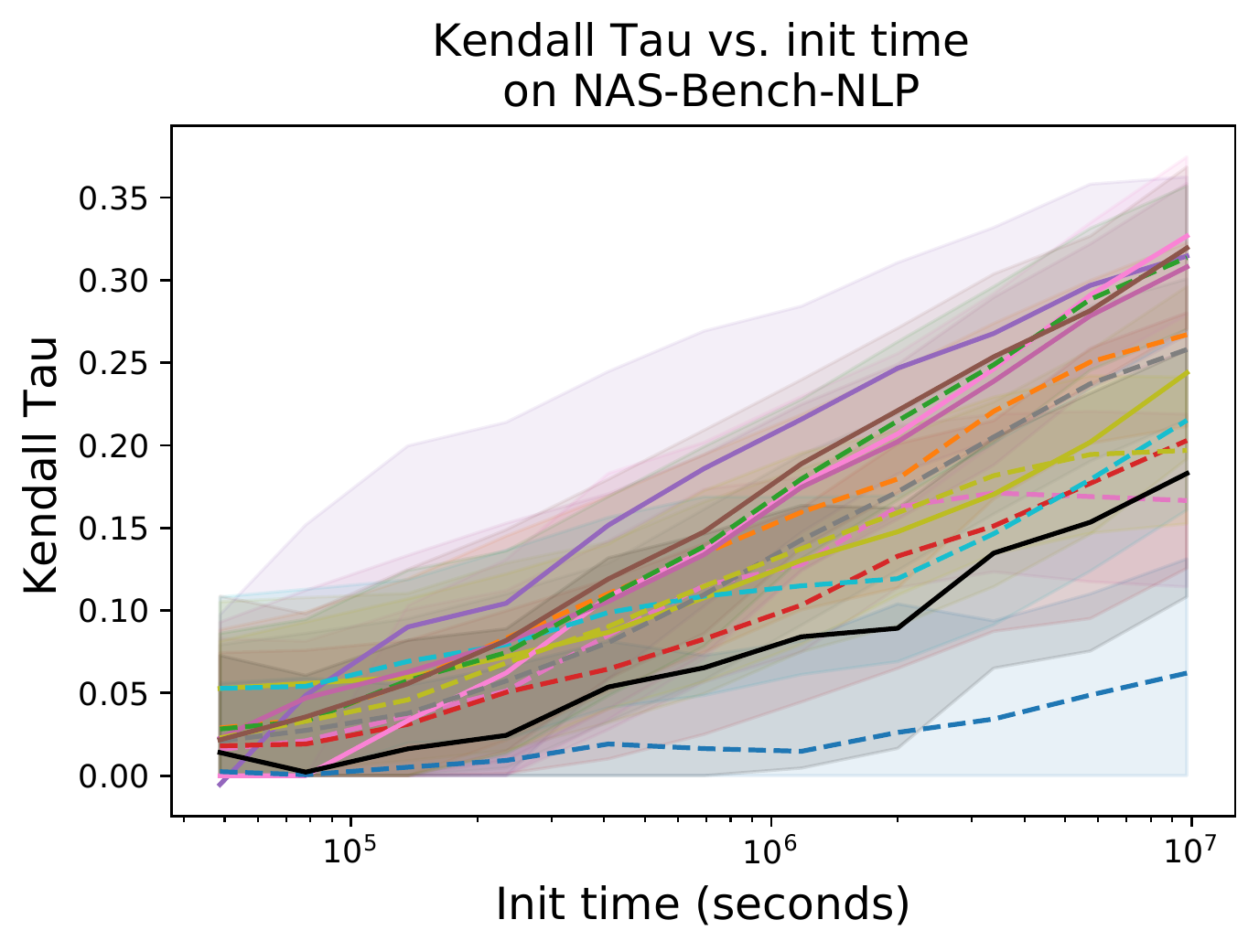}}
\raisebox{0.0\height}{\includegraphics[width=.29\columnwidth]{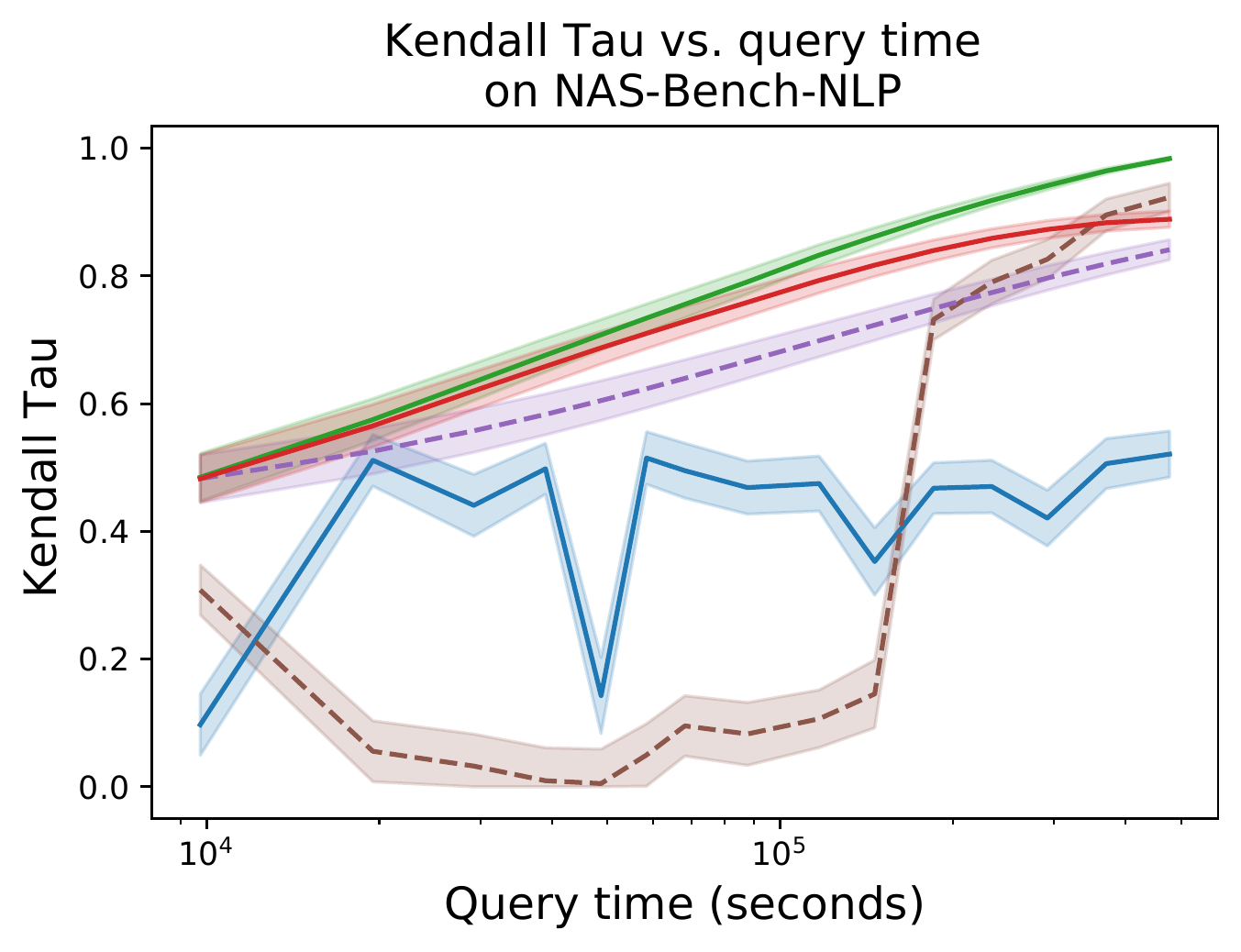}}
\raisebox{0.0\height}{\includegraphics[width=.28\columnwidth]{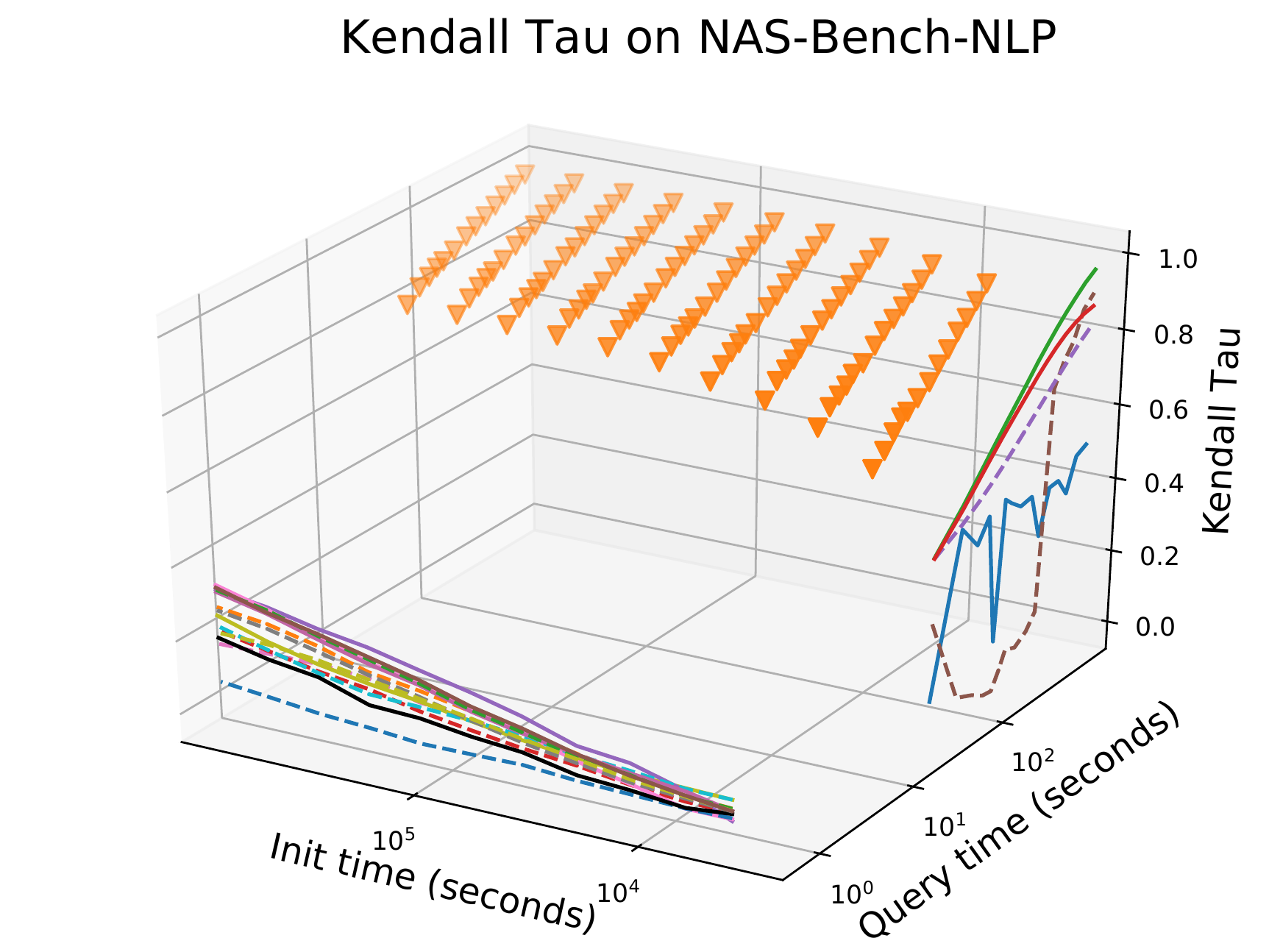}}
\raisebox{0.0\height}{\includegraphics[width=.9\columnwidth]{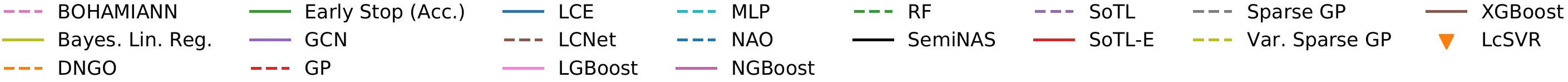}}
\raisebox{0.0\height}{\includegraphics[width=.23\columnwidth]{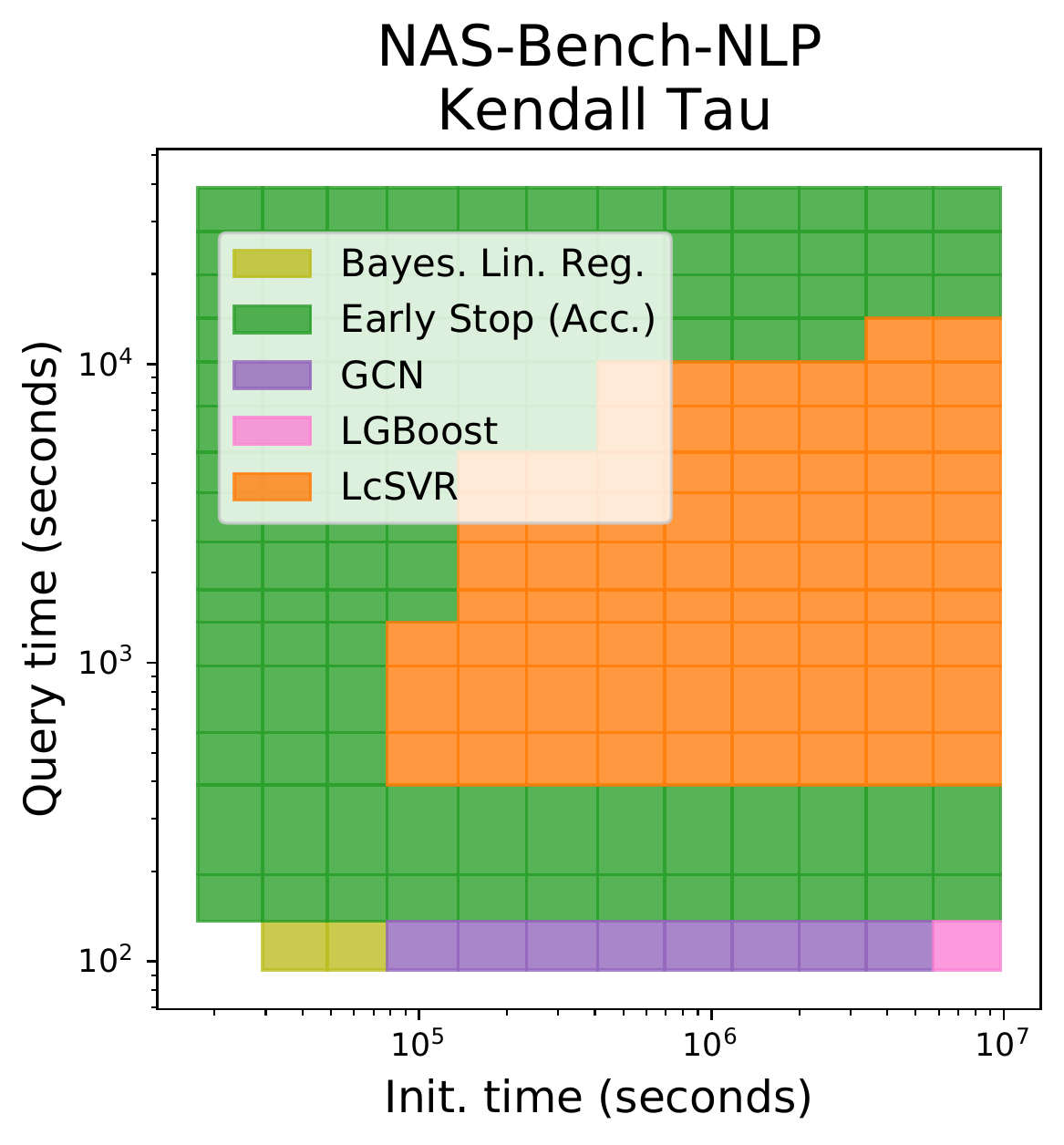}}
\raisebox{0.0\height}{\includegraphics[width=.23\columnwidth]{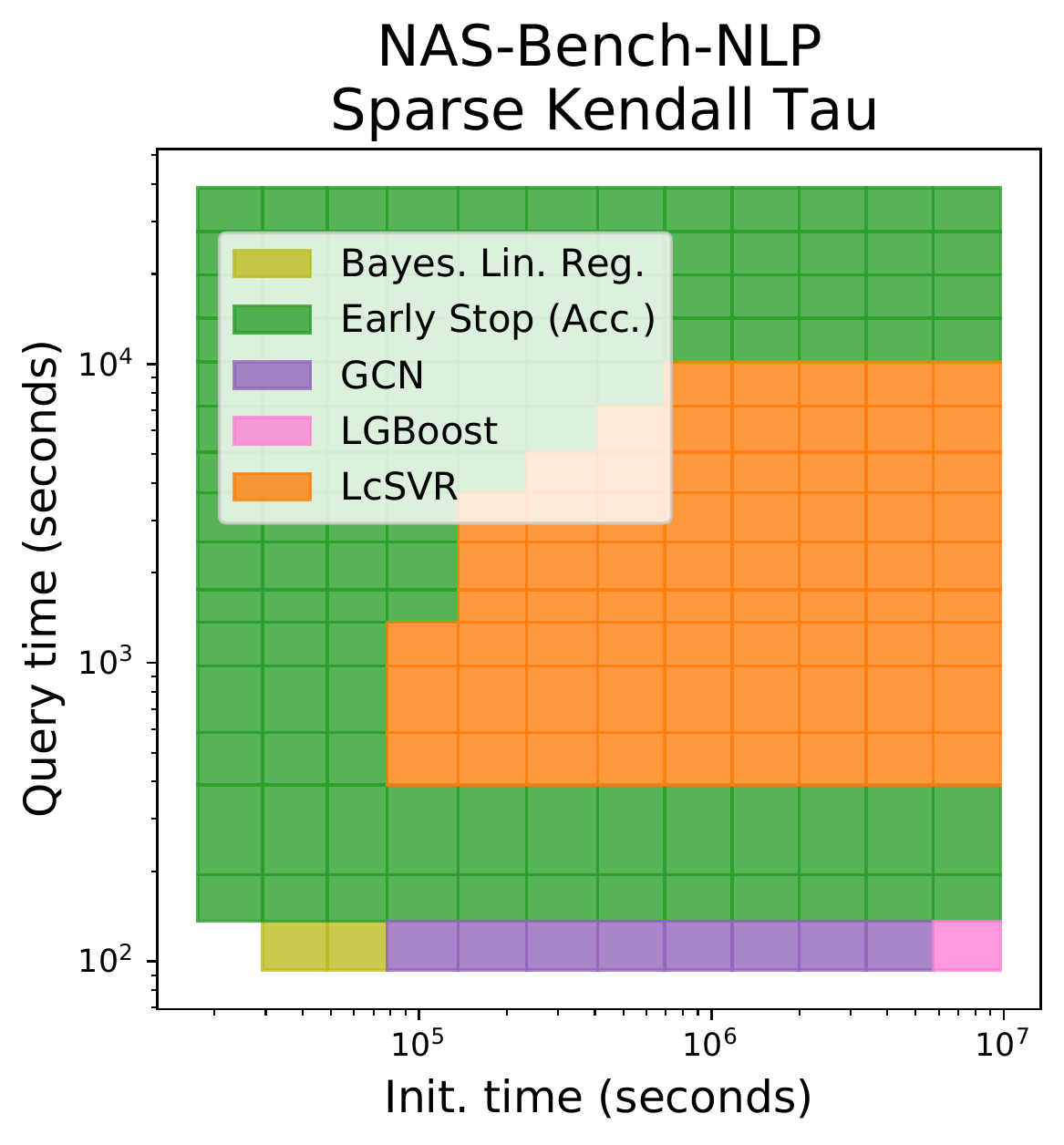}}
\raisebox{0.0\height}{\includegraphics[width=.23\columnwidth]{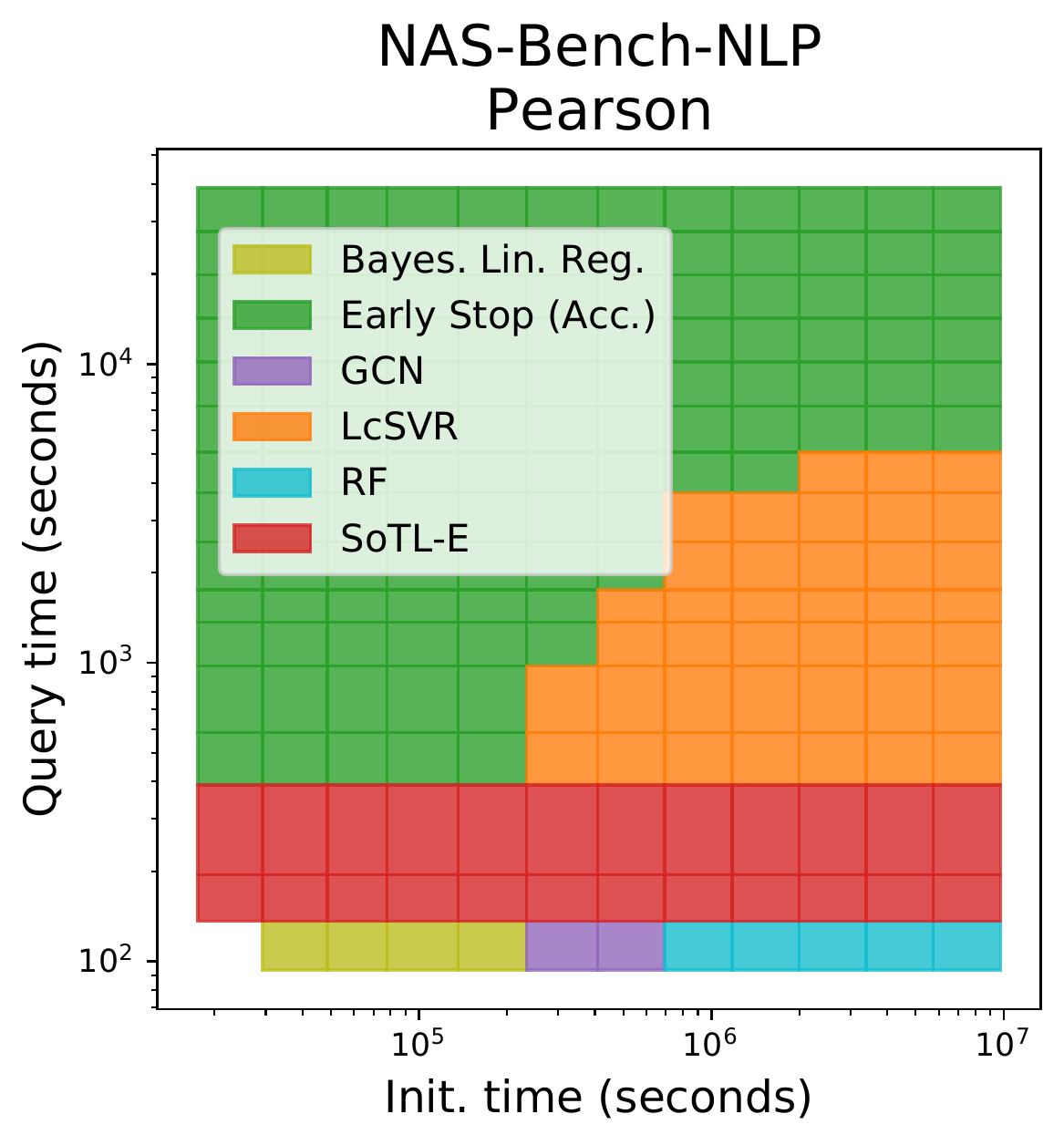}}
\raisebox{0.0\height}{\includegraphics[width=.23\columnwidth]{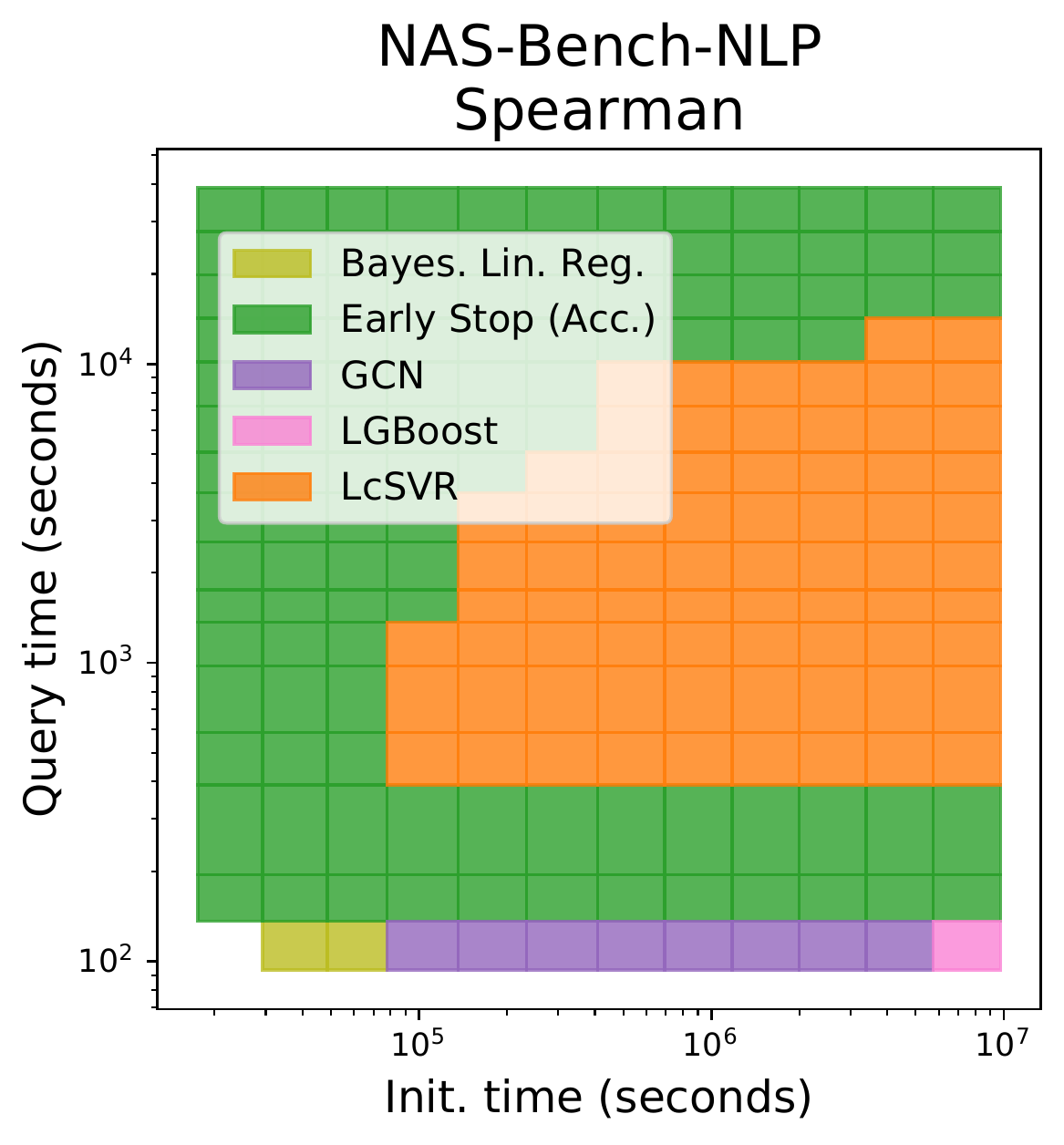}}
\caption{Full results for NAS-Bench-NLP with standard deviations shaded (top row).
The performance predictors with the highest metrics for all initialization time
and query time budgets on NAS-Bench-NLP (bottom row).}
\label{fig:nlp}
\end{figure}

Since Figures~\ref{fig:app} and~\ref{fig:paretos} take up one page each already,
we plot the results for the final search space, NAS-Bench-NLP, in Figure~\ref{fig:nlp}.
We see largely the same trends on this repository as well, with a few differences.
Since the NAS-Bench-NLP code is written in a much earlier version of
PyTorch (1.1), we were unable to implement the zero-cost predictors in
NAS-Bench-NLP. Furthermore, the SOTL and SOTL-E predictors do not perform
as well as Early Stop (Acc.) for NAS-Bench-NLP.
We see that LcSVR performs particularly well on NAS-Bench-NLP (as was also
the case with DARTS). This may be because the large size of the search space
gives a bigger benefit for hybrid methods which include LCE components, 
over model-based methods alone.

\begin{figure}
\centering
\includegraphics[width=.5\columnwidth]{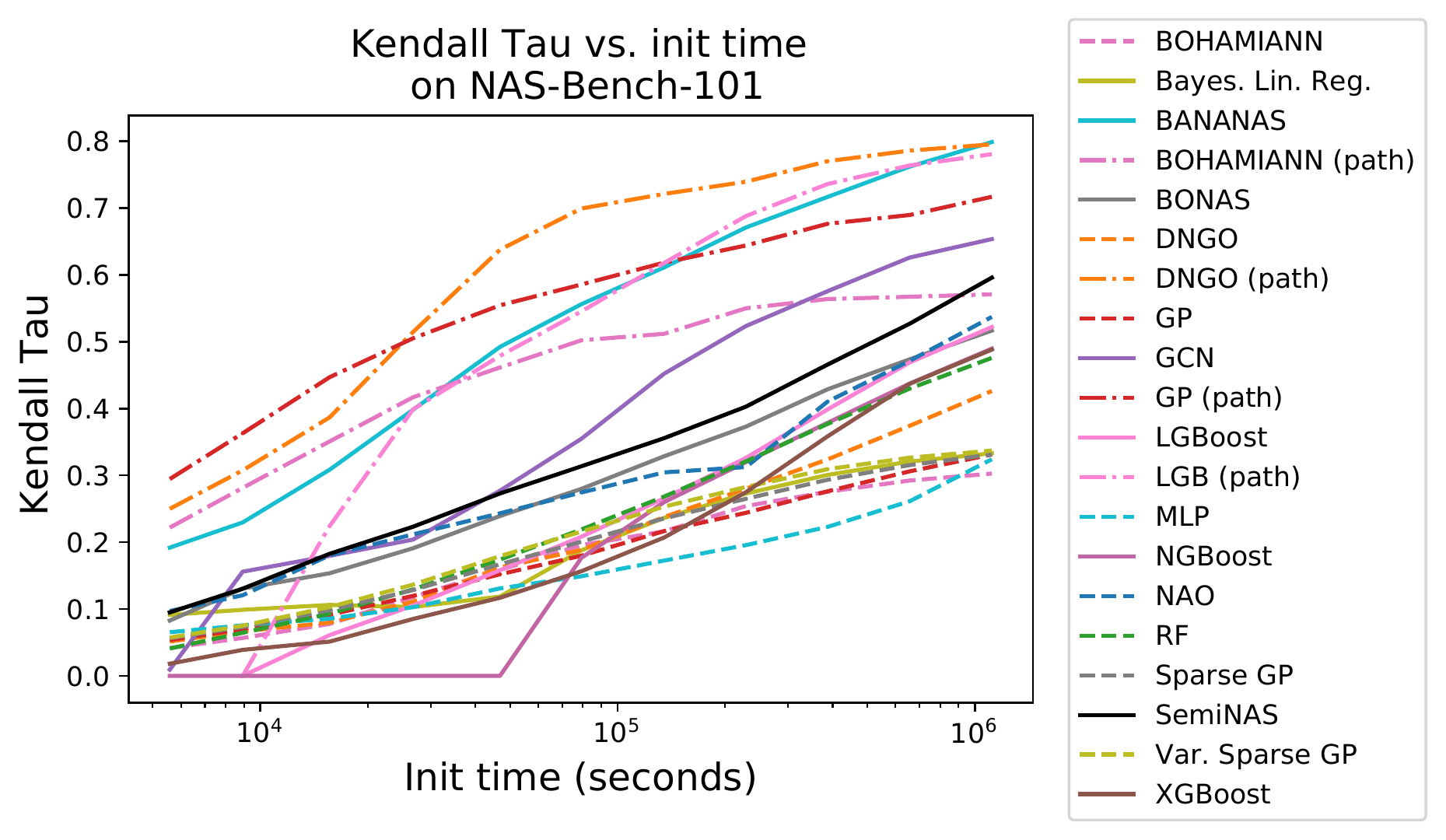}
\caption{
Kendall Tau rank correlation for predictors on NAS-Bench-101.
The GP-based and tree-based methods are not competitive until
used with the path encoding.}
\label{fig:nb101}
\end{figure}

Finally, recall that in Section~\ref{sec:experiments}, we discussed
the differences between NAS-Bench-101 and the other search spaces.
We mentioned one technical reason (the NAS-Bench-101 API only gives validation
accuracies at four epochs, and does not give the training loss for any
epochs) and one reason based on the differences in the search space itself:
the NAS-Bench-101 search space is more complex than NAS-Bench-201 and DARTS
because it allows any graph topology. Therefore, the path encoding is
particularly well-suited for NAS-Bench-101.
To test this explanation, in Figure~\ref{fig:nb101}
we run several of the simpler tree-based and GP-based 
predictors using the path encoding, and we see that these methods now surpass BANANAS.

\subsection{OMNI Details and Ablation}\label{subsec:omni}
In this section, we present more details and experiments for OMNI.
Recall that OMNI combines strong predictors from three different families:
SoTL-E, Jacobian covariance, and either NGBoost or SemiNAS, from the families of learning curve methods,
zero-cost methods, and model-based methods, respectively.
See Algorithm~\ref{alg:omni} for pseudo-code. 
Recall from Section~\ref{sec:prelim} that performance predictors have an initialization
stage that is called once, and a query stage that is called many times throughout the 
NAS algorithm.

\begin{algorithm}[t]
\caption{OMNI predictor}
\label{alg:omni}
\begin{algorithmic}
\STATE {\bfseries Input:} Search space $A$, dataset $D$, initialization time budget $B_{\text{init}}$, query time budget $B_{\text{query}}$.\\
\STATE \textbf{Initialization():}\\
 \begin{itemize}[topsep=0pt, itemsep=2pt, parsep=0pt, leftmargin=5mm]
     \item $\mathcal{D}_{\text{train}} \gets \emptyset$ 
     \item While $t<B_{\text{init}}$
     \begin{itemize}[topsep=0pt, itemsep=2pt, parsep=0pt, leftmargin=5mm]
         \item Draw an architecture $a$ randomly from $A$
         \item Train $a$ to completion to compute val.\ accuracy $y_a$
         \item $\mathcal{D}_{\text{train}} \gets \mathcal{D}_{\text{train}} \cup \{ (a, y_a) \}$
     \end{itemize}
     \item  Train an NGBoost model $m$ to predict the final val.\ accuracy of architectures from $\mathcal{D}_{\text{train}}$, using the architecture encoding, SoTL-E, and Jacob.\ cov.\ as input features.\\
 \end{itemize}
\STATE \textbf{Query}(architecture $a_{\text{test}}$):\\
\begin{itemize}[topsep=0pt, itemsep=2pt, parsep=0pt, leftmargin=5mm]
    \item While $t<B_{\text{query}}$, train $a_{\text{test}}$
    \item Compute SoTL-E using the partial learning curve, and compute Jacob.\ cov., 
    and the arch.\ encoding of $a_{\text{test}}$
    \item Predict val.\ acc.\ of $a_{test}$ using $m$ and the above features.
\end{itemize}
\end{algorithmic}
\end{algorithm}

Now we give an ablation study for OMNI.
We consider four different versions of OMNI:
(NGBoost + Jacob.\ Cov.), (SoTL-E + NGBoost), (Jacob.\ Cov.\ + SoTL-E), 
and (NGBoost + Jacob.\ Cov.\ + SoTL-E). Note that Jacob.\ Cov.\ + SoTL-E is created by using
Jacob.\ Cov.\ and SoTL-E as features in NGBoost without the architecture encoding as features.
All other OMNI variants use NGBoost with the architecture encodings as features.

\begin{figure*}[t]
\centering
\raisebox{0.0\height}{\includegraphics[width=.32\columnwidth]{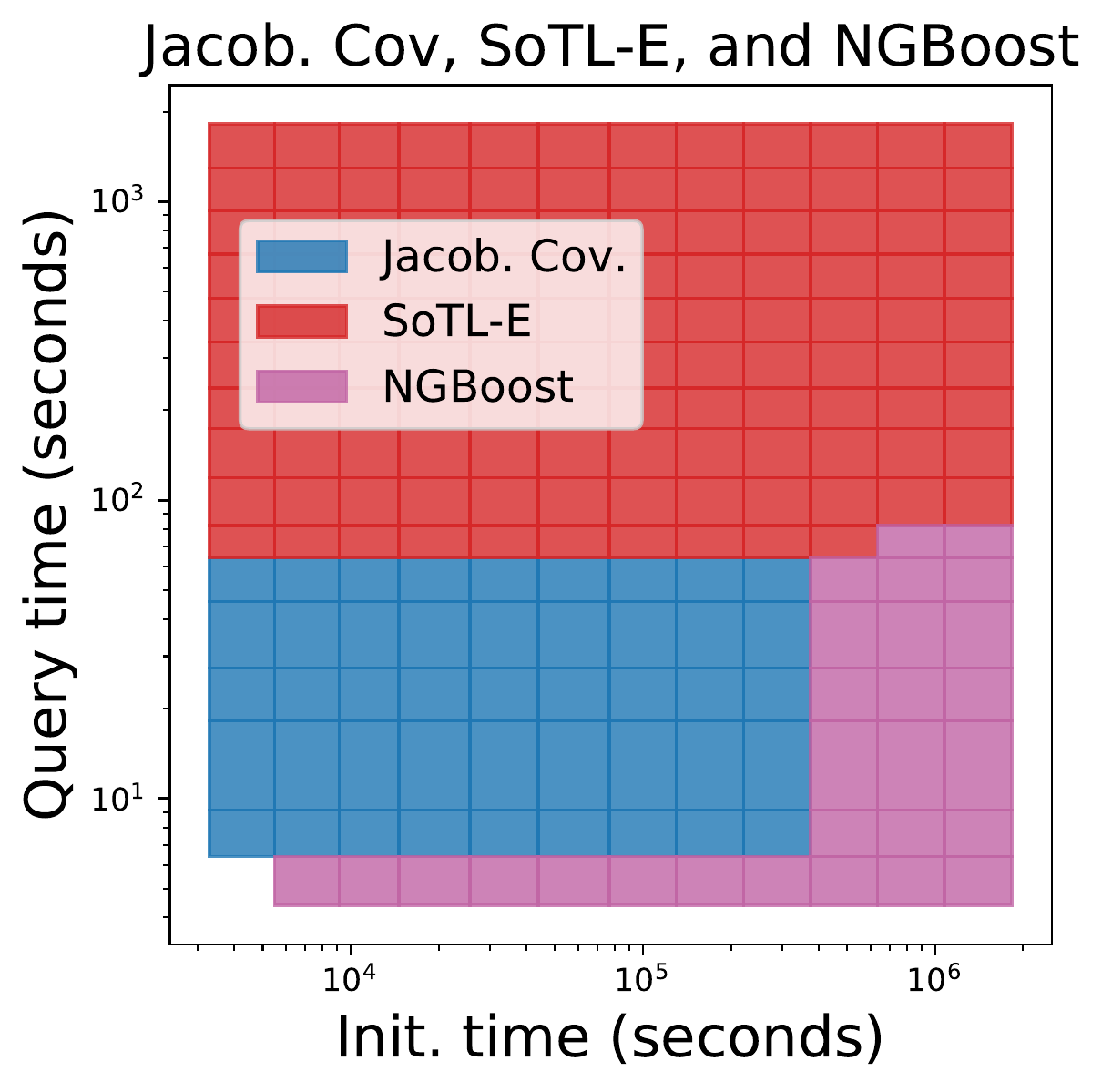}}
\raisebox{0.0\height}{\includegraphics[width=.32\columnwidth]{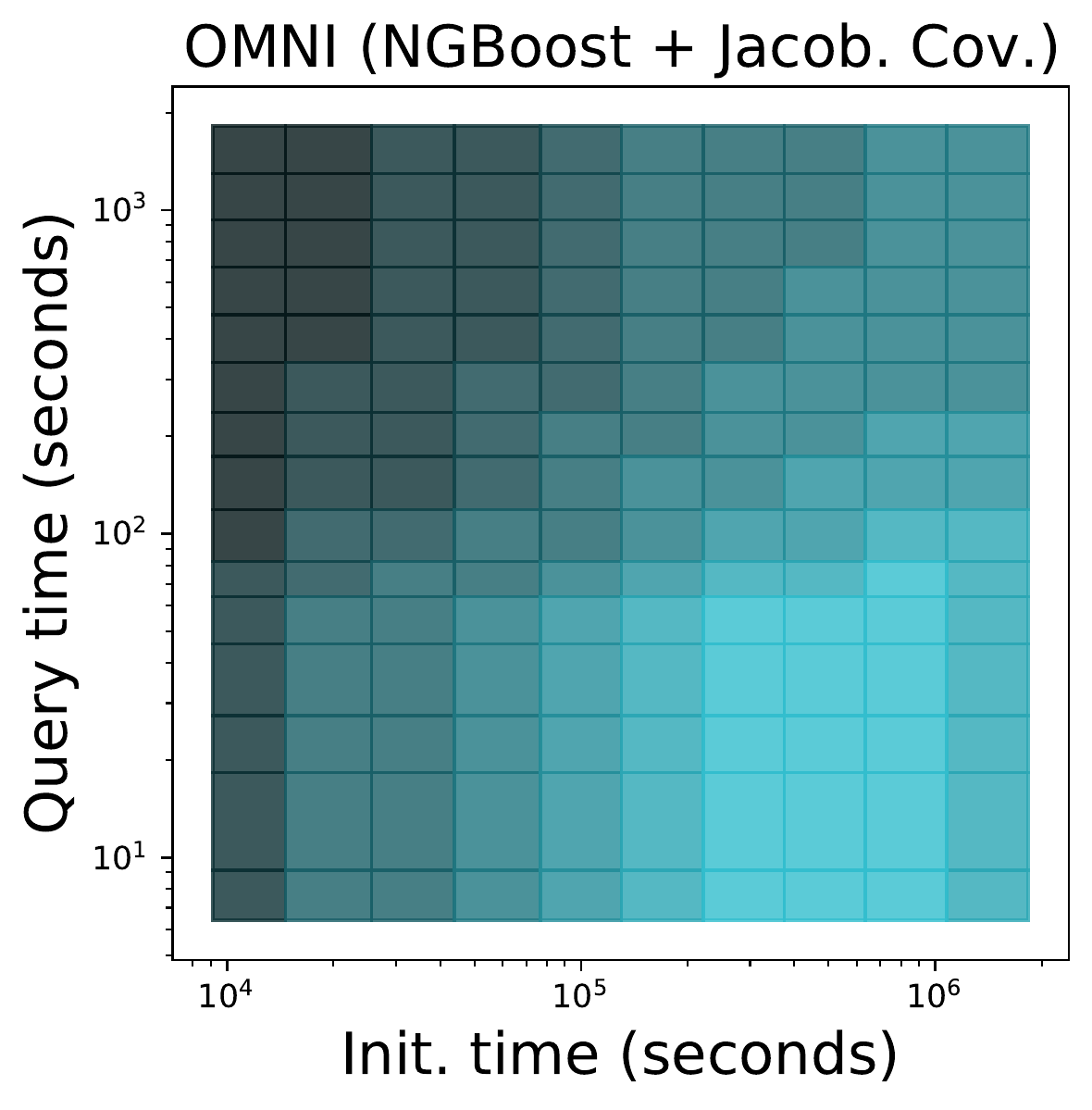}}
\raisebox{0.0\height}{\includegraphics[width=.32\columnwidth]{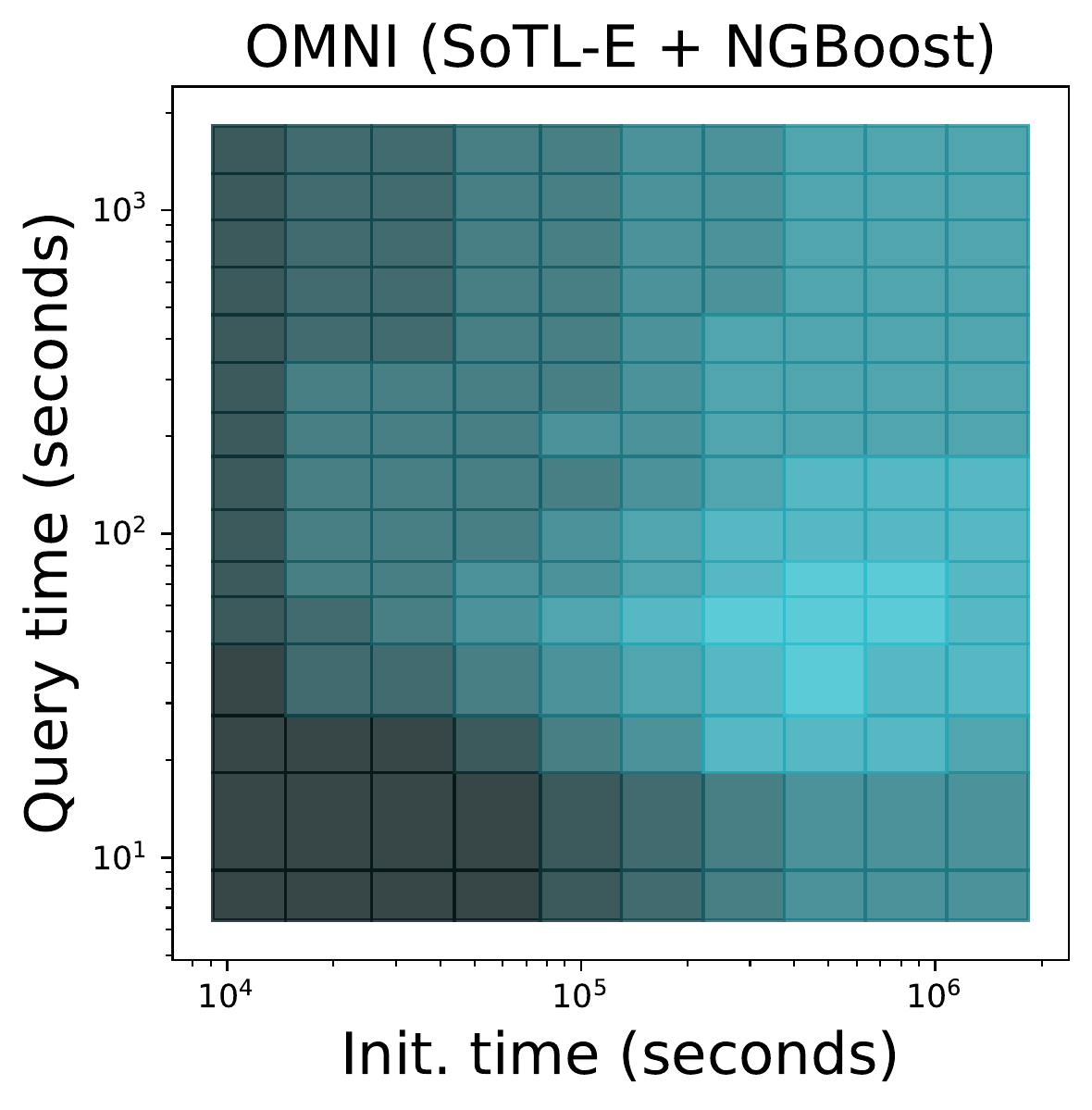}}
\raisebox{0.0\height}{\includegraphics[width=.32\columnwidth]{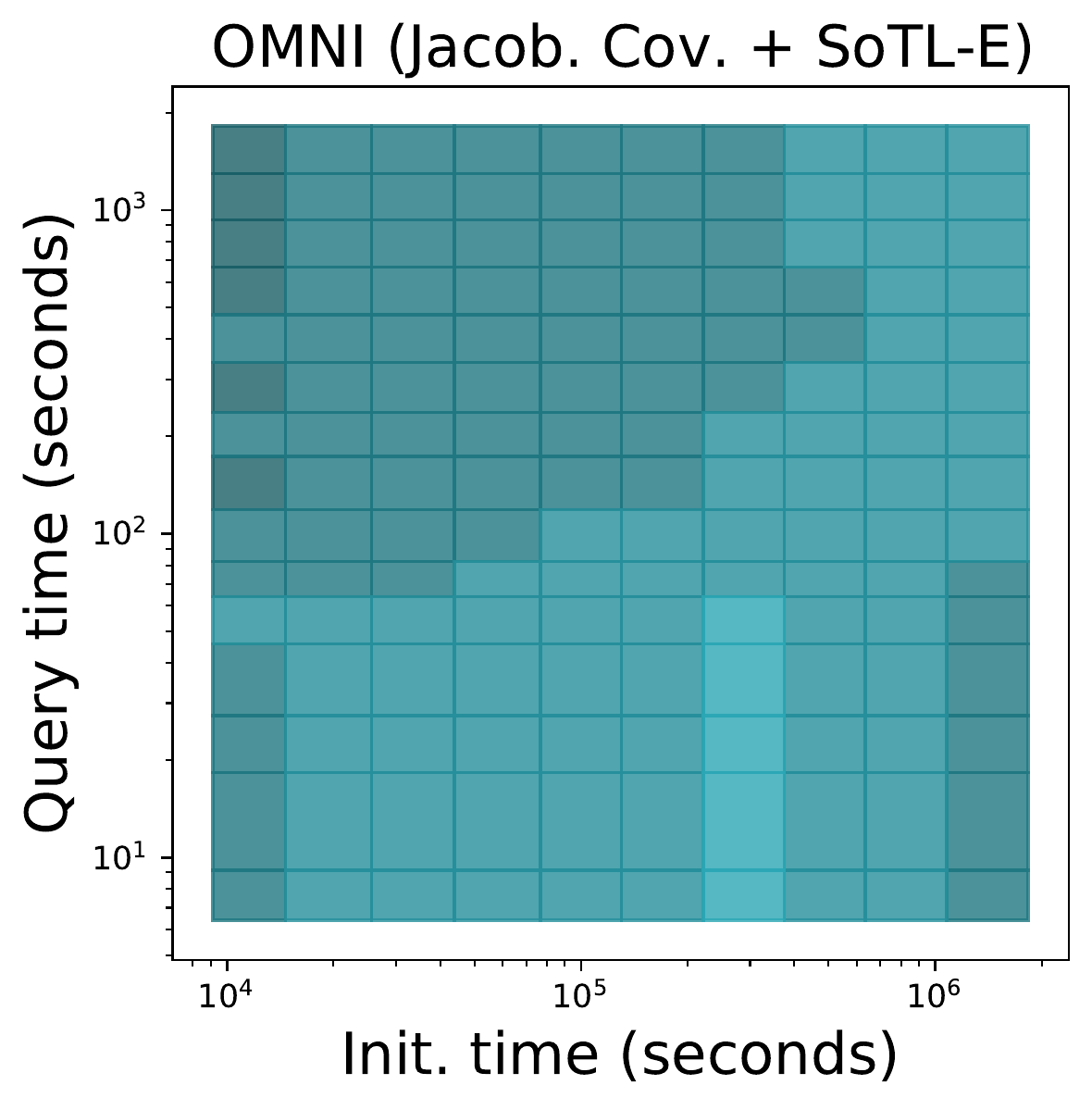}}
\raisebox{0.0\height}{\includegraphics[width=.32\columnwidth]{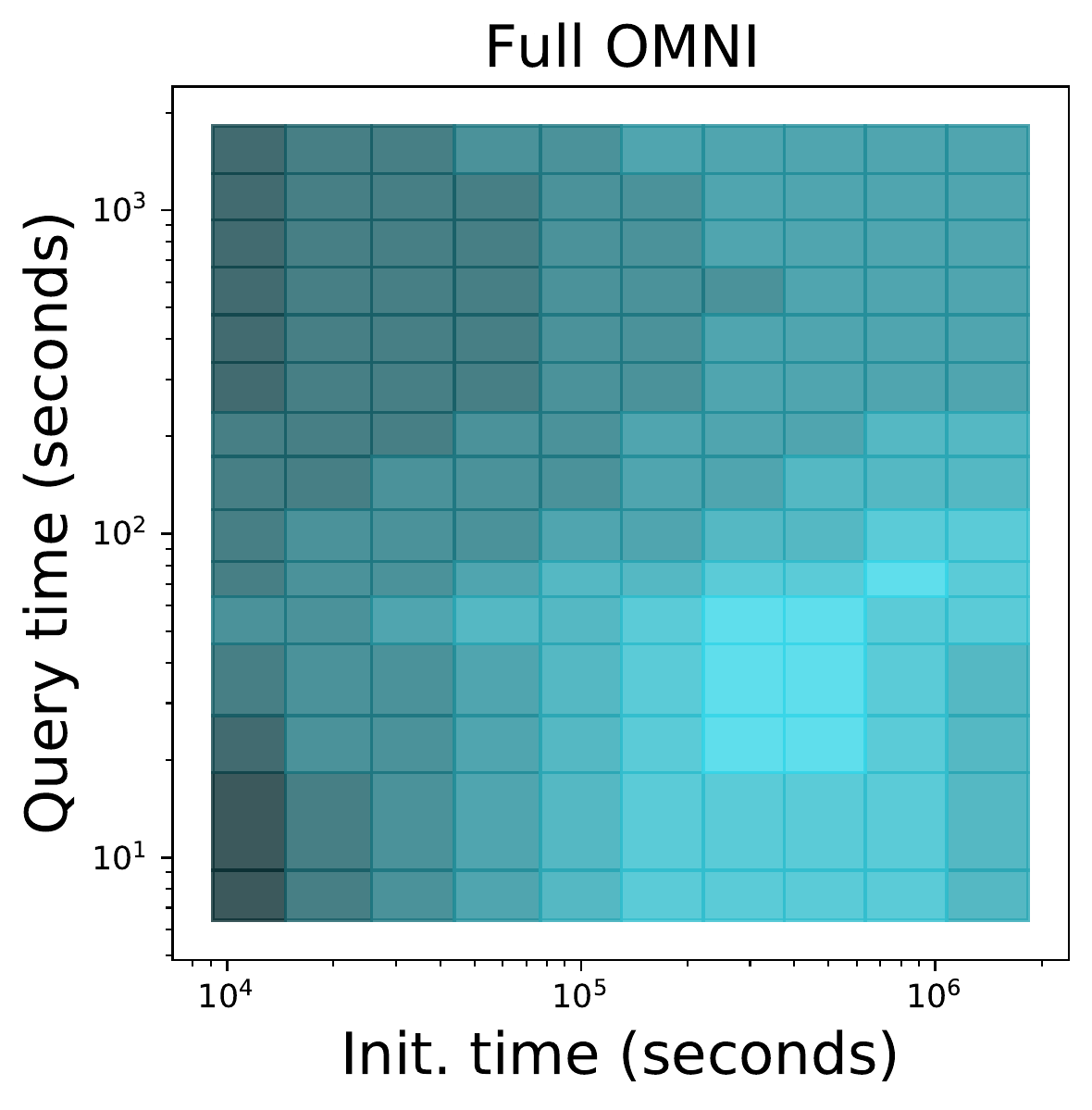}}
\raisebox{0.1\height}{\includegraphics[width=.14\columnwidth]{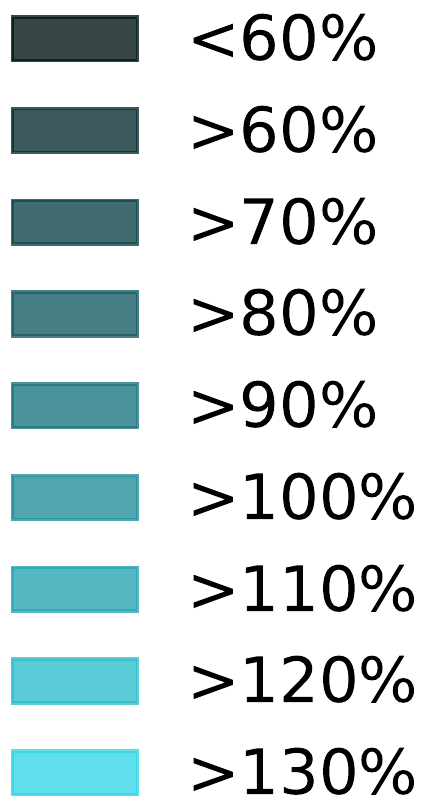}}
\caption{Ablation for OMNI. The first plot shows the best predictor in
\{Jacob.\ Cov.\, SoTL-E, and NGBoost\} for each initialization time and query time budget.
In the next four plots, we compute the percentage of the Kendall Tau value of the given OMNI
variant, compared to the first plot. For example, in the bottom-right corner of the first plot,
NGBoost achieves the highest Kendall Tau value from the set \{Jacob.\ Cov.\, SoTL-E, and NGBoost\};
In the bottom-right corner of the second plot, (NGBoost + Jacob.\ Cov.) achieves a Kendall Tau 
value that is 10\% higher than the Kendall Tau value achieved by NGBoost.
Therefore, combining Jacob.\ Cov.\ with NGBoost achieved a higher Kendall Tau 
value than the best individual predictor.
}
\label{fig:omni_ablation}
\end{figure*}

In Figure~\ref{fig:omni_ablation}, we plot the percentage of Kendall Tau compared to the best 
predictor in  \{Jacob.\ Cov.\, SoTL-E, and NGBoost\} for each initialization time and query time budget,
for each OMNI variant. 
We see that the variants (NGBoost + Jacob.\ Cov.) and (SoTL-E + NGBoost)
both give 20\% Kendall Tau improvements for some budget constraints, but do not perform well
for other budget constraints.
On the other hand, (Jacob.\ Cov.\ + SoTL-E) has fairly consistent performance across all budget
constraints. Finally, (NGBoost + Jacob.\ Cov.\ + SoTL-E) has consistent performance everywhere
and peaks at 30\% improvement.
This ablation study shows that predictors from all three families are needed to achieve
maximum performance.

\begin{figure}
\centering
\includegraphics[width=.24\columnwidth]{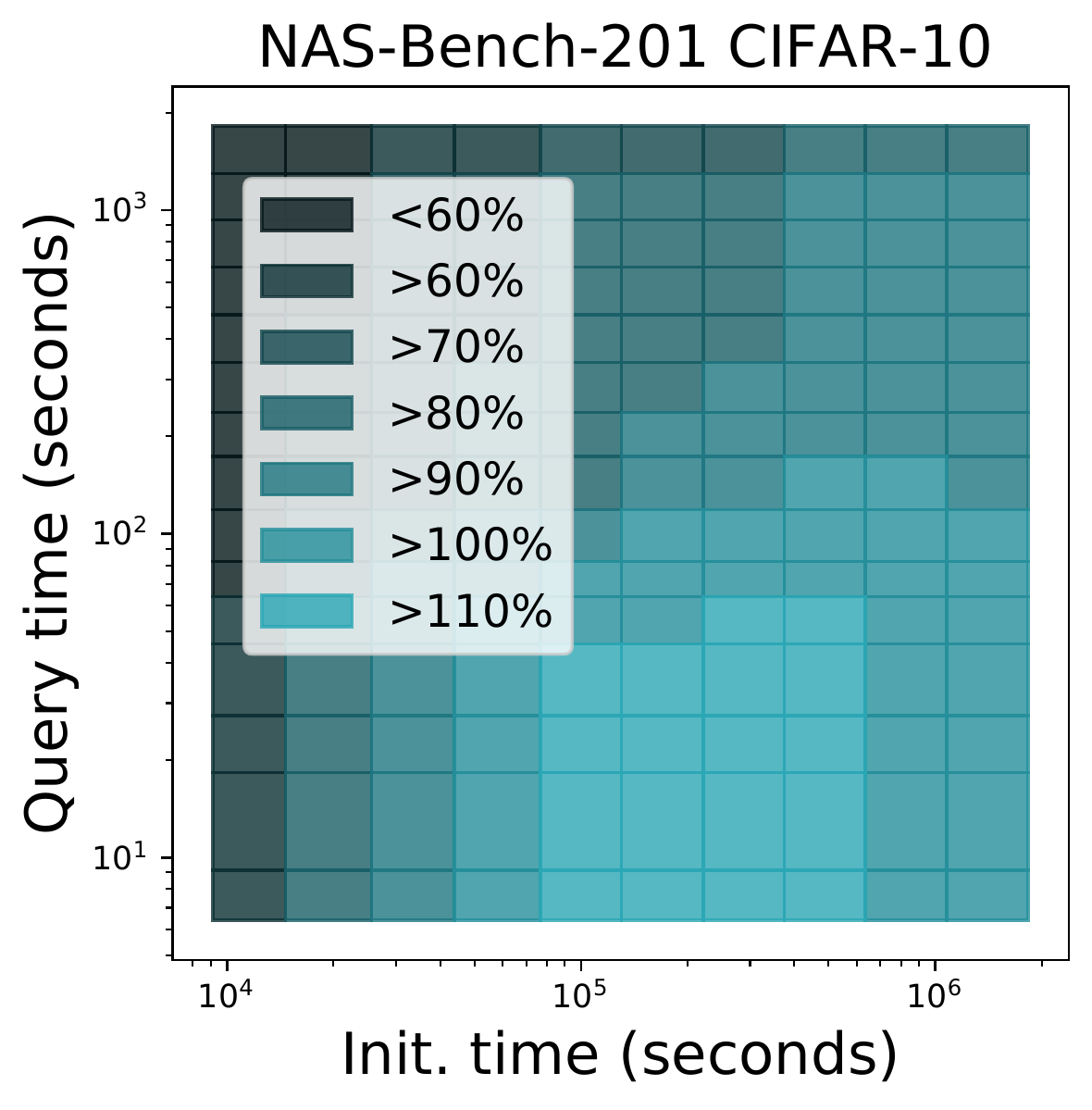}
\includegraphics[width=.24\columnwidth]{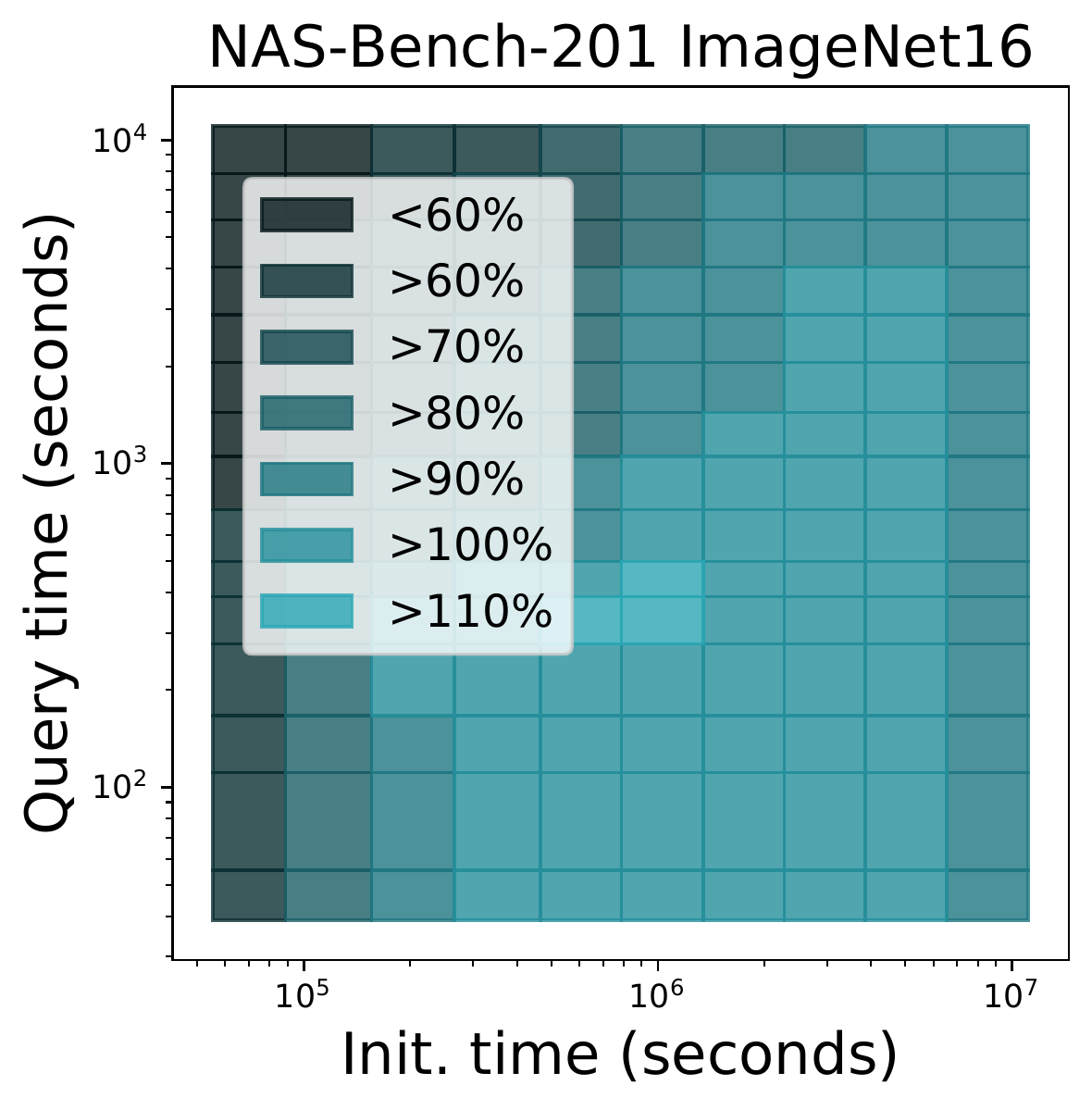}
\caption{Percentage of OMNI (SemiNAS)'s Kendall Tau value compared to 
the next-best predictors for each budget constraint.}
\label{fig:omni_seminas}
\end{figure}

\begin{figure}
\centering
\includegraphics[width=.24\columnwidth]{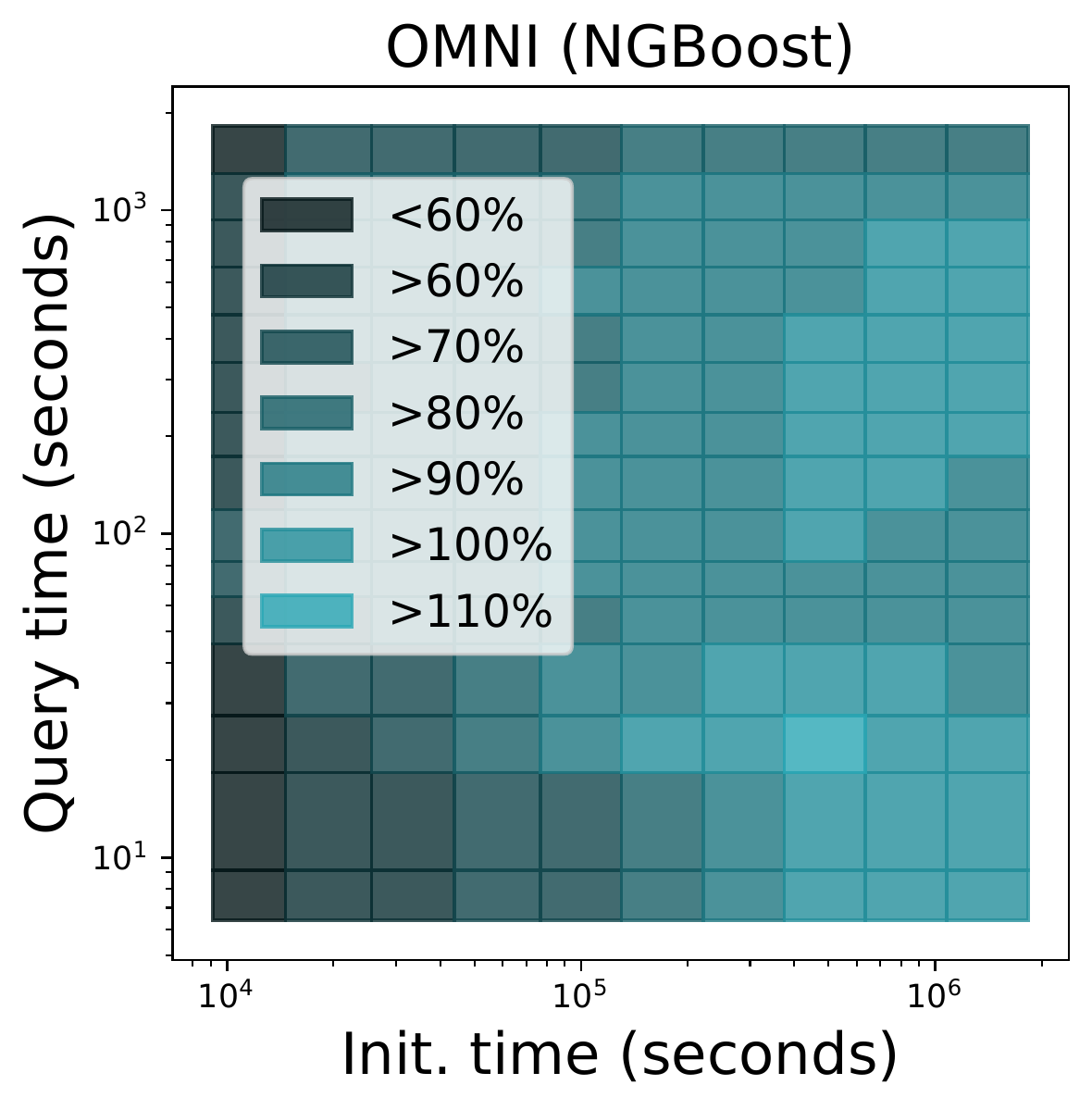}
\includegraphics[width=.24\columnwidth]{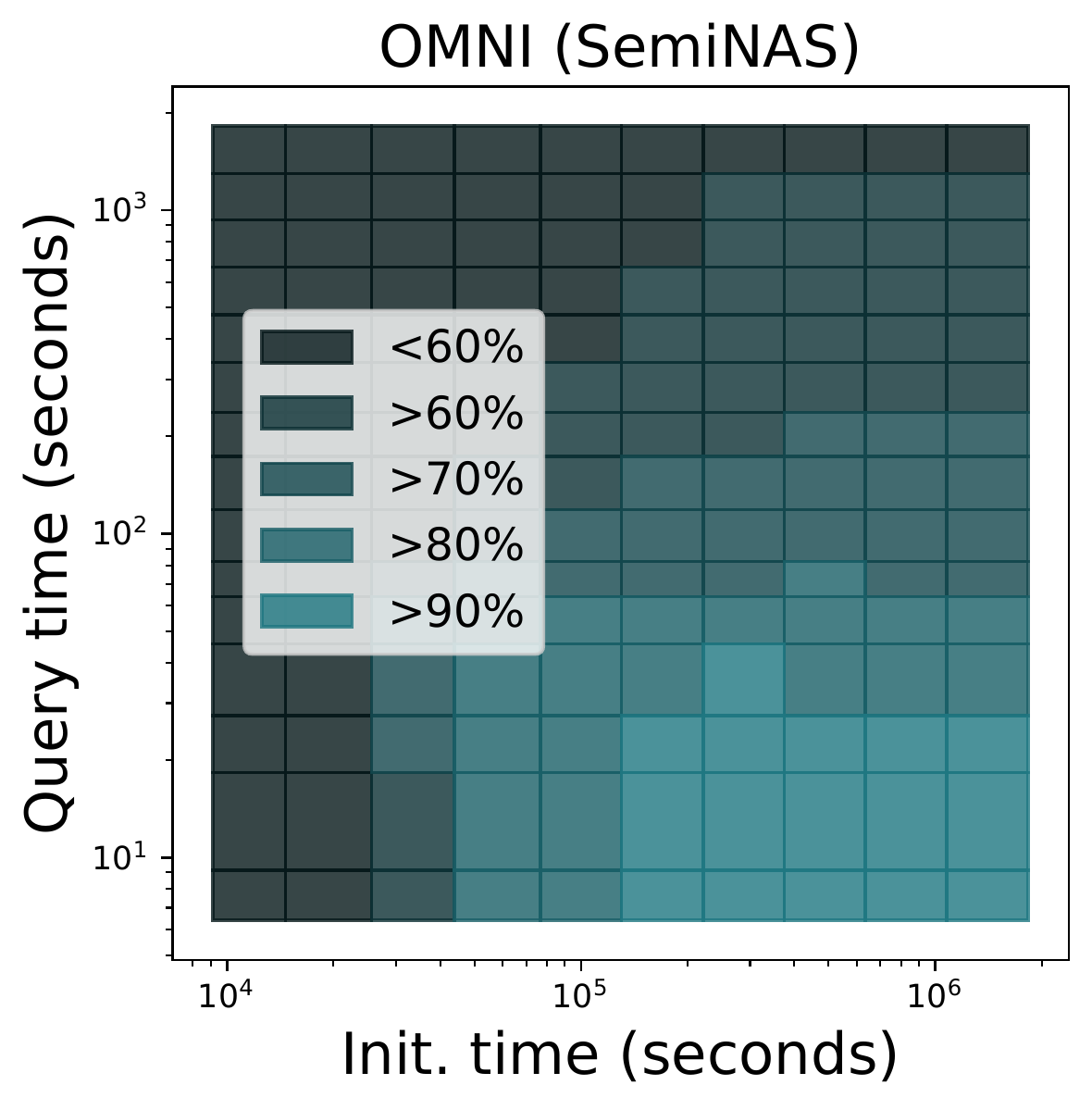}
\caption{Percentage of OMNI's Kendall Tau value compared to 
the next-best predictors for each budget constraint, in the mutation-based setting on NAS-Bench-201 CIFAR-10.}
\label{fig:omni_mutation}
\end{figure}

Next, we give additional experiments for OMNI in other settings.
Note that Figure \ref{fig:omni} used the version of OMNI with NGBoost. 
See Figure \ref{fig:omni_seminas} for the version of OMNI with SemiNAS, which performs worse.
Finally, see Figure \ref{fig:omni_mutation} for the performance of OMNI in the mutation-based setting
described in Section \ref{sec:experiments}. Note that it performs comparatively worse than in the standard
uniformly random setting.

\section{Additional Experiments on NAS-Bench-201}\label{app:experiments_201}

In this section, we give additional experiments carried out on NAS-Bench-201 CIFAR-10.
These experiments include implementing additional hybrid predictors, running zero-cost proxy baselines, computing the variance in random seeds when the training and testing datasets are fixed, and running the predictors with longer HPO budgets.

\subsection{Additional hybrid predictors}

In Section \ref{sec:experiments}, we implemented one predictor that was a hybrid learning curve + model-based method: LcSVR \citep{baker2017accelerating}. In this section, we implement two other hybrid predictors: LCNet \citep{lcnet} and ``LCE using previous builds'' \citep{chandrashekaran2017speeding}.
For both techniques, we used the original implementation.

We compute the Kendall Tau rank correlation of LcSVR, LCNet, and LC-prev-builds on a representative set of four different initialization times and three different query times, for twelve total settings on NAS-Bench-201 CIFAR-10. See Table \ref{tab:hybrids}.
We find that LC-prev-builds outperforms LcSVR when the initialization time is small, 
and when the query time is high. LCNet never outperformed LcSVR.

\begin{table}[t]
\caption{Comparison of hybrid learning curve + model-based methods on NAS-Bench-201 CIFAR-10, reporting the mean Kendall Tau rank correlation along with the standard deviation.}
\centering
\begin{tabular}{@{}c|c|c|c|c@{}}
\toprule
\multicolumn{1}{l}{\textbf{Init.\ time (s)}} & \multicolumn{1}{c}{\textbf{Query time (s)}} & \multicolumn{1}{c}{\textbf{LC-prev-builds}} & \multicolumn{1}{c}{\textbf{LCNet}} & \multicolumn{1}{c}{\textbf{LcSVR}} \\
\midrule 
$1.4e4$ & $64$ & $\mathbf{0.512\pm0.043}$ & $0.322\pm0.128$ & $0.427\pm0.181$ \\
$1.4e4$ & $238$ & $\mathbf{0.676\pm0.029}$ & $0.204\pm0.209$ & $0.484\pm0.172$ \\
$1.4e4$ & $932$ & $\mathbf{0.780\pm0.019}$ & $0.317\pm0.265$ & $0.497\pm0.174$ \\
$7.6e4$ & $64$ & $0.516\pm0.042$ & $0.388\pm0.104$ & $\mathbf{0.596\pm0.065}$ \\
$7.6e4$ & $238$ & $0.688\pm0.026$ & $0.471\pm0.084$ & $\mathbf{0.700\pm0.052}$ \\
$7.6e4$ & $932$ & $\mathbf{0.797\pm0.016}$ & $0.583\pm0.090$ & $0.740\pm0.048$ \\
$2.2e5$ & $64$ & $0.523\pm0.041$ & $0.418\pm0.074$ & $\mathbf{0.632\pm0.044}$ \\
$2.2e5$ & $238$ & $0.692\pm0.023$ & $0.472\pm0.097$ & $\mathbf{0.736\pm0.034}$ \\
$2.2e5$ & $932$ & $\mathbf{0.797\pm0.014}$ & $0.613\pm0.077$ & $0.795\pm0.023$ \\
$1.1e6$ & $64$ & $0.528\pm0.039$ & $0.452\pm0.081$ & $\mathbf{0.667\pm0.045}$ \\
$1.1e6$ & $238$ & $0.698\pm0.025$ & $0.472\pm0.101$ & $\mathbf{0.759\pm0.032}$ \\
$1.1e6$ & $932$ & $0.798\pm0.014$ & $0.637\pm0.076$ & $\mathbf{0.823\pm0.023}$ \\
\bottomrule
\end{tabular}
\label{tab:hybrids}
\end{table}

\subsection{Zero-cost proxy baselines}
In this section, we implement flops and params as baselines for zero-cost predictors. 
We compare the Kendall Tau rank correlation and Spearman rank correlation on NAS-Bench-201 CIFAR-10 for flops, params, and all of the zero-cost proxies implemented in Section \ref{sec:experiments}. Note that zero-cost predictors have just one setting: no initialization time and a query time of 5 seconds.
See Table \ref{tab:zerocost}.
Surprisingly, flops and params tie for the second highest Kendall Tau value out of all six zero-cost predictors on NAS-Bench-201 CIFAR-10, behind Jacobian covariance. For Spearman values, flops and params tie for third behind Jacobian covariance and SynFlow.
Note that for the case of NAS-Bench-201, since the graph structure of the cell is fixed, flops has a one-to-one relationship with params (meaning they get the same rank correlation results).

\begin{table}[t]
\caption{Comparison of baseline methods to zero-cost proxies on NAS-Bench-201 CIFAR-10.}
\centering
\begin{tabular}{@{}l|c|c@{}}
\toprule
\multicolumn{1}{l}{\textbf{Method}} & \multicolumn{1}{c}{\textbf{Kendall Tau}} & \multicolumn{1}{c}{\textbf{Spearman}} \\
\midrule 
Flops & 0.539 & 0.713 \\
Params & 0.539 & 0.713 \\
\midrule
Fisher & 0.218 & 0.299 \\
Grad Norm & 0.420 & 0.587 \\
Grasp & 0.330 & 0.481 \\
Jacob. Cov. & \textbf{0.575} & \textbf{0.743} \\
SNIP & 0.422 & 0.590 \\
SynFlow & 0.530 & 0.724 \\
\bottomrule
\end{tabular}
\label{tab:zerocost}
\end{table}

\subsection{Random seed experiments}
Our plots in Figure \ref{fig:app} give the standard deviation across 100 trials for each predictor across different random seeds, which varies the tran and test sets as well as any stochasticity of the predictor. Now, we perform another experiment where we keep the train and test sets fixed and measure the standard deviation across only the stochasticity of the predictor. For a representative set of eight predictors on NAS-Bench-201 CIFAR-10, we computed the standard deviation of 10 trials on a fixed train and test set, averaged over 50 trials of choosing new train and test sets (500 trials total per predictor). We used the median settings of initialization time and query time and reported the Kendall Tau rank correlation.
See Table \ref{tab:seeds}.
We find that Bayesian Linear Regression has the highest stochasticity.

\begin{table}[t]
\caption{Standard deviation of predictors when train and test sets are fixed.}
\centering
\begin{tabular}{@{}l|c|c|c@{}}
\toprule
\multicolumn{1}{l}{\textbf{Method}} & \multicolumn{1}{c}{\textbf{Mean}} & \multicolumn{1}{c}{\textbf{Std.\ Dev.}} & \multicolumn{1}{c}{\textbf{Std.\ Dev.\ w.\ fixed datasets}} \\
\midrule 
BANANAS & 0.254 & 0.050 & 0.032 \\
Bayes. Lin. Reg. & 0.291 & \textbf{0.065} & \textbf{0.115} \\
Jacob. Cov & 0.539 & 0.000 & 0.006 \\
NGBoost & 0.355 & 0.059 & 0.032 \\
SoTL-E & \textbf{0.623} & 0.032 & 0.000 \\
SynFlow & 0.529 & 0.000 & 0.001 \\
Var. Sparse GP & 0.486 & 0.054 & 0.000 \\
XGBoost & 0.390 & 0.052 & 0.031 \\
\bottomrule
\end{tabular}
\label{tab:seeds}
\end{table}

\subsection{Longer HPO budgets}

In Section \ref{sec:experiments}, all model-based predictors were run using 15 minutes of hyperparameter tuning. In this section, we test whether there is further improvement when increasing the hyperparameter tuning budget from 15 minutes to 1 hour.
We run this experiment on a representative set of ten model-based predictors on NAS-Bench-201 CIFAR-10, across all eleven initialization time settings from Section \ref{sec:experiments} (the query time for all model-based predictors is fixed at one second).
For each predictor, after computing the improvement to Kendall Tau rank correlation when increasing the hyperparameter tuning budget, we report the average improvement over all initialization time settings, as well as the average of the top two and average of the worst two initialization time settings.
See Table \ref{tab:1hr_hpo}.

There were no predictors which achieved non-negligible improvement across all initialization time settings. 
Similarly, none of the predictors saw non-negligible decreases in performance, although BONAS had the lowest worst two average, at -0.020, likely due to additional overfitting that happens in the 1 hour hyperparameter tuning budget.
For the top two average improvements, BANANAS saw the biggest improvements, and MLP, BONAS, and GCN 
also saw non-negligible improvements. The rest of the performance predictors saw little to no improvement. Interestingly, the largest improvements were all from deep learning based predictors.

\begin{table}[t]
\caption{Standard deviation of predictors when train and test sets are fixed.}
\centering
\begin{tabular}{@{}l|c|c|c@{}}
\toprule
\multicolumn{1}{l}{\textbf{Method}} & \multicolumn{1}{c}{\textbf{Avg.}} & \multicolumn{1}{c}{\textbf{Worst Two Avg.}} & \multicolumn{1}{c}{\textbf{Top Two Avg.}} \\
\midrule 
BANANAS & 0.012 & -0.010 & \textbf{0.059} \\
Bayes. Lin. Reg. & 0.002 & -0.006 & 0.014 \\
BOHAMIANN & 0.001 & -0.003 & 0.006 \\
BONAS & 0.008 & \textbf{-0.020} & 0.036 \\
DNGO & 0.000 & -0.008 & 0.007 \\
GCN & 0.008 & -0.016 & 0.035 \\
GP & 0.003 & -0.004 & 0.009 \\
MLP & 0.009 & -0.009 & 0.039 \\
Sparse GP & -0.003 & -0.008 & 0.006 \\
Var. Sparse GP & \textbf{0.017} & -0.002 & 0.032 \\
\bottomrule
\end{tabular}
\label{tab:1hr_hpo}
\end{table}

\section{Reproducibility Table} \label{app:reproducibility}

When giving a large-scale comparison of performance predictors, it is important that all of the methods are accurately implemented and optimized. We took a number of reasonable steps to ensure this, including \emph{(1)} using the original implementations whenever possible, and \emph{(2)} applying light hyperparameter tuning to all methods, to help control for the fact that different techniques received different amounts of hyperparameter tuning in their original release. The best way to ensure that all methods are properly implemented is to reproduce the results reported by the original authors of a given technique.

In this section, we present a table to clarify for which of the 31 predictors we have succeeded in reproducing published results, and for which predictors it is not possible. For each predictor, we mark whether the original paper (or first paper to run on a NAS benchmark) had \emph{(1)} at least one search space out of the ones we used, \emph{(2)} at least one initialization and query time that matches one of our settings, and \emph{(3)} at least one metric from our set of metrics. If yes to \emph{(1)}-\emph{(3)}, then we check whether we \emph{(4)} achieved nearly the same numbers as the original paper. See Table \ref{tab:reproduce}.

Of the 31 predictors, 14 were released before any of the NAS-Bench search spaces had come out. Two more did not give experiments in the setting of our paper. We are able to fairly compare the remaining 15 performance predictors (up to smaller changes in the experimental setting, such as different test set sizes) with either the original paper or the first paper to give results on a NAS benchmark. Our results are close to the original results, or in some cases stronger (due to our use of HPO). All 15 are within 0.04 of the reported rank correlation value or higher.


\begin{table}[t]
\caption{Reproducibility table, clarifying which predictors we have reproduced with respect to existing published results.\\
\footnotesize{
$^1$ Compared with first follow-up paper to give correlation results on a NAS benchmark.
$^2$ Larger test set.
$^3$ Approximated from a plot.
}
}
\label{tab:reproduce}
    
\centering
    \resizebox{\textwidth}{!}{
\begin{tabular}{@{}lclllccc@{}}
\toprule
\textbf{Predictor} & \textbf{Paper} & \textbf{Search space} & \textbf{Setting} & \textbf{Metric} & \textbf{Value} & \textbf{Ours} & \textbf{Diff.}\\
\toprule
BANANAS & \citep{bananas} & NAS-Bench-101 & 1000 train, & Pearson & 0.699 & 0.904 & +0.205 \\
&&& 100 test \\
\midrule
Bayes.\ Lin.\ Reg.\ & \citep{bishop2006pattern} & \multicolumn{3}{l}{Released before NAS-Bench search spaces} \\
\midrule
BOHAMIANN & \citep{springenberg2016bayesian} & \multicolumn{3}{l}{Released before NAS-Bench search spaces} \\
\midrule
BONAS & \citep{bonas} & NAS-Bench-101 & \multicolumn{3}{l}{Only used size 360k train set.} \\
\midrule
DNGO & \citep{snoek2015scalable} & \multicolumn{3}{l}{Released before NAS-Bench search spaces}  \\
\midrule
Early Stop. & \citep{ru2020revisiting}$^1$ & NAS-Bench-201 & 100 epochs, & Spearman & 0.85$^3$ & 0.850 & +0.0 \\
ValAcc &&& 1000 test$^2$ \\
\midrule
Early Stop. & \citep{ru2020revisiting}$^1$ & NAS-Bench-201&100 epochs, &Spearman&0.83$^3$ & 0.839 & +0.009 \\
ValLoss &&& 1000 test$^2$ \\
\midrule
Fisher & \citep{abdelfattah2021zerocost}$^1$ & NAS-Bench-201 & 15k test$^2$ & Spearman & 0.36 & 0.328 & -0.032 \\
\midrule
GCN & \citep{wen2019neural} & NAS-Bench101$^1$ & 1000 train, & Pearson & 0.607 & 0.793 & +0.186 \\
&&& 100 test \\
\midrule
GP & \citep{rasmussen2003gaussian} & \multicolumn{3}{l}{Released before NAS-Bench search spaces} \\
\midrule
Grad Norm & \citep{abdelfattah2021zerocost} & NAS-Bench-201 & 15k test$^2$ & Spearman & 0.58 & 0.587 & +0.007 \\
\midrule
Grasp & \citep{abdelfattah2021zerocost}$^1$ & NAS-Bench-201 & 15k test$^2$ & Spearman & 0.48 & 0.481 & +0.001 \\
\midrule
Jacob. Cov. & \citep{abdelfattah2021zerocost} & NAS-Bench-201 & 1000 test$^2$ & Pearson & 0.574 & 0.575 & +0.001 \\
\midrule
LCE & \citep{domhan2015speeding} & \multicolumn{3}{l}{Released before NAS-Bench search spaces} \\
\midrule
LCE-m & \citep{lcnet} & \multicolumn{3}{l}{Released before NAS-Bench search spaces} \\
\midrule
LcSVR & \citep{ru2020revisiting} & NAS-Bench-201$^1$ & 100 epochs, & Spearman & 0.93$^3$ & 0.931 & +0.001 \\
&&& 100 train, \\
&&& 1000 test$^2$\\
\midrule
LGBoost & \citep{ke2017lightgbm} & NAS-Bench101 & 1000 train, & Kendall Tau & 0.640 & 0.705 & +0.065 \\
&&& 100 test \\
\midrule
MLP & \citep{bananas}$^1$ & NAS-Bench101 & 1000 train, & Pearson & 0.400 & 0.447 & +0.047 \\
&&& 100 test \\
\midrule
NAO & \citep{luo2018neural} & \multicolumn{3}{l}{Does not have NAS-Bench search spaces} \\
\midrule
NGBoost & \citep{duan2020ngboost} & \multicolumn{3}{l}{Released before NAS-Bench search spaces} \\
\midrule
OneShot & \citep{zela2020understanding} & \multicolumn{3}{l}{Released before NAS-Bench search spaces} \\
\midrule
Rand. Forest & \citep{liaw2002classification} & \multicolumn{3}{l}{Released before NAS-Bench search spaces} \\
\midrule
RSWS & \citep{randomnas} & \multicolumn{3}{l}{Released before NAS-Bench search spaces} \\
\midrule
SemiNAS & \citep{seminas} & NAS-Bench-101 & \multicolumn{3}{l}{No predictor experiments} \\
\midrule
SNIP & \citep{abdelfattah2021zerocost}$^1$ & NAS-Bench-201  & 15k test$^2$ & Spearman & 0.58 & 0.590 & +0.01 \\
\midrule
SoTL & \citep{ru2020revisiting} & NAS-Bench-201 & 100 epochs, & Spearman & 0.96$^3$ & 0.963 & +0.003 \\
&&& 1000 test$^2$ \\
\midrule
SoTL-E & \citep{ru2020revisiting} & NAS-Bench-201 & 100 epochs, & Spearman & 0.93$^3$ & 0.932 & +0.002 \\
&&& 1000 test$^2$ \\
\midrule
Sparse GP & \citep{candela05} & \multicolumn{3}{l}{Released before NAS-Bench search spaces} \\
\midrule
SynFlow & \citep{abdelfattah2021zerocost} & NAS-Bench-201 & 15k test$^2$ & Spearman & 0.74 & 0.738 & -0.002 \\
\midrule
Var. Sparse GP & \citep{titsias2009variational} & \multicolumn{3}{l}{Released before NAS-Bench search spaces} \\
\midrule
XGBoost & \citep{chen2016xgboost} & \multicolumn{3}{l}{Released before NAS-Bench search spaces} \\
\bottomrule
\end{tabular}
}
\end{table}

\end{document}